\documentclass[journal]{IEEEtran}
\usepackage{amsmath,amsfonts}
\usepackage{algorithmic}
\usepackage{algorithm}
\usepackage{array}
\usepackage[caption=false,font=normalsize,labelfont=sf,textfont=sf]{subfig}
\usepackage{textcomp}
\usepackage{stfloats}
\usepackage{url}
\usepackage{verbatim}
\usepackage{graphicx}
\usepackage{cite}
\usepackage{amssymb}
\usepackage{cuted}
\usepackage{amsfonts}
\usepackage{array}
\usepackage{adjustbox}   
\usepackage{makecell}    
\usepackage{booktabs}    
\usepackage{multirow}    
\usepackage{epsfig}
\usepackage{float}
\usepackage[hidelinks]{hyperref}
\usepackage{xr-hyper}
\usepackage{cleveref}
\usepackage[table]{xcolor}
\usepackage{mathrsfs}
\usepackage{newfloat}
\usepackage{listings}
\usepackage{url}
\newcommand{\PG}[1]{\textcolor{black}{#1}}
\newcommand{\YL}[1]{\textcolor{black}{#1}}

\DeclareRobustCommand{\pspcs}{pSp-cs }
\DeclareRobustCommand{\pspcsref}{pSp-cs-Ref }

\begin{document}

\title{Learning Common and Salient Generative Factors Between Two Image Datasets}

\author{Yunlong~He,
        Gwilherm~Lesné,
        Ziqian~Liu,
        Michael Soumm, 
        and~Pietro~Gori
\thanks{All authors are with LTCI, Télécom Paris, Institut Polytechnique de Paris, 19 Place Marguerite Perey, 91120 Palaiseau, France.}
\thanks{Corresponding author: Pietro Gori (email: pietro.gori@telecom-paris.fr).}
}


\maketitle

\begin{abstract}
Recent advancements in image synthesis have enabled high-quality image generation and manipulation. Most works focus on: 1) conditional manipulation, where an image is modified conditioned on a given attribute, or 2) disentangled representation learning, where each latent direction should represent a distinct semantic attribute. In this paper, we focus on a different and less studied research problem, called Contrastive Analysis (CA). Given two image datasets, we want to separate the common generative factors, shared across the two datasets, from the salient ones, specific to only one dataset. Compared to existing methods, which use attributes as supervised signals for editing (e.g., glasses, gender), the proposed method is weaker, since it only uses the dataset signal. We propose a novel framework for CA, that can be adapted to both GAN and Diffusion models, to learn both common and salient factors. By defining new and well-adapted learning strategies and losses, we ensure a relevant separation between common and salient factors, preserving a high-quality generation. We evaluate our approach on diverse datasets, covering human faces, animal images and medical scans. Our framework demonstrates superior separation ability and image quality synthesis compared to prior methods.
\end{abstract}




\section{Introduction}
\label{sec:intro}
\IEEEPARstart{M}odern generative models, including Generative Adversarial Networks (GANs) \cite{goodfellow2014generative} and Diffusion Models (DMs) \cite{ho2020denoising}, have achieved remarkable advances in high-quality image synthesis. Beyond generation, a growing line of work aims to understand and control their latent spaces to enable intuitive image manipulation. These efforts underpin a range of downstream tasks, such as image editing \cite{shen2020interpreting}, attribute transformation \cite{harkonen2020ganspace}, and medical image analysis \cite{chartsias2019disentangled}.

Latent-controlled image manipulation requires inverting real images into an interpretable latent space that supports both faithful reconstruction and semantic editing. Such spaces have been extensively explored in GANs, notably the StyleGAN family~\cite{melnik2024face}, and have recently drawn increasing attention in DMs~\cite{preechakul2022diffusion, Kwon2023SemanticLatent, jeong2024training}. After inversion, the latent representation is modified by moving towards a direction that can be found by either using attribute supervision (\textit{e.g.,} gender, age, hair color) \cite{Abdal2021StyleFlow}, text-supervision \cite{Baykal2023CLIPInverter} or in an ``unsupervised'' way \cite{harkonen2020ganspace, voynov2020unsupervised} (actually, ``unsupervised'' methods are all based on inductive biases or latent assumptions. For instance, ``interesting'' directions may be orthogonal or with large variations). Among these works, almost all of them focus on learning (or using) semantically meaningful and disentangled directions within a \textit{single} dataset. Some recent works have proposed local editing based on reference images \cite{collins2020editing, kim2021stylemapgan} or attributes (e.g., age) \cite{alaluf2021sam} but none of them, to the best of our knowledge, has proposed using advanced generative models to compare and analyze \textit{two datasets} at the same time.

In this article, we focus on a rather new research area, called \textit{Contrastive Analysis} (CA)~\cite{abid2018exploring,weinberger2022moment,louiset2024sepvae,louiset_separating_2024,carton2024double}.
Given two datasets, we assume that one dataset (target) has some modified/added patterns with respect to the second one (background). This is the most studied assumption and the objective is to \textit{identify} and \textit{separate} the generative factors common to both datasets from the salient ones, that produce the modified/added patterns of the target dataset. Otherwise, one may also assume that both datasets have salient patterns, but this is a lesser studied assumption. 
For instance, in medical imaging, using the first assumption, one may aim at finding the generative factors characterizing a pathology that is only present in a dataset of patients and not in healthy subjects. These factors need to be separated from the common ones that model the healthy patterns. Another example, related to the second assumption, may be two datasets comprising face images with different attributes/characteristics (e.g., one with glasses and one with smiles). The goal would be to separate the generative factors modeling these salient attributes from the ones characterizing the common face tracts.

Existing CA methods are based on simple generative models, like Variational AutoEncoders (VAEs) \cite{abid2018exploring,weinberger2022moment,louiset2024sepvae,louiset_separating_2024} and DCGANs \cite{carton2024double}, and are thus characterized by a very low-quality image generation and reconstruction, which limits interpretability and factor separation.
Motivated by this, we propose a novel framework that can be adapted to advanced GANs and DMs, leveraging their high-quality synthesis and structured representations to learn common and salient generative factors between two datasets. Unlike prior works about image editing, we can learn in a weakly-supervised way (only dataset label and no attribute labels). This allows for a new semantic edition where common and salient factors can be swapped between datasets without a controlled supervision. Compared to existing image editing techniques, our method can perform precise and per-image attribute manipulation without the need to search for the best editing strength or direction, and also supports linear subspace analysis (e.g., interpolation with PCA). As a new framework for CA, our results surpass those of all existing baselines in both latent separation and generation quality.

\noindent Our contributions are summarized as follows: 

\begin{itemize}
\item We introduce a new framework that can be adapted to both GANs and DMs for CA, enabling the identification and separation of common and salient generative factors between two datasets and a high-quality image synthesis.
\item We apply the proposed framework to two of the most
advanced and used generative models: StyleGAN and Denoising Diffusion Implicit Models (DDIMs), introducing reconstruction losses and adversarial regularizations that promote correct separation. For StyleGAN, we also present a refined framework that further enhances image detail, enabling both high-fidelity reconstruction and effective attribute swapping.
\item We provide extensive experimental validation on various datasets, including human faces and medical images. Our method outperforms recent CA baselines in latent separation and synthesis quality, while also surpassing most SOTA encoder-based approaches.
\end{itemize}

\section{Related Work}
Separating common and salient latent factors between two datasets is a recent and emerging problem related to several established research areas. In the following, we review the most relevant works and highlight how our approach differs in objectives, assumptions, and data setup. 

\paragraph{Image-to-Image Translation}  
Image-to-image translation (I2I) aims to map images from one domain to another while preserving structural consistency. Early methods such as Pix2Pix~\cite{isola2017image} are trained with paired data, whereas CycleGAN~\cite{zhu2017unpaired} introduced cycle consistency for unpaired translation.
Subsequent approaches, including MUNIT~\cite{huang2018multimodal} and DRIT~\cite{lee2018diverse}, disentangle content and style to enable greater control. In contrast, our goal is not translation but statistical analysis: we learn latent representations that capture common structures (e.g., faces) and distinctive variations (e.g., presence or absence of glasses), providing interpretable insight into the underlying generative factors.

\paragraph{Counterfactual Analysis}
Counterfactual analysis seeks to identify the minimal features or patterns in an image that alter the decision of a pre-trained classifier. Recent methods address this problem with adversarial~\cite{jeanneret2023adversarial}, diffusion-based~\cite{weng2024fast}, and text-to-image~\cite{jeanneret2024text} approaches to generate realistic counterfactuals, i.e., minimally modified yet visually plausible inputs. Most existing techniques depend on classifier feedback to produce counterfactuals. Differently, our method estimates common and salient generative factors in latent space, without relying on classifier-based explanation.

\paragraph{Anomaly Detection}
Our work is related to unsupervised anomaly detection, which typically models the background (normal) distribution and flags target-domain samples that deviate from it. Anomalies are often interpreted via reconstruction errors~\cite{guillon2021detection} or attention maps~\cite{venkataramanan2020attention}. However, these methods prioritize detection over analyzing the underlying generative latent factors.

\paragraph{Disentanglement}
A closely related line of work is disentanglement, which enables controlled modification of semantically meaningful attributes (e.g., a person’s age or gender) by altering a single latent component~\cite{kim2018disentangling}. Existing methods commonly impose additional constraints, such as inductive biases~\cite{higgins2017beta}, weak supervision~\cite{locatello2020weakly}, etc., to obtain meaningful factorization of latent spaces. Despite their effectiveness, these approaches learn factors within a single dataset and do not explicitly identify common/salient variations across datasets. 

\paragraph{Contrastive Analysis}
CA aims to separate target-specific latent variations from background ones. Our method follows this paradigm. 
Representative approaches include CA variational autoencoders (CA-VAEs)~\cite{weinberger2022moment,louiset2024sepvae}, which model the target data using both common and salient factors, whereas background data are explained solely by common factors. Although CA-VAEs can capture contrastive variations, their reliance on VAEs limits reconstruction quality and interpretability. To address these limitations, Double InfoGAN~\cite{carton2024double} introduced a GAN-based CA model.
However, it relies on DCGAN~\cite{chen2016infogan} and requires training a generator from scratch, limiting scalability to high-resolution images. In contrast, our method leverages powerful pre-trained StyleGAN and DMs, along with their semantic latent spaces, enabling high-resolution reconstruction and more effective separation. 

\paragraph{Controllable Image Manipulation}
Recent advances in GANs and DMs have motivated methods for controlling the generative process, allowing images to be modified in an interpretable manner. GANs, particularly \textbf{StyleGAN}~\cite{karras2019style}, have received extensive attention for high-quality image synthesis and a well-structured latent space that enables editing by manipulating latent codes. To apply such edits, an image is first encoded into the generator’s latent space (GAN inversion). Numerous inversion methods have been proposed (see~\cite{melnik2024face} for a review), with the most widely studied strategy based on training an encoder that maps real images into StyleGAN’s latent space in one or a few forward passes. Beyond inversion, many works discover semantic directions in latent space for attribute editing~\cite{shen2020interpreting,harkonen2020ganspace}. In parallel, DMs~\cite{ho2020denoising} have emerged as strong competitors to GANs in generative quality. Its deterministic variant, \textbf{DDIM}~\cite{song2021ddim}, supports near-perfect image reconstruction, making it well-suited for controllable generation. Unlike the structured latent spaces of StyleGAN, intermediate variables in DMs are high dimensional and less semantically organized, which makes direct manipulation more challenging. Diffusion Autoencoders (DiffAE)~\cite{preechakul2022diffusion} address this by training dedicated encoders that produce meaningful, decodable latent codes. Another line of work~\cite{Kwon2023SemanticLatent,jeong2024training} treats the U-Net bottleneck (h-space) as a semantic latent space and models it with an asymmetric reverse process (Asyrp). These efforts have enhanced semantic control within the latent space of DMs.

Our work builds upon faithful inversion and semantic latent spaces of advanced GANs and DMs, and make a step further since we encode images into two distinct latent spaces encoding the common and salient latent generative factors between two datasets. Furthermore, contrarily to the existing image-editing methods, we don't use a supervised approach by exploiting labels of given attributes, such as age or glasses. Instead, we rather employ a weakly-supervised approach, where the only supervision signal is the dataset label.

\section{Methods}
\label{sec:methods}
\noindent We present our framework using two of the most advanced generative models: StyleGAN and DDIM, instantiated as \textbf{CS-StyleGAN} and \textbf{CS-Diffusion}, respectively. We focus on the \textit{background/target} assumption, which is the most studied setting in the CA literature. \YL{An extension of CS-StyleGAN for the \textit{multiple-salient} assumption is presented in Sec.~\ref{sec:multi}.}

\subsection{Problem Statement}
\label{sec:problem}
Given real images from two datasets, $X = \{x_i\}_{i=1}^{N}$ (background) and $Y = \{y_j\}_{j=1}^{M}$ (target), where \(x_i\) lacks distinctive attributes (e.g., faces without glasses or healthy subjects) and \(y_j\) contains dataset-specific features (e.g., glasses or pathology), our goal is to learn two generative factors: a \textbf{common} factor \(c \in \mathbb{R}^{d_c}\) capturing patterns shared between $X$ and $Y$ (e.g., facial structure or healthy tissue), and a \textbf{salient} factor \(s \in \mathbb{R}^{d_s}\) encoding distinctive variations unique to $Y$ (e.g., the presence of glasses or pathology).

\begin{figure*}[t]
\centering
\includegraphics[width=1\linewidth]{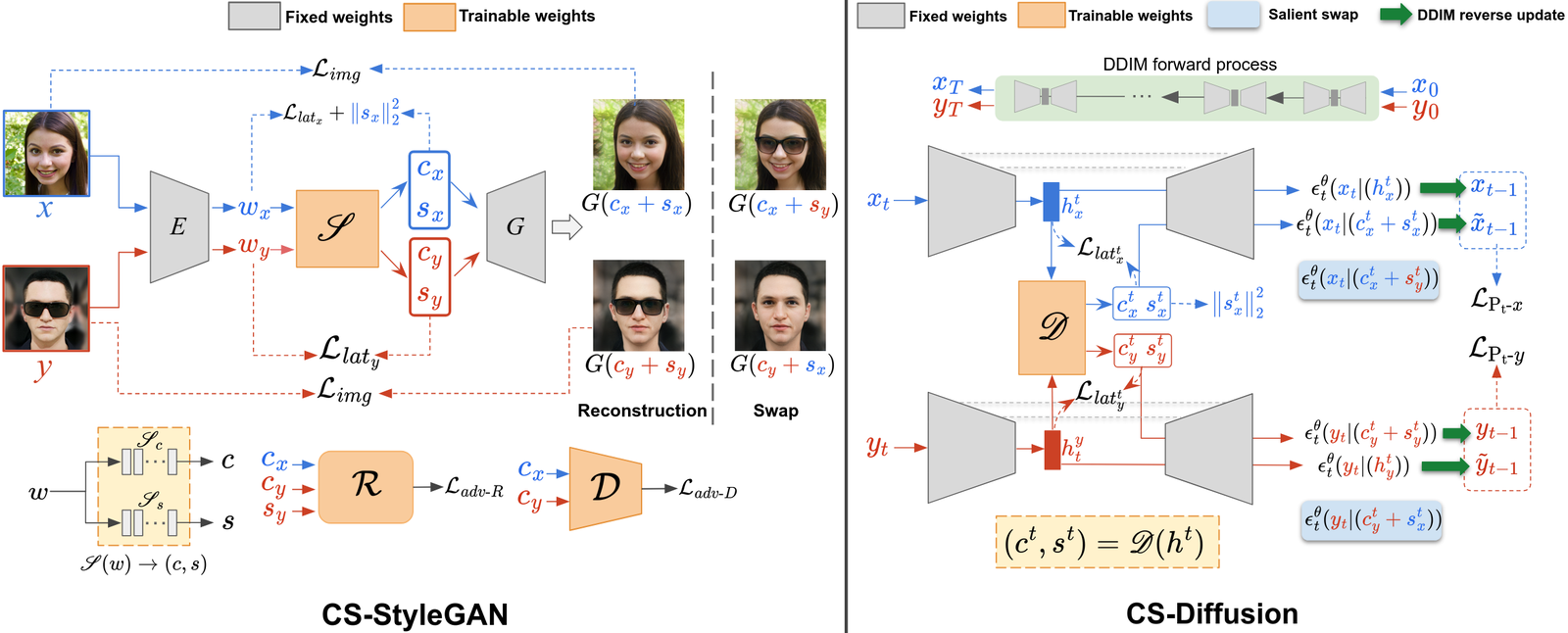}
\caption{
\textbf{Overview of the proposed frameworks adapted to StyleGAN and DMs.}
\textbf{Left: CS-StyleGAN}. Real images from both background ($X$, \textit{e.g.}, without glasses) and target ($Y$, \textit{e.g.}, with glasses) datasets are encoded into latent codes $w_x$ and $w_y$, respectively, by a pretrained encoder $E$. A separating network $\mathscr{S}$ decomposes each $w$ into a common factor $c$ (shared by $X$ and $Y$) and a salient factor $s$ (specific to $Y$). Training is \textit{weakly-supervised}, relying on dataset information only. Relevant separation between $c$ and $s$ is encouraged with $\mathcal{L}_{lat_x}$ and $\mathcal{L}_{lat_y}$ and by constraining $s_x$ to be zero with $\lVert s_x \lVert_2^2$, ensuring that $c_x$ fully encodes $X$.
Adversarial losses $\mathcal{L}_{adv\text{-}D}$ and $\mathcal{L}_{adv\text{-}R}$ align $c_x$ and $c_y$ and promote independence between $c$ and $s$. 
Image reconstruction loss $\mathcal{L}_{img}$ further ensures faithful reconstructions. \textbf{Right: CS-Diffusion.} Real images from two datasets are inverted to $x_T$ and $y_T$ via the DDIM forward process. At each reverse step, noisy latents $x_t$, $y_t$ are fed to a pretrained U-Net, extracting middle-layer features $h_x^t$, $h_y^t$. The separating network $\mathscr{D}$ maps these to $c$ and $s$ factors. Latent separation is encouraged by minimizing $\mathcal{L}_{{lat}^t_x}$, $\mathcal{L}_{{lat}^t_y}$, and $\lVert s^t_x \rVert_2^2$ during each DDIM reverse step, with faithful reconstruction enforced by minimizing $\mathcal{L}_{\mathrm{P_t}\text{-}x}$ and $\mathcal{L}_{\mathrm{P_t}\text{-}y}$.
During inference, for both methods, $c$ and $s$ from the same image are used for reconstruction, while attribute exchange is performed by swapping $s_x$ and $s_y$ before decoding.
}
\label{fig:architecture}
\end{figure*}

\subsection{Common and Salient (CS)-StyleGAN}
\label{sec:CS-StyleGAN}
For StyleGAN, we employ a pretrained encoder $E$ (e.g., pSp~\cite{richardson2021encoding}) to map real images into the latent space $w \in W$ (or a variant such as $W^+$ or $\mathcal{S}$). Let ${d_w}$ denote the dimensionality of the chosen space, so $w \in \mathbb{R}^{d_w}$. Given a background image $x$ and a target image $y$, their latent codes are denoted as $w_{x}$ and $w_{y}$, respectively. As shown in the left panel of 
Fig.~\ref{fig:architecture}, we introduce a \textbf{separating network} $\mathscr{S} \! = \! (\mathscr{S}_c,\mathscr{S}_s)$ that decomposes each $w$ into a \textbf{common} latent code $c\!=\!\mathscr{S}_c(w)$ and a \textbf{salient} latent code $s=\mathscr{S}_s(w)$. As typically done in CA \cite{weinberger2022moment,louiset2024sepvae}, we encourage the network's outputs to satisfy $w_x \!= \! c_x + s_x$ (with $s_x\! = \!s'$, typically $s' \!=\! 0$) for background images and $w_y\! = \!c_y + s_y$ for target images, so that $c_x$ fully represents the dataset $X$.\footnote{An ablation study about the (usual) choice $s_x=0$ can be found in Sec. III-C of the Supplementary (Suppl.)} 

\subsubsection{Latent Regularization}
\label{sec:regularization}
\YL{To promote faithful reconstruction of latent codes in the StyleGAN latent space, we define the following losses for optimizing $\mathscr{S}$}:
\begin{align}
    &\mathcal{L}_{lat_x} =\| \mathscr{S}_c(w_{x}) - w_{x} \|_2^2,
    \label{equ:loss_lat_x} \\
    &\mathcal{L}_{lat_y} = \| \mathscr{S}_c(w_{y}) + \mathscr{S}_s(w_{y}) - w_{y} \|_2^2,
    \label{equ:loss_lat_y}
\end{align}
Eq.~\eqref{equ:loss_lat_x} encourages $c_x=\mathscr{S}_c(w_{x})$ alone to reconstruct $w_x$, whereas Eq.~\eqref{equ:loss_lat_y} requires the joint contribution of the common and salient, $c_y$ and $s_y$, to reconstruct the target latents $w_y$. To ensure that $c_x$ fully encodes $X$,
we also minimize:
\begin{equation}
\mathcal{L}_{s_{x}} = \| \mathscr{S}_s(w_{x}) \|_2^2. 
\label{equ:loss_sx}
\end{equation}
The final latent reconstruction loss is:
\begin{equation}
\mathcal{L}_{lat} = \mathcal{L}_{lat_{x}} + \mathcal{L}_{lat_y} +\mathcal{L}_{s_{x}}. 
\label{equ:loss_lat}
\end{equation} 

The loss in Eq.~\eqref{equ:loss_lat} is necessary, but not enough. Indeed, it can not guarantee that: 1) common factors should encode the same information between $X$ and $Y$ (\textit{common consistency}) and 2) common and salient factors should be \textit{statistically independent} (\textit{i.e.,} they should not contain similar information). 

\noindent \textbf{\textit{Common consistency.}}
To force the common factors to encode the same information between $X$ and $Y$, we propose using an adversarial regularization loss. Specifically, a discriminator $\mathcal{D}$ is trained to distinguish whether a common factor $\mathscr{S}_c(w)$ comes from the background or target dataset. Then, the following adversarial objective is defined:
\begin{equation}
\begin{aligned}
    \mathcal{L}_{adv\text{-}D}
    = \min_{\mathscr{S}} \max_{\mathcal{D}} \big(
    &- \mathbb{E}_{w_y}[\log \mathcal{D}(\mathscr{S}_c(w_y))] \\
    &- \mathbb{E}_{w_x}[\log (1 - \mathcal{D}(\mathscr{S}_c(w_x)))]
    \big).
    \label{equ:loss_adv_D}
\end{aligned}
\end{equation}
\YL{where $\mathcal{D}$ is trained to classify whether a given common factor originates from $X$ or $Y$, while $\mathscr{S}$ is trained adversarially to fool $\mathcal{D}$, encouraging the distributions of $\mathscr{S}_c(w_x)$ and $\mathscr{S}_c(w_y)$ to become indistinguishable.}

\noindent \textbf{\textit{Common/salient statistical independence.}}
To encourage independence between common and salient factors, we introduce a regressor network $\mathcal{R}$ trained to predict the salient vectors $s_y$ and $s_x$ from the common ones $c_y$ and $c_x$. If $\mathcal{R}$ can successfully predict $s$ from $c$, it indicates that $c$ still contains information about $s$, and the separation is incomplete. We adversarially train $\mathscr{S}$ and $\mathcal{R}$ with the following loss:
\begin{equation}
\begin{aligned}
    \mathcal{L}_{adv\text{-}R}
    = \min_{\mathscr{S}} \max_{\mathcal{R}} \big(
        &- \mathbb{E}_{w_y}[\| \mathcal{R}(\mathscr{S}_c(w_y)) - \mathscr{S}_s(w_y) \|_2^2 ] \\
        &- \mathbb{E}_{w_x}[\| \mathcal{R}(\mathscr{S}_c(w_x)) \|_2^2 ] 
    \big),
    \label{equ:loss_adv_R}
\end{aligned}
\end{equation}
Minimizing this loss ensures that $c$ does not contain information that could be used to reconstruct $s$. We use $c$ as input and $s$ as output because the common factor typically carries more information than the salient one, making it important to prevent $c$ from carrying salient details. This regularization, together with the reconstruction and adversarial losses \YL{(Eqs.~\eqref{equ:loss_lat} and~\eqref{equ:loss_adv_D})}, forces $s_y$ to encode \YL{exclusively} the salient information of $Y$. \PG{A comparison with mutual-information based regularization can be found in Sec. III-E of the Suppl., where we show that the proposed regressor-based
regularization achieves the best results both for image quality and latent separation.}

\subsubsection{Image-Space Losses}
As is common in the literature, we also employ losses 
\YL{to enforce faithful reconstructions} in the image (pixel) space. Given a background image $x$ and a target image $y$, we map these to $c_{x}$ (resp. $s_x$) and $c_y$ (resp. $s_y$) using the separating network $\mathscr{S}$, and reconstruct the images with the pretrained StyleGAN generator $G$. For simplicity, we express image-space losses as functions of a real image $a$ and its reconstructed image $b$. First, we include the standard pixel-wise $\mathcal{L}_2$ loss: 
\begin{equation}
    \mathcal{L}_2(a, b) = \|a - b\|_2^2 \label{equ:l2_loss}.
\end{equation}
Second, we use the LPIPS loss~\cite{zhang2018unreasonable} to measure perceptual differences between $a$ and $b$:
\begin{equation}
    \mathcal{L}_{\text{lpips}}(a, b) = \|A(a) - A(b)\|_2^2,
    \label{equ:lpips_loss}
\end{equation}
where $A(\cdot)$ is a pretrained AlexNet used to extract perceptual features. Additionally, for face images, we calculate identity loss as the cosine similarity between two samples:
\begin{equation}
    \mathcal{L}_{\text{id}}(a, b) = 1 - \langle I(a), I(b) \rangle,
    \label{equ:id_loss}
\end{equation}
where $I(\cdot)$ is the pretrained ArcFace~\cite{deng2019arcface} network. Combining Eqs.~\eqref{equ:l2_loss}, \eqref{equ:lpips_loss}, and \eqref{equ:id_loss}, we obtain: 
\begin{equation}
    \mathcal{L}_{\mathrm{rec}}(a, b) = \mathcal{L}_2(a, b) + \lambda_{lpips}  \mathcal{L}_{\text{lpips}}(a, b) + \lambda_{id} \mathcal{L}_{\text{id}}(a, b),
    \label{equ:img_rec}
\end{equation}
where $\lambda_{lpips}$ and $\lambda_{id}$ are user-defined loss weights.
We applied Eq.~\eqref{equ:img_rec} to both $x$ and $y$ reconstructions, giving the total image-space loss:
\begin{equation}
    \mathcal{L}_{img} = \mathcal{L}_{\mathrm{rec}}(x, G(c_{x})) + \mathcal{L}_{\mathrm{rec}}(y, G(c_{y} + s_{y})).
    \label{equ:training_img_loss}
\end{equation}

The final loss function used to train the network $\mathscr{S}$ is:
\begin{equation}
    \mathcal{L_\text{sep}} = \lambda_{1}\mathcal{L}_{img} + \lambda_{2}\mathcal{L}_{lat} + \lambda_{3}\mathcal{L}_{adv\text{-}R} + \lambda_{4}\mathcal{L}_{adv\text{-}D},
    \label{equ:sep}
\end{equation}
where $\lambda_{1}, \lambda_{2}, \lambda_{3}, \lambda_{4}$ are user-defined hyperparameters. 

\begin{figure*}[ht!]
\begin{center}
   \includegraphics[width=0.7\linewidth]{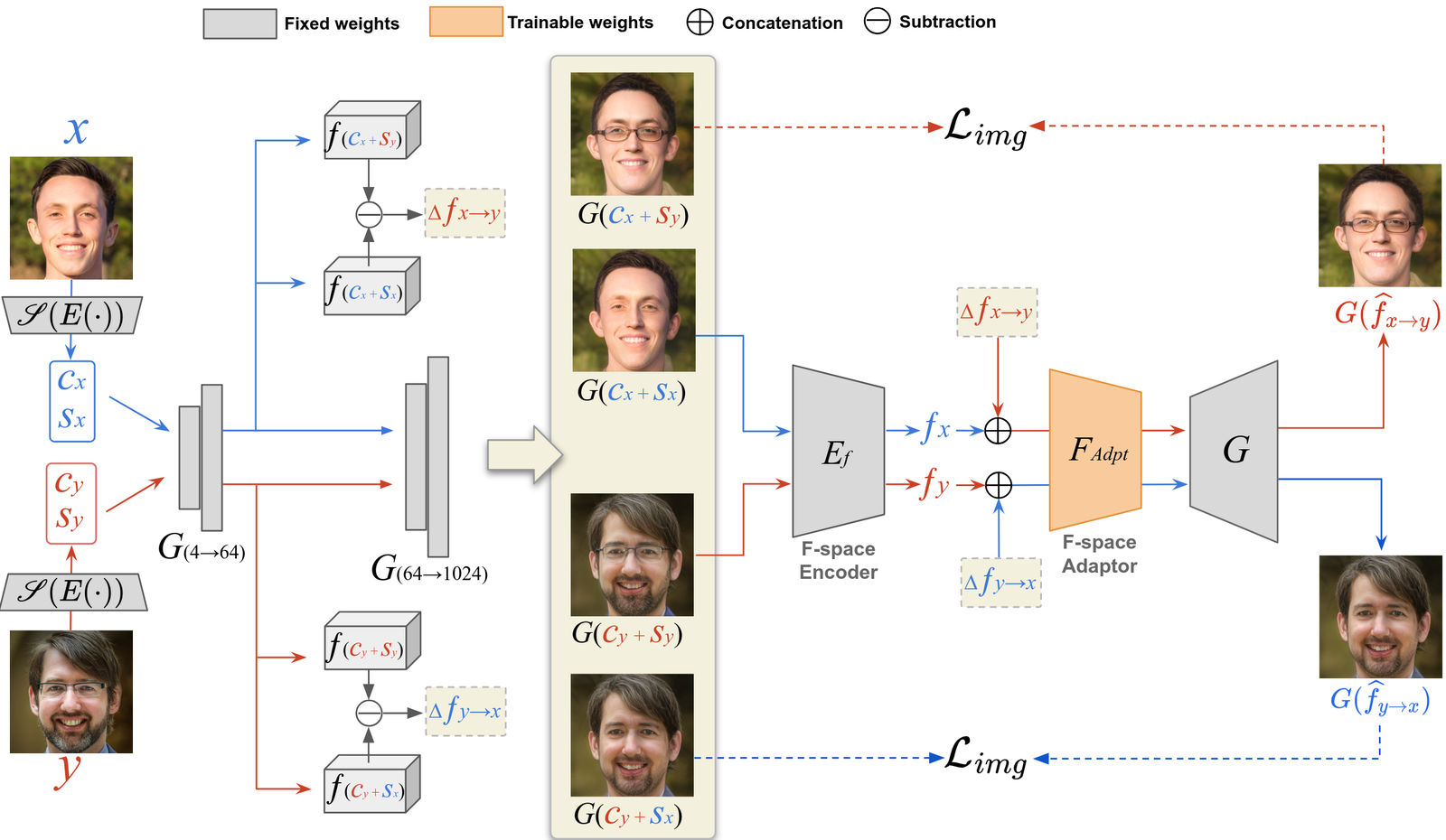}
    \caption{\textbf{Overview of the F-space refinement with pre-trained CS-StyleGAN.} 
    Real images from \( X \) and \( Y \) are encoded into latent codes \( (c_x, s_x) \) and \( (c_y, s_y) \) using a pre-trained encoder-separator network \( \mathscr{S}(E(\cdot)) \). These codes are then fed into the fixed StyleGAN2 generator \( G \) to produce reconstructions \( G(c_x + s_x) \), \( G(c_y + s_y) \), and swapped images \( G(c_x + s_y) \), \( G(c_y + s_x) \). Semantic feature shifts \( \Delta f_{x \rightarrow y} \) and \( \Delta f_{y \rightarrow x} \) are computed by subtracting the generator’s intermediate feature maps. A frozen F-space encoder \( E_f \) maps \( G(c_x+s_x) \) and \( G(c_y+s_y) \) to features \( f_x \) and \( f_y \), respectively. These \( f \) are concatenated with feature shifts \( \Delta f \) and passed through a trainable adapter \( F_{\text{Adpt}} \). The adapted features are then decoded by \( G \) to produce swapped images: \( G(\widehat{f}_{x \rightarrow y}) \) and \( G(\widehat{f}_{y \rightarrow x}) \). The adapter is trained using a reconstruction loss \( \mathcal{L}_{x \rightarrow y} \) that compares \( G(c_x+s_y) \) and \( G(\widehat{f}_{x \rightarrow y}) \), with a symmetric counterpart \( \mathcal{L}_{y \rightarrow x} \) defined analogously. To further enhance reconstruction fidelity, real images from \( X \) and \( Y \) are processed through \( E_f \), \( F_{\text{Adpt}} \), and \( G \) to compute the loss \( \mathcal{L}_r \). Additionally, adversarial losses $\mathcal{L}_{\text{adv-}x}$ and $ \mathcal{L}_{\text{adv-}y}$ are also minimized, to encourage $G(\widehat{f}_{y \rightarrow x})$ and $G(\widehat{f}_{x \rightarrow y})$ to resemble $x$ and $y$, respectively.
    }
\label{fig:refine_framework}
\end{center}
\end{figure*}

\subsubsection{F-space Refinement} 
\label{sec:f_refine}
Once $\mathscr{S}$ is trained, we encode $x$ and $y$ as $(c_x,s_x)$ and $(c_y,s_y)$ using $E$ and $\mathscr{S}$. The generator $G$ reconstructs $x$ and $y$ from $c_x+s_x$ and $c_y+s_y$, and swaps salient attributes by feeding $c_x+s_y$ and $c_y+s_x$ into $G$ (see Fig.~\ref{fig:architecture}, left). However, manipulating images in the low-dimensional space (e.g., $W^+$) often limits reconstruction quality. \YL{Recent work~\cite{bobkov2024devil} introduced a network that maps semantic shifts from the $W^+$ space to the intermediate StyleGAN feature space ($F$-space), improving detail preservation in image editing (see Fig. 21 in the Suppl.). We adapt and extend this idea as a refinement stage to improve CA results (see Fig.~\ref{fig:refine_framework}). Specifically, we extract intermediate generator features $f$ by feeding the learned common and salient factors into the StyleGAN2 generator $G$ (e.g., layer-9 output), denoted $f(c+s) \in \mathbb{R}^{512 \times 64 \times 64}$. To isolate the effect of salient swapping between images $x$ and $y$, we define shifts in the feature space $F$ for the two swap directions:}
\begin{align}
\Delta f_{x \rightarrow y} = f(c_x + s_x) - f(c_x + s_y), \label{equ:delta_x2y} \\
\Delta f_{y \rightarrow x} = f(c_y + s_y) - f(c_y + s_x). \label{equ:delta_y2x}
\end{align}
\YL{Here, $\Delta f_{x \to y}$ quantifies the feature differences induced by replacing the salient pattern of $x$ with that of $y$ (at fixed $c_x$), while $\Delta f_{y \to x}$ quantifies the converse differences (at fixed $c_y$).
Meanwhile, the features of the reconstructions $G(c_x + s_x)$ and $G(c_y + s_y)$ can be extracted using a pretrained encoder $E_f$: \(f_x = E_f(G(c_x + s_x))\) and \(f_y = E_f(G(c_y + s_y))\).
In line with~\cite{bobkov2024devil}, we condition the shifts on the features
by concatenating them and feeding the result into an adapter $F_{\text{Adpt}}$: 
$\widehat{f}_{x \rightarrow y} = F_{\text{Adpt}}(f_x \oplus \Delta f_{x \rightarrow y})$
and $\widehat{f}_{y \rightarrow x} = F_{\text{Adpt}}(f_y \oplus \Delta f_{y \rightarrow x})$, where $\oplus$ denotes concatenation.}
\YL{The adapter is trained such that decoding the adapted features with $G$ yields the edited images, by minimizing:}
\begin{align}
\mathcal{L}_{x \rightarrow y} &= \mathcal{L}_{\mathrm{rec}}(G(c_x + s_y), \; G(\widehat{f}_{x \rightarrow y})), \label{eq:loss_xy} \\
\mathcal{L}_{y \rightarrow x} &= \mathcal{L}_{\mathrm{rec}}(G(c_y + s_x), \; G(\widehat{f}_{y \rightarrow x})),
\label{eq:loss_yx}
\end{align} 
Since Eqs.~\eqref{eq:loss_xy} and~\eqref{eq:loss_yx} use only synthetic images and may harm real-image performance, we also train on real images:
\begin{align}
\mathcal{L}_{\text{r}} = \mathcal{L}_{\mathrm{rec}}(x,\; \widehat{x}) + \mathcal{L}_{\mathrm{rec}}(y,\; \widehat{y}),
\end{align}
where $\widehat{x}\!=\!G(F_\text{Adpt}(E_f(x)))$, $\widehat{y}\!=\!G(F_\text{Adpt}(E_f(y)))$. To encourage realism in the swapped images, we additionally incorporate an adversarial loss: $\mathcal{L}_{\text{adv}}=\mathcal{L}_{\text{adv-}x} + \mathcal{L}_{\text{adv-}y}$, where $\mathcal{L}_{\text{adv-}x}$ is applied to $G(\widehat{f}_{y \rightarrow x})$ against $x$, and $\mathcal{L}_{\text{adv-}y}$ to $G(\widehat{f}_{x \rightarrow y})$ against $y$. The overall refinement objective is:
\begin{align}
\mathcal{L}_{\text{refine}} = \mathcal{L}_{x \rightarrow y} + \mathcal{L}_{y \rightarrow x} + \mathcal{L}_{\text{r}} + \lambda_{adv} \mathcal{L}_{adv}.
\end{align}

\subsubsection{Multiple-Salient Assumption}
\label{sec:multi}
\begin{figure*}[ht!]
\begin{center}
   \includegraphics[width=0.8\linewidth]{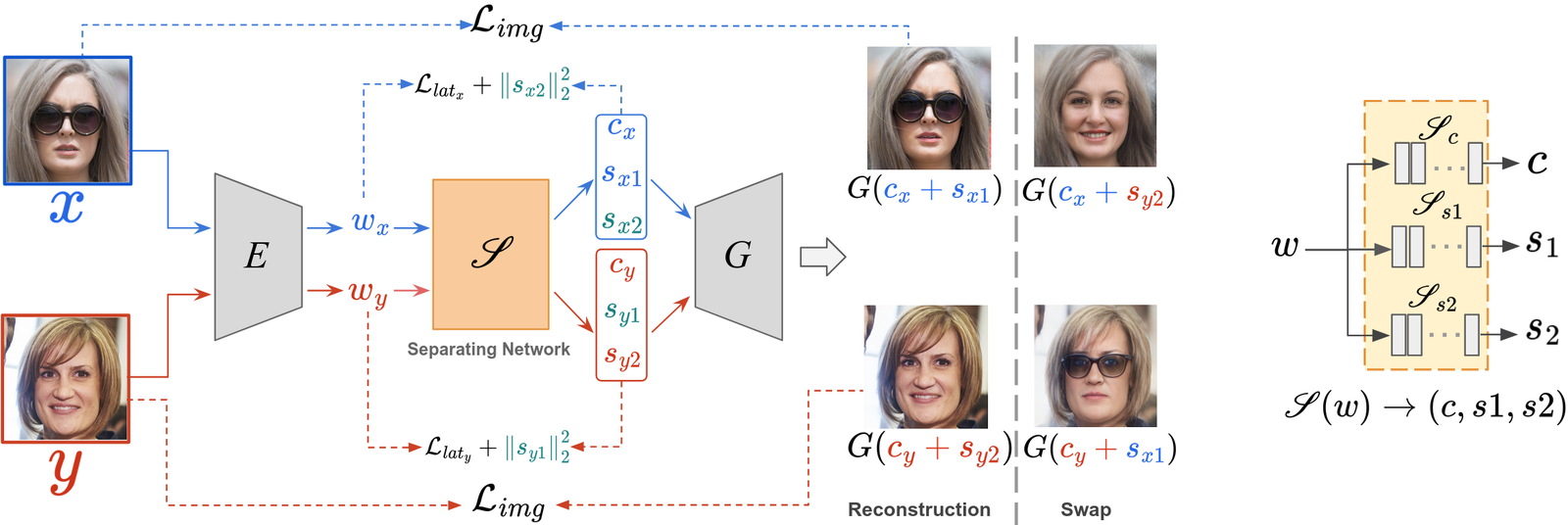}
\end{center}
\caption{\textbf{Overview of the framework for the \textit{multiple-salient} assumption}. Real images from both $X$ (e.g., with glasses only) and $Y$ (e.g., with smile only) datasets are encoded into latent codes $w_x$ and $w_y$ by a pretrained encoder $E$. A separating network $\mathscr{S}$ decomposes each $w$ into a common factor $c$ (shared by $X$ and $Y$) and two salient factors: $s_1$ (encoding attributes unique to $X$) and $s_2$ (encoding attributes unique to $Y$). Relevant separation between $c$ and $s$ is encouraged by minimizing $\mathcal{L}_{lat_x}$ and $\mathcal{L}_{lat_y}$ (Eqs.~\eqref{equ:s1s2_latx_loss} and \eqref{equ:s1s2_laty_loss}) and by minimizing $\lVert s_{x2} \rVert_2^2$ and $\lVert s_{y1} \rVert_2^2$ (i.e., both $s_{x2}$ and $s_{y1}$ should be equal to 0). The image reconstruction loss $\mathcal{L}_{img}$ (Eq.~\eqref{equ:s1s2_img_loss}) further ensures faithful reconstructions. 
} 
\label{fig:arch_two_salients}
\end{figure*}

As shown in Fig.~\ref{fig:arch_two_salients}, we also introduce a new CS-StyleGAN adapted for the \textbf{\textit{multiple-salient}} case, in which the separating network $\mathscr{S}$ is designed to have three outputs: $\mathscr{S}(w) \rightarrow (c, s_1, s_2)$. We train one output, $s_{1}$, to encode only the distinctive patterns in the $X$ dataset, and the other, $s_2$, to encode those in the $Y$ dataset. We train $\mathscr{S}$ by  minimizing the following latent losses:
\begin{align}
    \mathcal{L}_{lat_x} &= \| \mathscr{S}_c(w_{x}) + \mathscr{S}_{s1}(w_{x}) - w_{x} \|_2^2, \label{equ:s1s2_latx_loss} \\
    \mathcal{L}_{lat_y} &= \| \mathscr{S}_c(w_{y}) + \mathscr{S}_{s2}(w_{y}) - w_{y} \|_2^2. \label{equ:s1s2_laty_loss}
\end{align}
The image-space loss is defined as:
\begin{align}
    \mathcal{L}_{img} = \mathcal{L}_{\mathrm{rec}}\left(x, G(c_{x}+s_{x1})) + \mathcal{L}_{\mathrm{rec}}(y, G(c_{y} + s_{y2})\right),
    \label{equ:s1s2_img_loss}
\end{align}
where $\mathcal{L}_{\text{rec}}(a, b)$ is defined in Eq.~\eqref{equ:img_rec}.
We also minimize $\|s_{x2}\|^2_2$ to ensure that the distinctive features of $X$ are completely encoded by $s_{x1}$, and minimize $\|s_{y1}\|^2_2$ to ensure that the distinctive features of $Y$ are fully captured by $s_{y2}$. 

\subsection{CS-Diffusion}
DDIM is a variant of DMs that enables deterministic and efficient sampling. It serves as the core sampling method \PG{for two state-of-the-art methods that learn a semantic latent space of DMs:} Asyrp \cite{Kwon2023SemanticLatent} and DiffAE~\cite{preechakul2022diffusion}.

\subsubsection{Common and Salient in h-space} 
Most DMs use a U-Net as denoising neural network that predicts either noise, denoised data, or other intermediary targets. In Asyrp, the bottleneck of the U-Net (i.e., the h-space) is used as a controllable semantic latent space. \YL{We adopt the DDIM reverse process of Asyrp and express it for our CA setting as follows:}
\begin{align}
x_{t-1} &= \sqrt{\alpha_{t-1}}\, \mathbf{P}_t\left(\epsilon_t^{\theta}(x_t)\right) + \mathbf{D}_t\left(\epsilon_t^{\theta}(x_t)\right) + \sigma_t z_t^x, \label{equ:x_t_next}\\
y_{t-1} &= \sqrt{\alpha_{t-1}}\, \mathbf{P}_t\left(\epsilon_t^{\theta}(y_t)\right) + \mathbf{D}_t\left(\epsilon_t^{\theta}(y_t)\right) + \sigma_t z_t^y, \label{equ:y_t_next}
\end{align}
where $x_t$ and $y_t$ are the noisy versions of $x$ (background) and $y$ (target) at step $t$ of the reverse process; \YL{$\mathbf{P}_t(\epsilon_t^{\theta}(x_t))$ denotes the predicted clean, noise-free sample $x_0$ from $x_t$;} $\mathbf{D}_t(\epsilon_t^{\theta}(x_t))$ adjusts $x_t$ in the transition to $x_{t-1}$. \YL{The same definitions apply analogously for \(y_t\).} Following prior work, we omit the stochastic terms $\sigma_t z_t^x$ and $\sigma_t z_t^y$ for editing tasks.

Let $h_x^t$ and $h_y^t$ denote the $h$-space outputs for $x_t$ and $y_t$. As shown in Fig.~\ref{fig:architecture} (right), we introduce a separating network $\mathscr{D}$ that decomposes $h^t$ at each $t$ into common and salient factors: $c_x^t = \mathscr{D}_c(h^t_{x})$, $s_x^t = \mathscr{D}_s(h^t_{x})$, $c_y^t = \mathscr{D}_c(h^t_{y})$, and $s_y^t = \mathscr{D}_s(h^t_{y})$. As before, our goal is to train $\mathscr{D}$ such that, at each $t$, $h^t_x = c_x^t + s_x^t$ (typically $s_x^t = 0$ so that $c_x^t$ fully represents the background dataset $X$), and $h^t_y = c_y^t + s_y^t$ for target images.
We minimize the following losses: 
\begin{align}
    \mathcal{L}_{lat_t} = \mathcal{L}_{lat^t_x} + \left\| s_x^t \right\|^2_2 + \mathcal{L}_{lat^t_y},
\end{align}
where $\mathcal{L}_{lat^t_x} = \| c_x^t - h^t_x \|_2^2$ and $\mathcal{L}_{lat^t_y} = \| (c_y^t + s_y^t) - h^t_y \|_2^2$.
As observed by Asyrp and our own experiments, modifying the h-space in both $P_t$ and $D_t$ brings negligible change to the image. Thus, at each timestep, we omit $D_t(\cdot)$ and compute image-space losses on $P_t(\cdot)$ rather than on $x_{t-1}$ (or $y_{t-1}$):
\begin{align}
\mathcal{L}_{\mathrm{P_t}\text{-}x} &= \mathcal{L}_{\mathrm{rec}}\left(\mathbf{P}_t(\epsilon_t^{\theta}(x_t | h_x^t)), \; \mathbf{P}_t(\epsilon_t^{\theta}(x_t | c_x^t))\right), \label{equ:loss_img_x_diff} \\
\mathcal{L}_{\mathrm{P_t}\text{-}y} &= \mathcal{L}_{\mathrm{rec}}\left(\mathbf{P}_t(\epsilon_t^{\theta}(y_t  | h_y^t)),\; \mathbf{P}_t(\epsilon_t^{\theta}(y_t  | (c_y^t+s_y^t)))\right),
\end{align}
where $\epsilon_t^{\theta}(z_t | h_z^t) = \epsilon_t^{\theta}(z_t)$, $\forall z \in \{x, y\}$, as $h_z^t$ is the original U-Net bottleneck output. The final loss to train $\mathscr{D}$ is:
\begin{align}
\mathcal{L}_{\text{h-space}} =\mathcal{L}_{lat_t} + \mathcal{L}_{\mathrm{P_t}\text{-}x} + \mathcal{L}_{\mathrm{P_t}\text{-}y}.
\label{equ:h-space}
\end{align}

\subsubsection{Common and Salient in DiffAE}
DiffAE introduces a semantic encoder $E_{\mathrm{sem}}$ that maps the input image $x$ into a semantic representation $E_{\mathrm{sem}}(x) \rightarrow z_{\mathrm{sem}} $, and a stochastic encoder $E_{\mathrm{sto}}$ that maps $x$ and $z_{\mathrm{sem}}$ to a noise-injected image $x_T$ via a conditional forward process: $E_{\mathrm{sto}}(x, z_{\mathrm{sem}}) \rightarrow x_T$. The image is then conditionally generated with the DDIM decoder: $Dec(z_{\mathrm{sem}}, x_T)$. We use the same architecture for $\mathscr{S}$ as is used for StyleGAN’s $W$ space ($w \in \mathbb{R}^{512}$). We first use the pretrained $E_{\mathrm{sem}}$ to encode $x$ and $y$ as $z_{\mathrm{sem}\text{-}x}$ and $z_{\mathrm{sem}\text{-}y}$, and $E_{\mathrm{sto}}$ to generate their noisy versions $x_T$ and $y_T$. These semantic codes are mapped to $c_x$, $s_x$, $c_y$, and $s_y$ via $\mathscr{S}$. We then minimize losses in Eqs.~\eqref{equ:loss_lat_x}--\eqref{equ:loss_sx}, with $z_{\mathrm{sem}\text{-}x}$ and $z_{\mathrm{sem}\text{-}y}$ replacing $w_x$ and $w_y$. Additionally, we minimize the image-space losses defined as:
{\small
\begin{equation}
\mathcal{L}_{\text{img}}=\mathcal{L}_{\mathrm{rec}}(x, Dec(c_x, x_T)) 
+ \mathcal{L}_{\mathrm{rec}}(y, Dec(c_y + s_y, y_T)).
\label{equ:img_loss_diffae}
\end{equation}
}

\section{Experiments}
\label{sec:experiments}
In the following, we present the datasets, comparison baselines, evaluation methods, and results. Implementation/architecture details, ablation studies, and additional results are provided in the Suppl. Material. Our code is available at \url{https://github.com/yunlongH/CA-with-stylegan2-pSp/}.
\begin{table}[!ht]
    \caption{Summary of Attributes and $X$/$Y$ Datasets for FFHQ and CelebA-HQ. Train/Test counts are given per dataset (e.g., ``10k'' indicates 10k for each of $X$ and $Y$). {\bfseries Case 1 (multiple attributes)}: Y has glasses and smile, absent in X. {\bfseries Case 2 (multiple salients)}: glasses only in X; smile only in Y.}
    \label{tab:n_training}
    \centering
    \footnotesize
    \resizebox{1.0\columnwidth}{!}{
    \begin{tabular}{llllll}
        \toprule
        \textbf{Dataset} & \textbf{Attribute} & \textbf{$X$} & \textbf{$Y$} & \textbf{Train} & \textbf{Test} \\
        \midrule
        \multirow{7}{*}{\textbf{FFHQ}}
          & Glasses          & No Glasses                 & Glasses                   & 10k   & 3.5k \\
          & Head pose             & Frontal                    & Turned Left/Right         & 8k    & 1.6k \\
          & Gender           & Male                       & Female                    & 18.5k & 1.7k \\
          & Age              & Young                      & Old                       & 14.7k & 1.7k \\
          & Smile            & Non-Smiling                & Smiling                   & 16.9k & 4.2k \\
          & Glasses \& Smile (case 1)  & No Glasses \& Non-Smiling  & Glasses \& Smiling        & 9.6k  & 2.5k \\
          & Glasses \& Smile (case 2)  & Glasses \& Non-Smiling     & No Glasses \& Smiling     & 3k    & 1k   \\
        \midrule
        \multirow{2}{*}{\textbf{CelebA-HQ}}
          & Smile            & Non-Smiling                & Smiling                   & 11k   & 2.5k \\
          & Gender           & Male                       & Female                    & 9k    & 2k   \\
        \bottomrule
    \end{tabular}
    }
\end{table}
\subsection{Datasets} 
Experiments are conducted on 4 datasets: FFHQ \cite{karras2019style}, CelebA‐HQ \cite{karras2018progressive} (human faces), 
AFHQv2 \cite{choi2020starganv2} (cats and dogs), 
and BraTS2023 \cite{menze2015brats} (brain MRI). 
For each dataset, we split the data into $X$ and $Y$ datasets using one or two attributes (e.g., “No Glasses” vs. “Glasses”). We balance the splits so that $X$ and $Y$ contain the same number of images in both the training and test sets. Table~\ref{tab:n_training} summarizes the attributes and the per-split train/test counts for FFHQ and CelebA-HQ. Since the official FFHQ release does not include attribute labels, we use a public annotation set.\footnote{\url{https://github.com/DCGM/ffhq-features-dataset}} For BraTS2023, we use 8{,}000 healthy ($X$) and 8{,}000 tumored ($Y$) images for training, and 2{,}000 per class for testing. For AFHQv2, we use 4{,}500 cat ($X$) and 4{,}500 dog ($Y$) images for training, and 500 per class for testing.


\subsection{Comparison Baselines}
We compare our method with prior CA baselines, including MM-cVAE~\cite{weinberger2022moment}, SepVAE~\cite{louiset2024sepvae}, and Double InfoGAN~\cite{carton2024double}. 
We further evaluate our method by comparing it with recent StyleGAN-based encoders on image reconstruction and editing, including pSp~\cite{richardson2021encoding}, e4e~\cite{tov2021designing}, HyperStyle~\cite{alaluf2022hyperstyle}, ReStyle~\cite{alaluf2021restyle}, IDInvert~\cite{zhu2020indomain}, StyleSpace~\cite{wu2021stylespace}, StyleGAN3-Editing~\cite{alaluf2022third}, Feature-Style~\cite{yao2022featurestyle}, StyleRes~\cite{pehlivan2023styleres}, and SFE~\cite{bobkov2024devil}. For attribute editing, we use semantic directions learned with InterFaceGAN~\cite{shen2020interpreting} and StyleSpace~\cite{wu2021stylespace}. We also include diffusion-model baselines: DiffAE~\cite{preechakul2022diffusion}, Asyrp ($h$-space)~\cite{Kwon2023SemanticLatent}, and TIME~\cite{jeanneret2024text} (a text-to-image model designed for counterfactual analysis). All methods are evaluated on the same test set for fairness.

\subsection{Evaluation Methods}
To evaluate our method, we consider two aspects: (1) reconstruction and edited-image quality, and (2) latent separation quality. 
Image quality is assessed both qualitatively and quantitatively using LPIPS~\cite{zhang2018unreasonable}, L2, MS-SSIM~\cite{radford2021learning}, identity similarity (ID-Sim; via ArcFace feature distance; see Eq.~\eqref{equ:id_loss}), and FID~\cite{heusel2017gans}. Furthermore, to evaluate the performance of image editing when using  FFHQ and CelebA-HQ, we compute identity similarity between a real image and its edited version of the same subject. Realism is assessed using FID with respect to real images from $X$ (FID-X) or $Y$ (FID-Y). In addition, we use a classifier trained on real images to evaluate edited outputs, with target labels set to the desired post-edit classes. To evaluate latent separation, we train a classifier on the learned latent factors to predict attribute presence. In this way, we can determine whether attribute information is encoded in the common or salient latent space.

\begin{figure*}[ht!]
    \centering
    \includegraphics[width=1.0\linewidth]{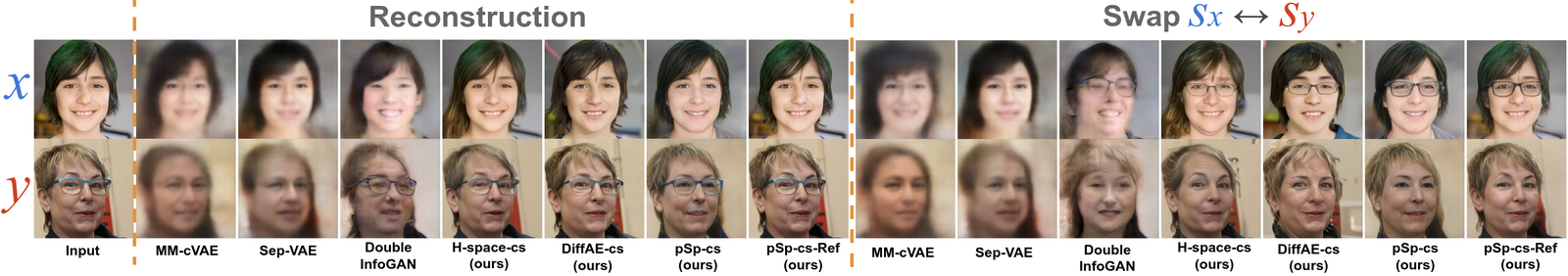}
    \caption{Qualitative comparison with CA baselines. The two datasets contain people without ($X$, background) and with ($Y$, target) glasses. \textbf{pSp-cs} is the base CS–StyleGAN model with a pSp encoder, and \textbf{pSp-cs-Ref} applies F-space refinement to its outputs (common and salient factors).}
    \label{fig:ca_baselines}
\end{figure*}
\begin{table*}[!ht]
\caption{Quantitative Comparison of Reconstruction and Attribute Swapping. Best results in bold, second-best underlined.}
\label{tab:ca_quality_comp}
\centering
\footnotesize
\resizebox{\textwidth}{!}{
\begin{tabular}{lcccccccccccccc}
\toprule
\multirow{3}{*}{\textbf{Model}} & 
\multicolumn{5}{c}{\textbf{Reconstruction quality ($X$ dataset)}} & 
\multicolumn{5}{c}{\textbf{Reconstruction quality ($Y$ dataset)}} & 
\multicolumn{2}{c}{\textbf{Swap $X \rightarrow Y$ (+Glasses)}} & 
\multicolumn{2}{c}{\textbf{Swap $Y\rightarrow X$ (-Glasses)}} \\
\cmidrule(lr){2-6} \cmidrule(lr){7-11} \cmidrule(lr){12-13} \cmidrule(lr){14-15}
& LPIPS$\downarrow$ & MSE$\downarrow$ & MS-SSIM$\uparrow$ & ID-Sim$\uparrow$ & FID-X$\downarrow$ 
& LPIPS$\downarrow$ & MSE$\downarrow$ & MS-SSIM$\uparrow$ & ID-Sim$\uparrow$ & FID-Y$\downarrow$
& ID-Sim$\uparrow$ & FID-Y$\downarrow$ 
& ID-Sim$\uparrow$ & FID-X$\downarrow$ \\
\midrule
MMc-VAE          & 0.600 & 0.018 & 0.584 & 0.119 & 172.498  & 0.608 & 0.018 & 0.583 & 0.151 & 158.550 & 0.085 & 166.042 & 0.077 & 190.874 \\
Sep-VAE          & 0.556 & 0.015 & 0.633 & 0.159 & 145.980  & 0.573 & 0.016 & 0.619 & 0.169 & 142.732 & 0.142 & 160.309 & 0.140 & 148.384 \\
Double InfoGAN   & 0.369 & 0.028 & 0.604 & 0.277 & 72.892   & 0.361 & 0.023 & 0.604 & 0.248 & 57.715  & 0.158 & 58.891  & 0.153 & 76.836  \\
\midrule
DiffAE-cs       & 0.188 & 0.011 & 0.831 & 0.732 & \underline{29.816}   & \underline{0.118} & 0.006 & 0.906 & \underline{0.877} & \underline{16.018}  & 0.570 & 39.050  & 0.540 & \underline{38.598}  \\
H-space-cs      & \underline{0.179} & \underline{0.004} & \underline{0.908} & \underline{0.838} & 30.776   & 0.190 & \underline{0.004} & \underline{0.907} & 0.840 & 29.271  & \underline{0.604} & \underline{36.736}  & \underline{0.624} & 39.486  \\
pSp-cs & 0.190 & 0.015 & 0.752 & 0.756 & 41.709   & 0.183 & 0.014 & 0.740 & 0.790 & 33.095  & 0.601 & 46.696  & 0.650 & 60.309  \\
pSp-cs-Ref  & \textbf{0.019} & \textbf{0.001} & \textbf{0.983} & \textbf{0.992} & \textbf{2.406} 
                & \textbf{0.020} & \textbf{0.001} & \textbf{0.982} & \textbf{0.991} & \textbf{2.201} 
                & \textbf{0.804} & \textbf{23.302} & \textbf{0.809} & \textbf{25.222} \\
\bottomrule
\end{tabular}
}
\end{table*}
\begin{table*}[t!]
\caption{Results of common and salient separation. 5-fold average classification accuracy and standard deviation (std) on the FFHQ dataset. \YL{The training datasets $X$ and $Y$ contain facial images without and with glasses, respectively.}
For glasses: $\Delta = |0.5 - C| + |1.0 - S|$; for other attributes: $\Delta = |1.0 - C| + |0.5 - S|$. Best results in bold.
} \label{tab:ca_baselines}
\centering
\footnotesize
\renewcommand{\arraystretch}{1.2}
\setlength{\tabcolsep}{5pt}
\resizebox{\textwidth}{!}{
\begin{tabular}{l ccc ccc ccc ccc}
\toprule
\multirow[b]{2}{*}{\textbf{Model}} &
  \multicolumn{3}{c}{\textbf{No Glasses vs. Glasses}} &
  \multicolumn{3}{c}{\textbf{Male vs. Female}} &
  \multicolumn{3}{c}{\textbf{Head Pose (Frontal vs. Right/Left)}} &
  \multicolumn{3}{c}{\textbf{Smile vs. Non-Smiling}} \\
\cmidrule(lr){2-4}\cmidrule(lr){5-7}\cmidrule(lr){8-10}\cmidrule(lr){11-13}
 & $C\downarrow$ & $S\uparrow$ & $\Delta\downarrow$
 & $C\uparrow$ & $S\downarrow$ & $\Delta\downarrow$
 & $C\uparrow$ & $S\downarrow$ & $\Delta\downarrow$
 & $C\uparrow$ & $S\downarrow$ & $\Delta\downarrow$ \\
\midrule
MM-cVAE        & 0.58 ± 0.01 & 0.68 ± 0.01 & 0.40 
              & 0.64 ± 0.02 & 0.56 ± 0.02 & 0.42 
              & 0.73 ± 0.01 & 0.69 ± 0.01 & 0.46 
              & 0.66 ± 0.01 & 0.60 ± 0.02 & 0.56 \\
SepVAE         & 0.59 ± 0.01 & 0.70 ± 0.01 & 0.39 
              & 0.65 ± 0.01 & 0.59 ± 0.03 & 0.44 
              & 0.69 ± 0.02 & 0.68 ± 0.01 & 0.49 
              & 0.65 ± 0.01 & 0.57 ± 0.01 & 0.58 \\
Double InfoGAN & 0.65 ± 0.01 & 0.82 ± 0.01 & 0.33 
              & 0.60 ± 0.03 & 0.55 ± 0.01 & 0.45 
              & 0.71 ± 0.01 & 0.57 ± 0.02 & 0.36 
              & 0.64 ± 0.00 & 0.57 ± 0.01 & 0.43 \\
\textbf{Ours (pSp-cs)}  & \textbf{0.52 ± 0.03} & \textbf{0.98 ± 0.01} & \textbf{0.04} 
              & \textbf{0.80 ± 0.01} & \textbf{0.51 ± 0.03} & \textbf{0.21} 
              & \textbf{0.95 ± 0.01} & \textbf{0.52 ± 0.05} & \textbf{0.07} 
              & \textbf{0.90 ± 0.01} & \textbf{0.52 ± 0.01} & \textbf{0.12} \\
\midrule
\textcolor{blue}{Expected} 
              & \textcolor{blue}{0.5} & \textcolor{blue}{1.0} & \textcolor{blue}{0} 
              & \textcolor{blue}{1.0} & \textcolor{blue}{0.5} & \textcolor{blue}{0} 
              & \textcolor{blue}{1.0} & \textcolor{blue}{0.5} & \textcolor{blue}{0} 
              & \textcolor{blue}{1.0} & \textcolor{blue}{0.5} & \textcolor{blue}{0} \\
\bottomrule
\end{tabular}
}
\end{table*}

\subsection{Results}
\label{sec:results}
\subsubsection{Comparison with CA Baselines}
\label{sec:comp_ca}
In Fig.~\ref{fig:ca_baselines}, we compare our methods with prior CA baselines on reconstruction and salient-attribute swapping using high-quality FFHQ data. We use as background ($X$) face images of people without glasses and as target ($Y$) people with glasses. Our -cs models produce reconstructions that are noticeably sharper, more realistic, and better at preserving identity and fine details than previous approaches, which often show blurring or visual artifacts. For salient swapping, our models clearly transfer the glasses while maintaining identity and face details. \YL{Additional qualitative examples are shown in Fig.~22 of the Suppl..} Table~\ref{tab:ca_quality_comp} reports the quantitative results for reconstruction and attribute swapping, where our -cs models achieve significantly better performance than all CA baselines. Table~\ref{tab:ca_baselines} evaluates how well the learned latent factors separate attribute information. Since glasses are the only salient attribute in $Y$, all other attributes (e.g., smile) should be encoded in the common factors. Our model achieves the lowest separation metric $\Delta$ for all attributes, demonstrating more accurate semantic separation compared to previous CA methods.
\begin{figure*}[ht!]
    \centering
    \includegraphics[width=0.9\linewidth]{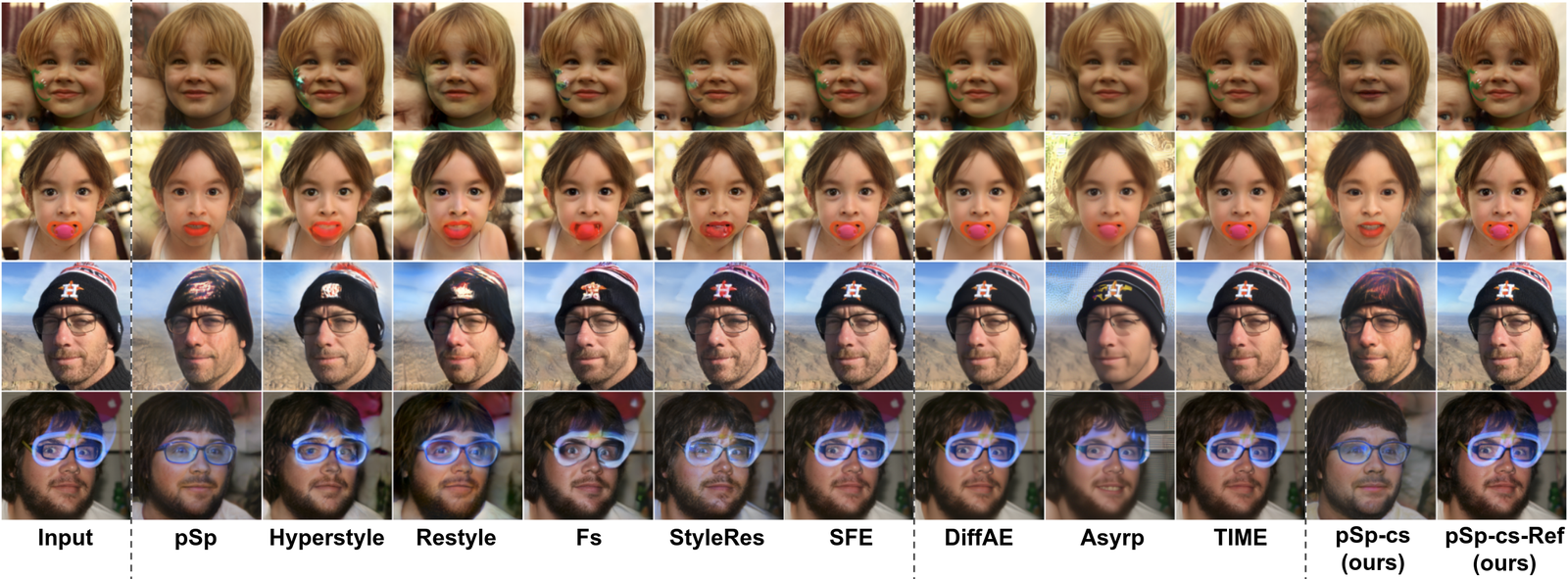}
    \caption{Comparison of image reconstructions from StyleGAN-based encoders (Cols. 2–7), DMs (Cols. 8–10), and our methods (Cols. 11–12).}
    \label{fig:recon_quality_stylegan}
\end{figure*}
\subsubsection{Reconstruction Quality}
\label{sec:recon}

Fig.~\ref{fig:recon_quality_stylegan} compares our reconstructions with those produced by StyleGAN-based encoders and three DMs. Relative to the StyleGAN baselines, our pSp-cs-Ref yields noticeably higher fidelity, successfully recovering fine-grained structures such as green graffiti and subtle facial accessories; among prior methods, only SFE achieves a comparable level of detail. Consistent with the quantitative results reported in  \YL{Table~XII of the Suppl.,} our method matches SFE on MS-SSIM and identity similarity (differences $<\!0.01$), while obtaining a lower FID. Compared with DMs, our reconstructions are both visually and quantitatively very similar, achieving a lower FID (e.g., \(2.406\) vs. DiffAE), with differences $<\!0.01$  in the remaining metrics.

\begin{figure*}[ht!]
    \centering
    \includegraphics[width=0.9\linewidth]{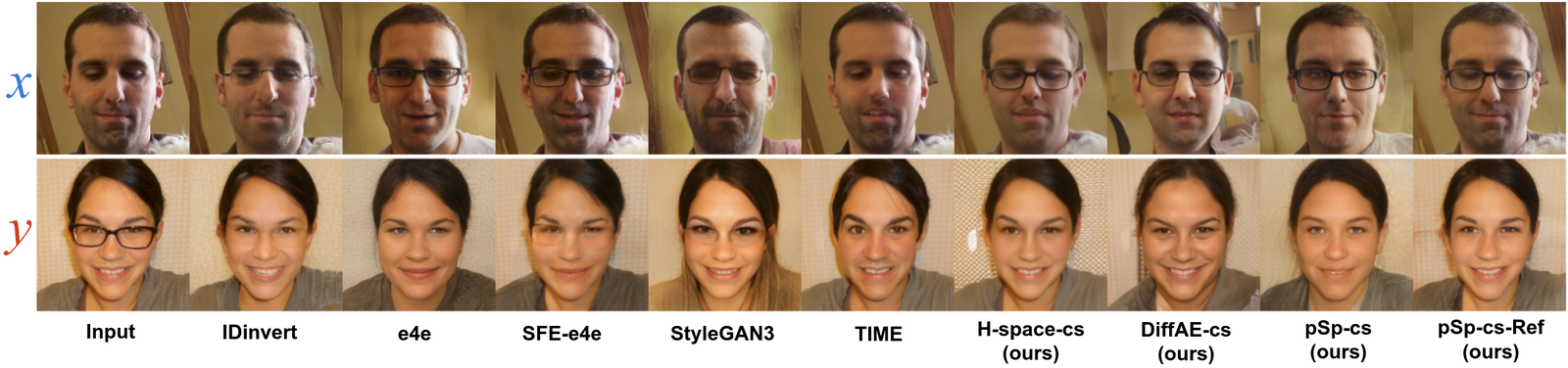}
    \caption{Visual comparison of our method and SOTA approaches for attribute editing.
    Columns 2--5 show supervised methods; columns 6--10 show weakly-supervised (dataset-guided) approaches, including TIME and our -cs models.}
    \label{fig:attr_edit_compare}
\end{figure*}

\subsubsection{Performance of Attribute Editing}
\label{sec:comp_editing}
In Fig.~\ref{fig:attr_edit_compare}, we present qualitative results for attribute (glasses) editing, comparing our method with strong baselines that achieve over $90\%$ attribute-classification accuracy (see Table~\ref{tab:attr_edit_compare}), as well as with TIME. Both supervised approaches and TIME introduce undesirable changes, such as altering the smile, degrading identity, or failing to fully remove glasses. In contrast, pSp-cs-Ref performs clean and precise edits while preserving facial realism, identity, and unrelated attributes. Table~\ref{tab:attr_edit_compare} reports the corresponding quantitative evaluation, including both editing metrics (ID-Sim, FID) and attribute-classification metrics (Accuracy, AUC) computed using a high-accuracy glasses classifier trained on real images (above $98\%$ on the test set). More details about the classifier can be found in the Suppl. Sec. II-B. A successful attribute-editing method should shift edited samples from the source distribution toward the desired post-edit distribution, which is reflected by achieving a lower FID in the target than in the source (the expected FID trend). Compared with both supervised baselines and TIME, pSp-cs-Ref attains the highest ID-Sim, the lowest FID, and strong classification accuracy, while maintaining the expected FID trend. Both h-space-cs and DiffAE-cs underperform with respect to pSp-cs-Ref, probably because they are less regularized. Exploring more suitable regularizations for DMs is left for future work. More visual examples can be found in Fig. 23 in the Suppl..

\begin{table*}[t]
\caption{Quantitative Comparison Results for Editing Performance.
Expected classification results are those obtained from the classifier, trained on real images, on the test set. \textbf{SL}: supervised learning (classifier-guided); \textbf{WSL}: weakly-supervised learning.}
\label{tab:attr_edit_compare}
\centering
\footnotesize
\resizebox{1.0\textwidth}{!}{%
\begin{tabular}{c c c
                *{3}{c}
                *{3}{c}
                *{2}{c}}
\toprule
& & &
\multicolumn{3}{c}{\textbf{$X \rightarrow Y$ (Add glasses)}} &
\multicolumn{3}{c}{\textbf{$Y \rightarrow X$ (Remove glasses)}} &
\multicolumn{2}{c}{\textbf{Attribute Classification (X+Y)}} \\
\cmidrule(lr){4-6}\cmidrule(lr){7-9}\cmidrule(lr){10-11}
\textbf{} & \textbf{Inversion} & \textbf{Direction} &
ID\mbox{-}Sim $\uparrow$ & FID\mbox{-}X & FID\mbox{-}Y $\downarrow$ &
ID\mbox{-}Sim $\uparrow$ & FID\mbox{-}X $\downarrow$ & FID\mbox{-}Y &
Accuracy $\uparrow$ & AUC $\uparrow$ \\
\midrule
\multirow{9}{*}{\textbf{SL}} 
& e4e                & InterFaceGAN  & 0.577 & 60.619 & 42.425 & 0.592 & 61.919 & 68.895 & 0.956 & 0.956 \\
& pSp                & Stylespace  & 0.418 & 66.265 & 53.985 & 0.625 & 45.430 & 50.022 & 0.850 & 0.968 \\
& SFE-e4e            & InterFaceGAN  & 0.667 & 40.060 & 23.314 & 0.744 & 30.302 & 35.675 & 0.973 & 0.975 \\
& SFE-e4e            & Stylespace  & 0.620 & 36.171 & 35.504 & 0.704 & 29.653 & 27.540 & 0.891 & 0.981 \\
& SFE-pSp            & InterFaceGAN  & 0.710 & 35.752 & 24.342 & 0.789 & 32.134 & 22.050 & 0.795 & 0.985 \\
& SFE-pSp            & Stylespace  & 0.505 & 44.742 & 39.099 & 0.713 & 37.317 & 24.301 & 0.703 & 0.954 \\
& ID-Invert          & InterFaceGAN   & 0.488 & 29.493 & 35.792 & 0.390 & 31.081 & 55.505 & 0.923 & 0.964 \\
& StyleGAN3 + pSp    & InterFaceGAN   & 0.538 & 64.981 & 37.690 & 0.582 & 56.302 & 62.793 & 0.915 & 0.918 \\
& StyleGAN3 + e4e    & InterFaceGAN   & 0.697 & 60.337 & 57.697 & 0.699 & 61.260 & 61.961 & 0.748 & 0.794 \\
\midrule
\multirow{5}{*}{\textbf{WSL}}
& \multicolumn{2}{c}{TIME (Counterfactual analysis)} & 0.686 & 26.492 & 27.609 & 0.692 & 27.632 & 25.929 & 0.834 & 0.844 \\
& \multicolumn{2}{c}{DiffAE\mbox{-}cs}         & 0.570 & 54.076 & 39.050  & 0.540 & 38.598 & 44.067 & 0.909 & 0.912 \\
& \multicolumn{2}{c}{H\mbox{-}space\mbox{-}cs}  & 0.604 & 39.736 & 36.709 & 0.624 & 39.486 & 51.774 & 0.813 & 0.814 \\
& \multicolumn{2}{c}{\pspcs}             & 0.601 & 57.033 & 46.696 & 0.650 & 55.088 & 60.309 & \textbf{0.975} & \textbf{0.990} \\
& \multicolumn{2}{c}{\pspcsref}     & \textbf{0.804} & 25.508 & \textbf{23.302} & \textbf{0.809} & \textbf{25.222} & 31.625 & 0.974 & 0.988 \\
\midrule
& \multicolumn{2}{c}{\textcolor{blue}{Expected}} 
& \textcolor{blue}{1.000} & \multicolumn{2}{c}{\textcolor{blue}{FID\mbox{-}Y $<$ FID\mbox{-}X}} 
& \textcolor{blue}{1.000} & \multicolumn{2}{c}{\textcolor{blue}{FID\mbox{-}X $<$ FID\mbox{-}Y}} 
& \textcolor{blue}{0.981} & \textcolor{blue}{0.993} \\
\bottomrule
\end{tabular}%
}
\end{table*}

\begin{figure}[ht!]
\begin{center}
   \includegraphics[width=0.8\linewidth]{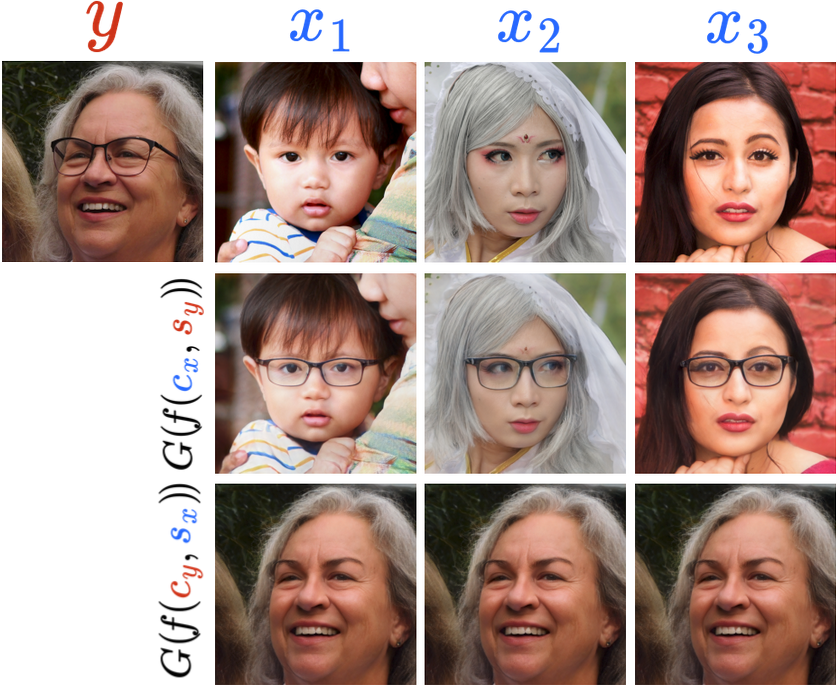}
   \caption{Faithful salient swap. Glasses are correctly encoded in $s_y$ and realistically added to the three background images $x_1$, $x_2$, $x_3$. As expected, no glass information is encoded in $s_x$.}
\label{fig:y_x1x2x3}
\end{center}
\end{figure}

\begin{figure}[ht!]
\begin{center}
   \includegraphics[width=0.9\linewidth]{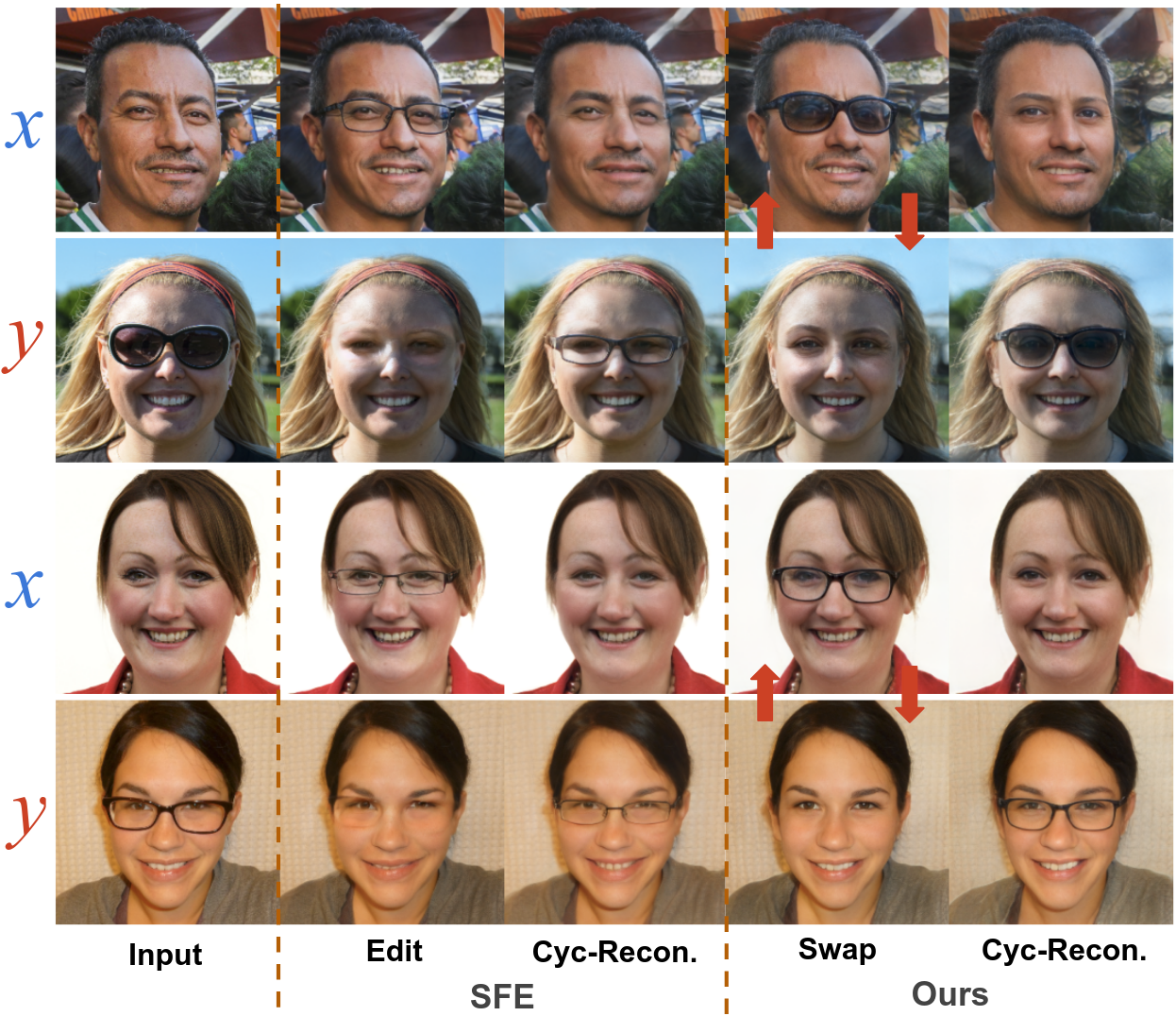}
\caption{\textbf{Cycle reconstruction.} For each input (leftmost), we show its SFE edit (Col.~2) and our salient-factor swap (Col.~4). The cycle reconstructions (Cols.~3 and 5) are obtained by applying the inverse edit or swap to the generated results. SFE subtracts the \textbf{same} global direction from the encoded latent, while our method re-exchanges the \textbf{image-specific} salient factors.}
\label{fig:cycle_recon}
\end{center}
\end{figure}

\subsubsection{Faithful Salient Swaps}
In Fig.~\ref{fig:y_x1x2x3}, we show results obtained by swapping the salient factors of a target image with those from three different background images. We denote the swapped image in the $F$-space as $G(f(c_x, s_y))$, 
which is equivalent to the previously defined $G(\hat{f}_{x\to y})$ in Sec.~\ref{sec:f_refine}, and is obtained by replacing the salient factor of $x$ with that of $y$ while keeping the common factor $c_x$ fixed. Similarly, $G(f(c_y, s_x))$ is equivalent to $G(\hat{f}_{y\to x})$, which is produced by replacing the salient factor of $y$ with that of $x$. It can be observed that the information about glasses is correctly encoded in $s_y$ while the common information about the face traits, gender, hair style, etc., is encoded in $c$. The swapping is highly realistic and faithful. Furthermore, we can check that no glass information is encoded in the estimated $s_x$, as expected, since the three faces at the bottom are exactly the same. More examples can be found in Fig.~24 in the Suppl.. 

\subsubsection{SFE Comparison}
Fig.~\ref{fig:cycle_recon} compares our method with SFE (using e4e and InterFaceGAN, the top-performing SL baseline in Table~\ref{tab:attr_edit_compare}) for cycle reconstruction. In this setting, edited images from SFE and swapped images from our pSp-cs-Ref are each fed back into their respective model to recover the original image. SFE applies a global ``glasses'' edit, thus adding always the same kind of glasses. Differently, our method estimates the correct kind of glasses (either eyeglasses or sunglasses), preserving many fine-grained details. Moreover, \YL{Fig.~25 of the Suppl.} 
reveals an edit-strength trade-off for SFE: larger $\alpha$ strengthens the edit (higher Acc) but harms identity, while smaller $\alpha$ under-edits and yields worse FID. By contrast, for both adding and removing glasses, our method achieves the lowest FID and highest Acc, maintaining high identity in a \emph{single} salient-swap pass (no $\alpha$ tuning). A similar comparison with IDInvert and InterfaceGAN can be found in Fig. 26 of the Suppl..

\begin{figure}[ht!]
    \centering
    \includegraphics[width=0.9\linewidth]{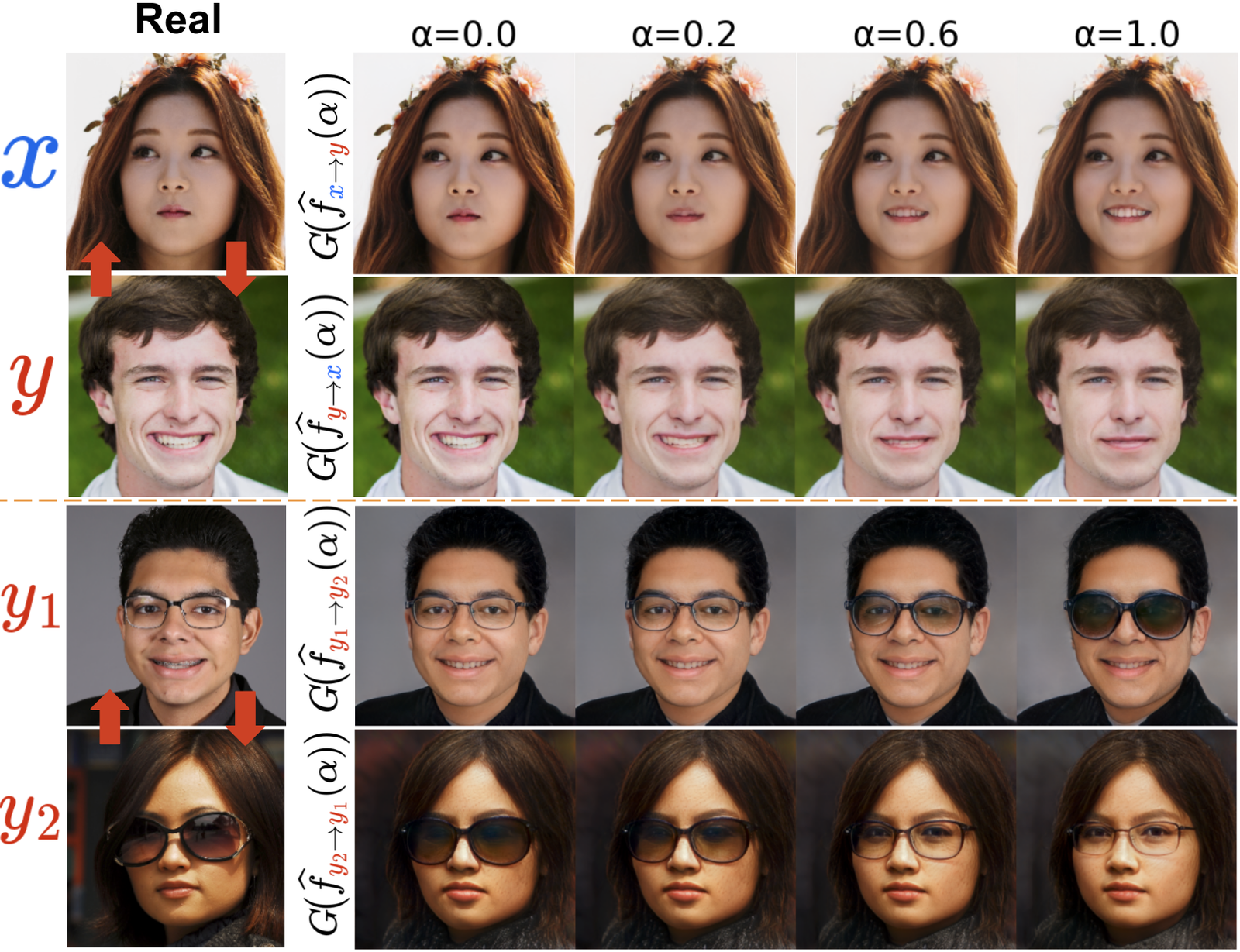}
    \caption{\textbf{Interpolation along salient factors under the \textit{background/target} assumption.}
    Rows 1–2: Interpolations between background ($X$) and target ($Y$) images, where the salient attribute present only in $Y$ is \emph{smile}. 
    Rows 3–4: Interpolations between target images, where the salient attribute present only in $Y$ is \emph{glasses}. Red arrows indicate the transfer of the corresponding salient attribute from one image to another.}
    \label{fig:interp_xy_y1y2}
\end{figure}

\subsubsection{Interpolation Along Salient Factors} 
\label{sec:interpolation}
Fig.~\ref{fig:interp_xy_y1y2} shows interpolations along salient factors under the \textit{background/target} setting. For the top two rows, we encode background image $x$ and target image $y$ into $(c_x,s_x)$ and $(c_y,s_y)$, compute the semantic shifts $\Delta f_{x\rightarrow y}$ and $\Delta f_{y\rightarrow x}$ by swapping their salient factors using Eqs.~\eqref{equ:delta_x2y} and \eqref{equ:delta_y2x}, and interpolate in $F$-space via: $\widehat{f}_{x \rightarrow y}(\alpha)\!=\!F_{\text{Adpt}}(f_x \oplus \alpha\,\Delta f_{x\!\rightarrow\!y})$ and $\widehat{f}_{y \rightarrow x}(\alpha)\!=\!F_{\text{Adpt}}(f_y \oplus \alpha\,\Delta f_{y \rightarrow x})$. We then synthesize $G(\widehat{f}_{x\rightarrow y}(\alpha))$  (row~1 of Fig.~\ref{fig:interp_xy_y1y2}), where increasing $\alpha\!\in\![0,1]$ gradually replaces the salient pattern of $x$ with that of $y$. Likewise, $G(\widehat{f}_{y\rightarrow x}(\alpha))$ (row~2) smoothly replaces the salient pattern of $y$ with that of $x$. The transitions are smooth and realistic: only the salient attribute changes, while identity, hairstyle, pose, and background remain consistent. For the bottom two rows, we apply the same procedure to two target images $y_1$ and $y_2$ by swapping their salient factors to compute the feature shifts, yielding smooth exchanges of both the salient attribute and its style (e.g., transitioning between sunglasses and eyeglasses). More interpolation examples between two images are provided in the \YL{Suppl. (Figs.~27 and~28).}

\begin{figure}[t!]
    \centering
    \includegraphics[width=0.9\linewidth]{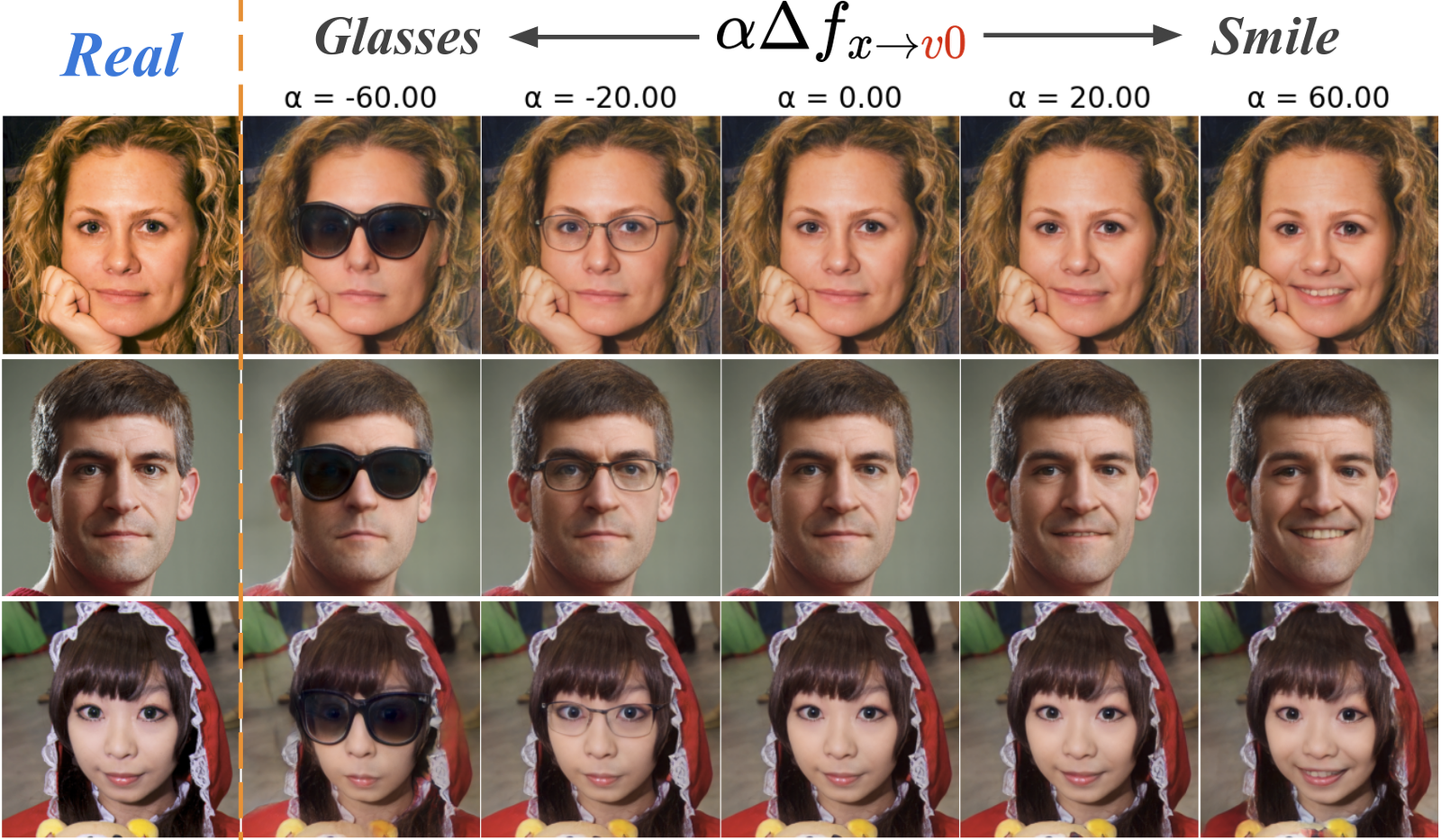}
    \caption{\textbf{Semantic traversal along the first PCA direction $v_0$ of learned salient features.} 
    Each row shows a test image from $X$ whose salient $s_x$ has been swapped with values sampled along $v_0$ of a PCA computed using salient from $Y$. The salient attributes in $Y$, not in $X$, are glasses and smile.}
    \label{fig:interp_pca}
\end{figure}

 In Fig.~\ref{fig:interp_pca}, we swap the salient factor of images from $X$ with salient values sampled along the first direction of a PCA computed using the salient factors of 10k test images from $Y$. The salient attributes of $Y$ are glasses and smile. The results show that the dominant variation in the salient space is the direction going from glasses to smile, as expected. Additional interpolation results using the first three PCA directions \(v_0, v_1, v_2\) are provided in the \YL{Suppl. (Figs.~29--31).}



\subsubsection{Evaluation on Diverse Attributes \& Datasets}
\label{sec:results_multi}
\begin{figure}[ht!]
    \centering
    \includegraphics[width=1.0\linewidth]{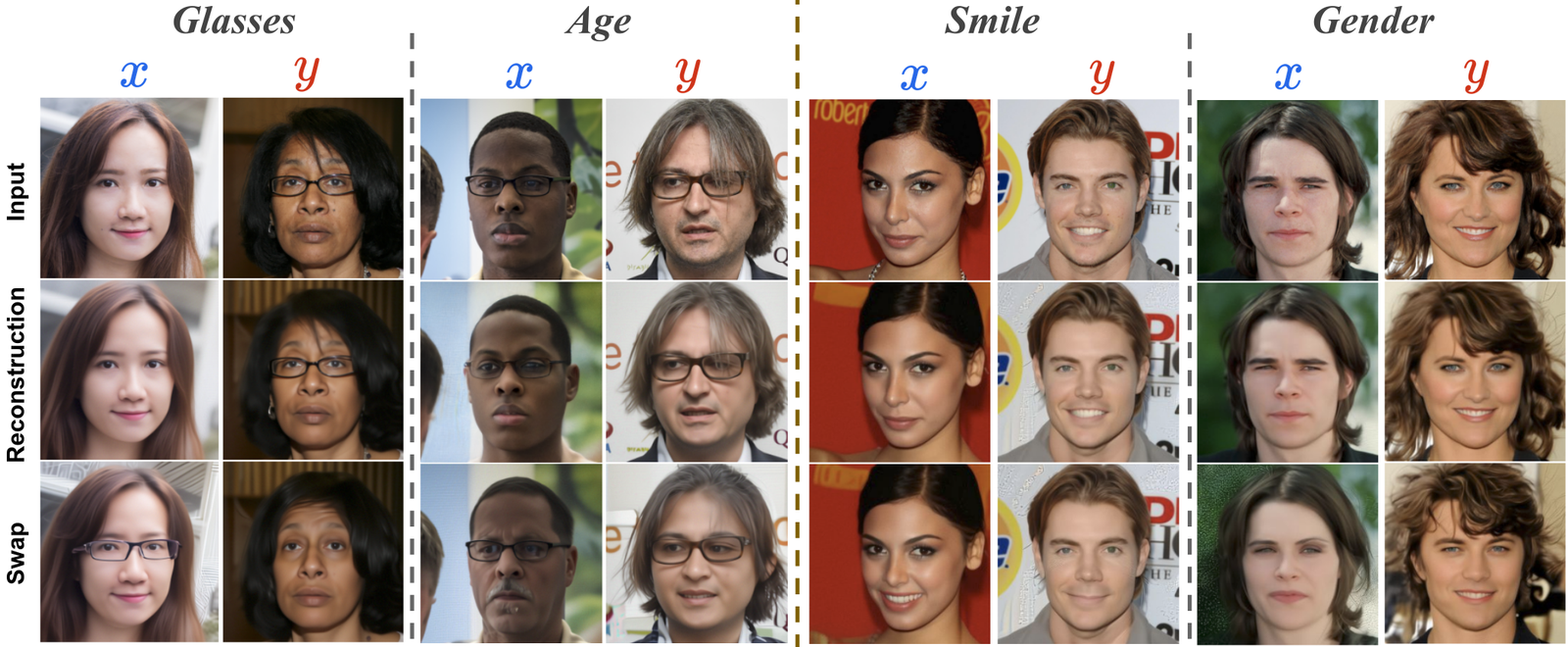}
    \caption{Results of \textbf{h-space-cs} on FFHQ (glasses, age) and CelebA-HQ (smile, gender) under the \textit{background/target} assumption.}
    \label{fig:diff_Asyrp}
\end{figure}

In Fig.~\ref{fig:diff_Asyrp}, we show results of our \textbf{h-space-cs} model trained under the background/target setting using the FFHQ \textit{glasses} and \textit{age} attributes, as well as the CelebA-HQ \textit{smile} and \textit{gender} attributes (see Table~\ref{tab:n_training} for details of the corresponding $X$/$Y$ splits). \YL{Suppl. Figs.~32 and 33} provide additional results: \textbf{pSp-cs-Ref} on the FFHQ \textit{smile}, \textit{age}, \textit{head pose}, and \textit{gender} attributes, and \textbf{DiffAE-cs} on FFHQ \textit{glasses}, respectively. As before, the separation uses only \textit{dataset}-level labels, with no attribute supervision. All of our -cs models produce high-quality reconstructions and clean, attribute-specific swaps, enabling controllable edits. \YL{Table~XIII in the Suppl.} reports qualitative results of pSp-cs-Ref trained on each of the FFHQ $X$/$Y$ dataset splits listed in Table~\ref{tab:n_training}. As expected, the training attribute is encoded in the salient factor, while the remaining attributes reside in the common factor.


\begin{figure}[ht!]
\begin{center}
   \includegraphics[width=1.0\linewidth]{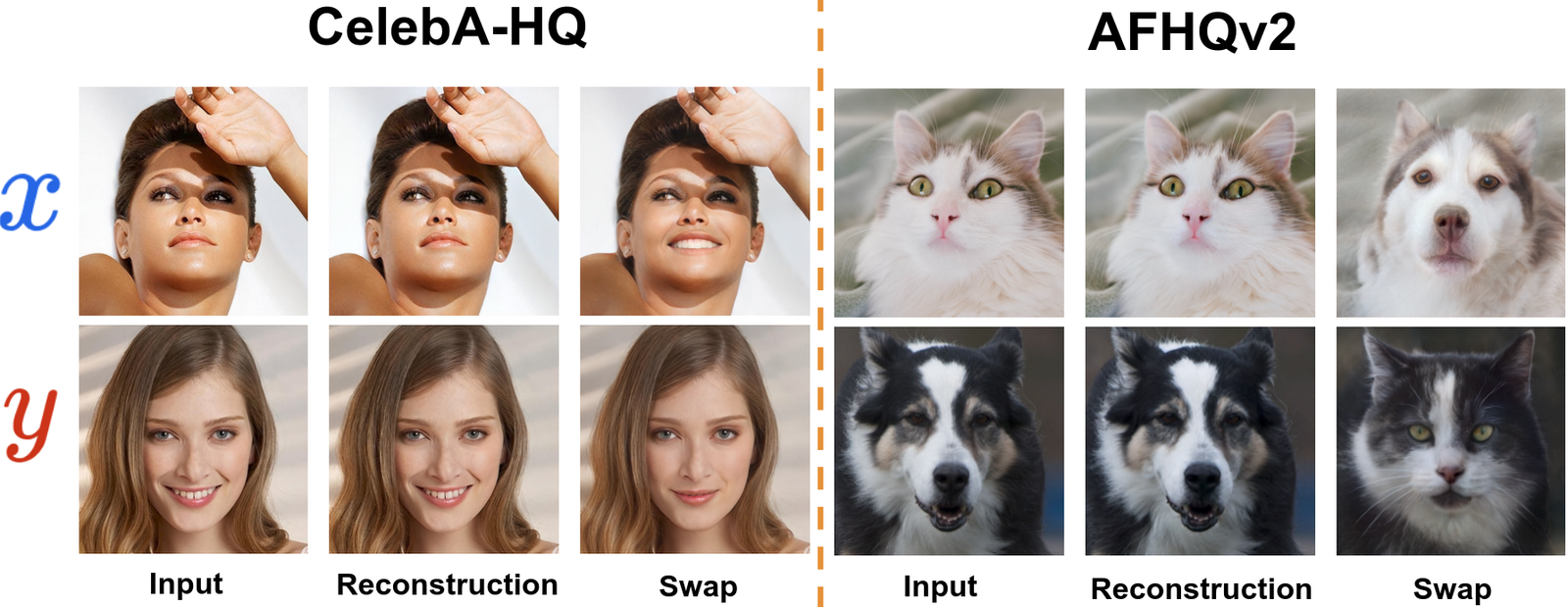}
\end{center}
    \caption{Results of pSp-cs-Ref on CelebA-HQ (non-smiling vs. smiling) and AFHQv2 (cat vs. dog).}
\label{fig:afhq_celebahq}
\end{figure}

Fig.~\ref{fig:afhq_celebahq} shows results of our \textbf{pSp-cs-Ref} on AFHQv2 and CelebA-HQ under the \textit{background/target} setting, demonstrating high-quality reconstruction and attribute swapping. Table~\ref{tab:several_datasets} and Suppl. Table~XIV reports quantitative results on AFHQv2, BraTS, and CelebA-HQ (gender and smile), including Double InfoGAN, the best-performing CA baseline in our evaluation. Our -cs models consistently deliver high-quality reconstructions, achieve more accurate latent separation and enable effective image editing across all datasets. \YL{See Fig.~34 of the Suppl.} for additional examples on AFHQv2 (cat vs.\ dog).

\begin{figure}[ht!]
    \centering
    \includegraphics[width=0.9\linewidth]{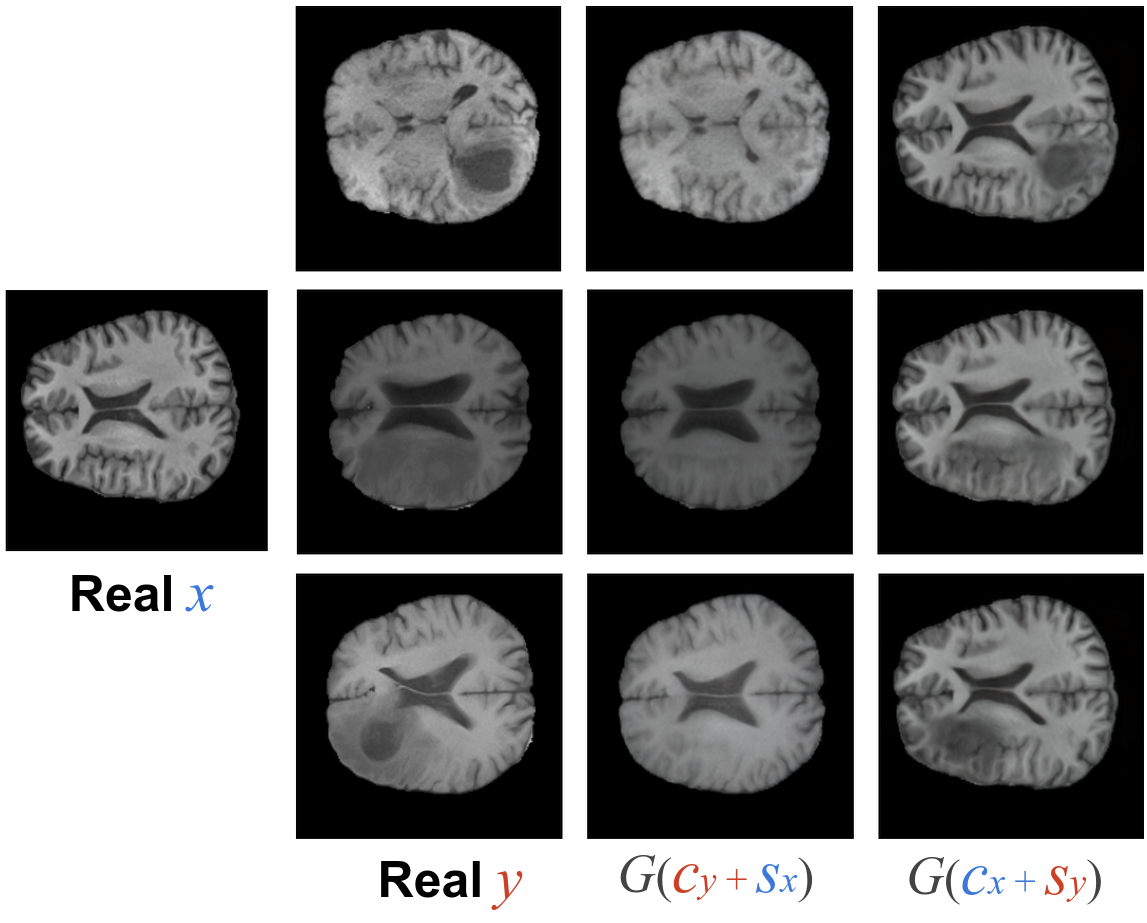}
    \caption{\textbf{Brain MRI results.} Swapping result between MR images of healthy brains ($X$) and brains with tumors ($Y$).}
    \label{fig:brats}
\end{figure}

In Fig.~\ref{fig:brats}, we show swapping results between MR images of healthy brains ($X$) and brains with tumors ($Y$). Our method effectively learns the common and salient (tumor) factors even with a relatively small training set (8K samples). \YL{Suppl. Fig.~35 and Fig.~36} provide additional examples, \YL{with Fig.~36} specifically visualizing interpolations between healthy and tumor-affected brain MR scans. The smooth and consistent transitions indicate that our model captures tumor-specific variations in the salient space while preserving the overall brain structure. These results demonstrate the effectiveness of our method in medical imaging as well.

\begin{table*}[ht!]
\caption{Quantitative results for AFHQv2, BraTS, and CelebA-HQ (gender and smile) datasets under the {\normalfont\itshape background/target} assumption.}
\label{tab:several_datasets}
\centering
\footnotesize
\resizebox{\textwidth}{!}{%
\begin{tabular}{l l
                *{5}{c}
                *{5}{c}
                *{3}{c}}
\toprule
& &
\multicolumn{5}{c}{\textbf{Reconstruction (X)}} &
\multicolumn{5}{c}{\textbf{Reconstruction (Y)}} &
\multicolumn{3}{c}{\textbf{Attribute Swapping}} \\
\cmidrule(lr){3-7}\cmidrule(lr){8-12}\cmidrule(lr){13-15}
\textbf{Dataset} & \textbf{Model} &
LPIPS$\downarrow$ & MSE$\downarrow$ & MS\mbox{-}SSIM$\uparrow$ & ID\mbox{-}Sim$\uparrow$ & FID$\downarrow$ &
LPIPS$\downarrow$ & MSE$\downarrow$ & MS\mbox{-}SSIM$\uparrow$ & ID\mbox{-}Sim$\uparrow$ & FID$\downarrow$ &
FID\mbox{-}Y$\downarrow$ & FID\mbox{-}X$\downarrow$ & Acc$\uparrow$ \\
\midrule
\multirow{3}{*}{AFHQv2}
& Double InfoGAN & 0.548 & 0.024 & 0.435 & 0.669 & 117.727 & 0.535 & 0.027 & 0.453 & 0.612 & 119.213 & 126.933 & 129.996 & 0.766 \\
& pSp\mbox{-}cs         & 0.210 & 0.018 & 0.584 & 0.855 & 22.100  & 0.246 & 0.020 & 0.551 & 0.778 & 37.432  & 82.638  & 89.192  & 0.845 \\
& pSp\mbox{-}cs\mbox{-}Ref      & \textbf{0.035} & \textbf{0.002} & \textbf{0.971} & \textbf{0.986} & \textbf{6.661}   & \textbf{0.028} & \textbf{0.001} & \textbf{0.979} & \textbf{0.987} & \textbf{5.872}   & \textbf{32.273}  & \textbf{42.390}  & \textbf{0.856} \\
\midrule
\multirow{3}{*}{BraTS}
& Double InfoGAN & 0.269 & 0.013 & 0.631 & 0.650 & 235.742 & 0.260 & 0.010 & 0.625 & 0.646 & 207.727 & 228.459 & 239.073 & 0.971 \\
& pSp\mbox{-}cs         & 0.064 & 0.003 & 0.842 & 0.881 & 30.487  & 0.061 & 0.002 & 0.849 & 0.893 & 27.661  & 48.663  & 56.665  & 0.998 \\
& pSp\mbox{-}cs\mbox{-}Ref      & \textbf{0.006} & \textbf{0.0002} & \textbf{0.997} & \textbf{0.997} & \textbf{5.618}   & \textbf{0.006} & \textbf{0.0002} & \textbf{0.997} & \textbf{0.997} & \textbf{4.709}   & \textbf{37.275}  & \textbf{43.473}  & \textbf{0.999} \\
\midrule
\multirow{3}{*}{CelebA\mbox{-}HQ Gender}
& Double InfoGAN & 0.394 & 0.024 & 0.602 & 0.179 & 130.277 & 0.375 & 0.021 & 0.610 & 0.199 & 81.527  & 88.475  & 149.370 & 0.704 \\
& pSp\mbox{-}cs         & 0.167 & 0.013 & 0.717 & 0.834 & 27.655  & 0.179 & 0.015 & 0.723 & 0.786 & 33.238  & 76.442  & 85.337  & 0.732 \\
& pSp\mbox{-}cs\mbox{-}Ref      & \textbf{0.016} & \textbf{0.001} & \textbf{0.986} & \textbf{0.991} & \textbf{3.245}   & \textbf{0.019} & \textbf{0.001} & \textbf{0.990} & \textbf{0.990} & \textbf{3.438}   & \textbf{72.708}  & \textbf{77.649}  & \textbf{0.760} \\
\midrule
\multirow{3}{*}{CelebA\mbox{-}HQ Smile}
& Double InfoGAN & 0.336 & 0.023 & 0.614 & 0.197 & 76.384  & 0.308 & 0.020 & 0.633 & 0.227 & 67.494  & 67.475  & 82.482  & 0.674 \\
& pSp\mbox{-}cs         & 0.204 & 0.017 & 0.682 & 0.698 & 34.139  & 0.180 & 0.015 & 0.698 & 0.768 & 30.317  & 37.728  & 37.500  & 0.689 \\
& pSp\mbox{-}cs\mbox{-}Ref      & \textbf{0.018} & \textbf{0.001} & \textbf{0.985} & \textbf{0.991} & \textbf{2.663}  & \textbf{0.018} & \textbf{0.001} & \textbf{0.985} & \textbf{0.992} & \textbf{2.353}   & \textbf{17.173}  & \textbf{18.802}  & \textbf{0.756} \\
\bottomrule
\end{tabular}%
}
\end{table*}

\subsubsection{Multiple-attribute and multiple-salient Settings} 
\label{sec:CA_assumptions}
\begin{figure*}[!t]
\centering
    \subfloat[]{\includegraphics[width=0.47\linewidth]{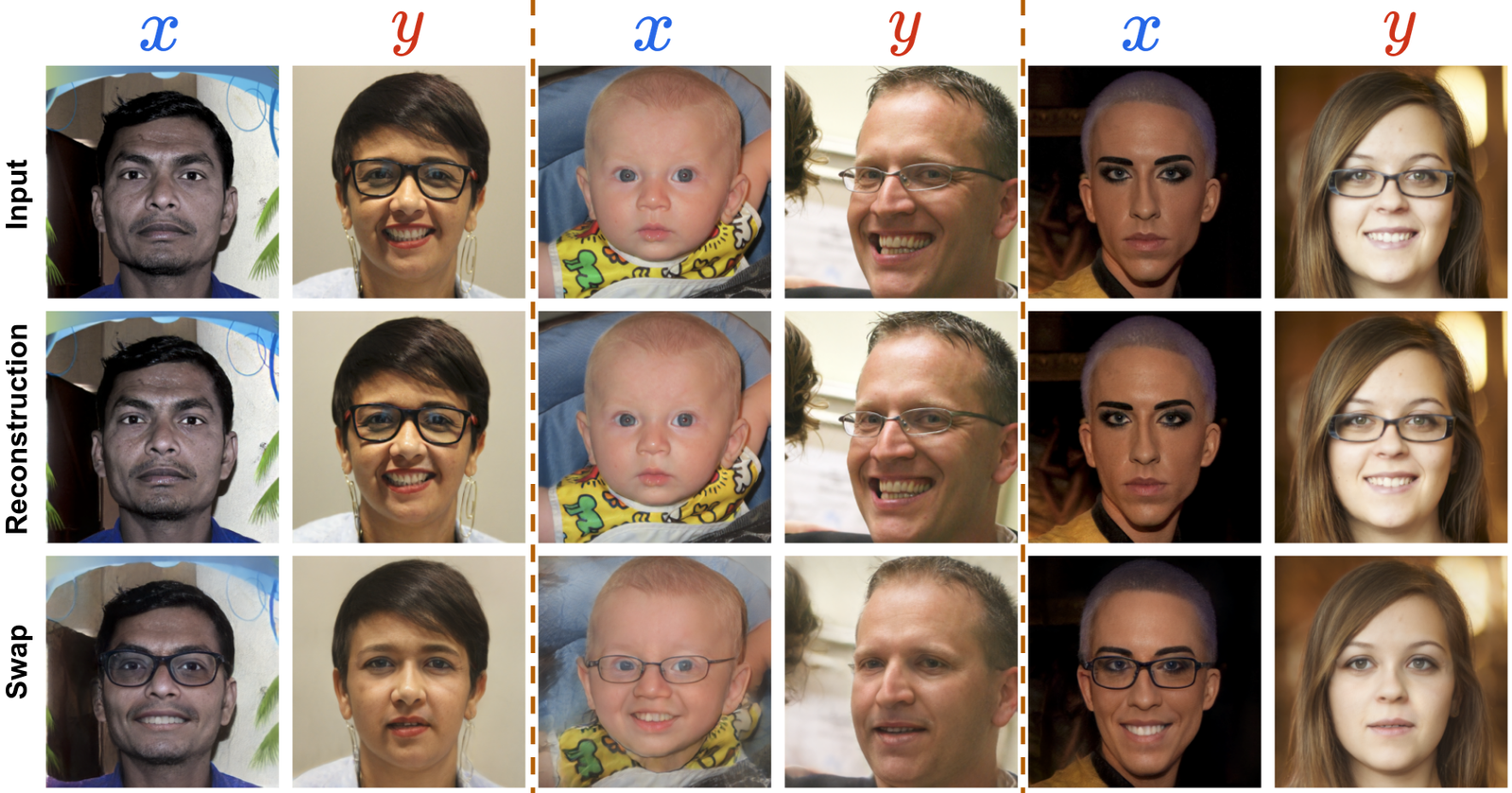}}%
    \hfil
    \subfloat[]{\includegraphics[width=0.47\linewidth]{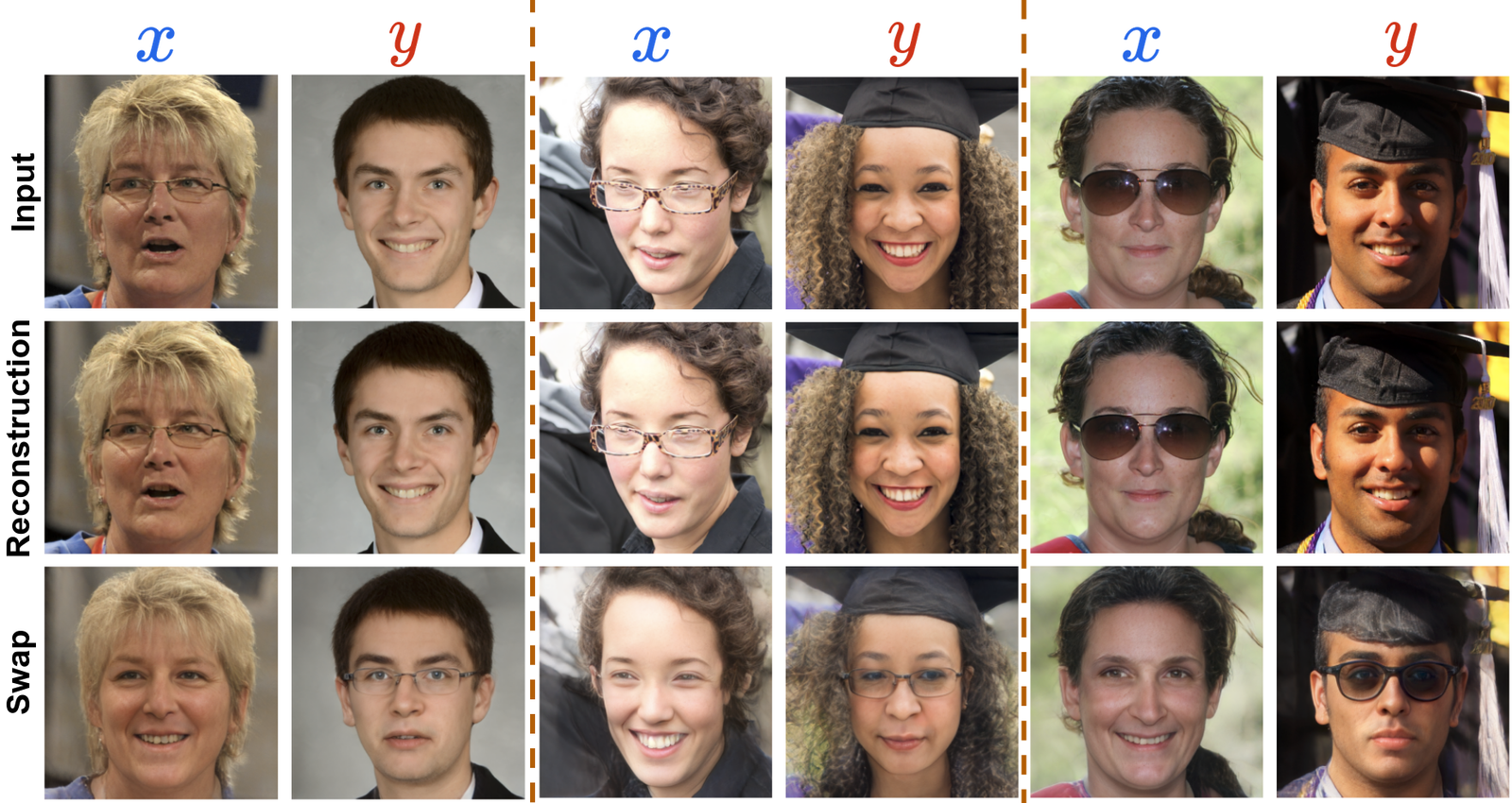}}%
    \caption{(a) \textbf{Multiple-attribute case}: $Y$ contains two salient attributes (glasses, smile) absent from $X$.
    (b) \textbf{Multiple–salient case}: each of $X$ and $Y$ contains a salient attribute (glasses only in $X$, smile only in $Y$).}
    \label{fig:multi_cases}
\end{figure*}

In Fig.~\ref{fig:multi_cases}(a), we show reconstruction and swapping results under the \textit{multiple-attribute} assumption, where the target set $Y$ contains two salient attributes (glasses and smile) that are absent from the background set $X$. Fig.~\ref{fig:multi_cases}(b) shows results for the \textit{multiple-salient} setting, where each of $X$ and $Y$ contains a salient pattern (glasses in $X$, smile in $Y$). These results demonstrate that our approach adapts effectively in both scenarios. \YL{See Suppl. Fig.~37 for more examples.} Quantitative results for the two cases are reported in the Suppl., in Table XIII for latent separation and in Table XVII for image reconstruction quality, respectively. 

\subsection{Dataset Bias Challenges}
\noindent Here, we inspect the presence of biases in the datasets. \YL{Table~XVI in the Suppl. reports} correlations between the attributes used to define training splits and those present in the data. For example, the ``No Glasses vs. Glasses'' split is age-imbalanced ($\rho$=0.366). This is also analyzed in \YL{Fig.~38 of the Suppl.} where we can see that $X$ (No Glasses) skews to the left (younger subjects), whereas $Y$ (Glasses) contains more older subjects. Such imbalance can degrade performance and induce leakage across factors. For instance, as reported in \YL{Suppl. Table~XIII}, when training on the Glasses attribute (row 1, glasses are present in Y but not in X), the age classification accuracies are $0.84$ for $c$ and $0.64$ for $s$, indicating that age information incorrectly leaks into the salient factor. \PG{Additional bias-control strategies, such as class-balanced sampling, targeted augmentation or specific regularization as in \cite{barbano_unbiased_2023}, could be used. This is left as future work.}


\section{Conclusions and Persepctives}
\label{sec:conclusions}
Learning semantically meaningful latent factors has long been a central goal in GAN research and has recently gained increasing attention in diffusion models. Here, we propose a novel framework that can work with both StyleGAN and DDIM, to learn common and salient generative factors between two datasets. Our framework extends the CA paradigm and achieves superior performance over previous CA baselines in both high-quality image generation, editing (i.e., salient attribute swapping) and latent space separation. 

In this work, for the first time in CA, we proposed a framework to analyze two complex settings: multiple attributes and multiple salient. However, they were tested on rather simple datasets, with just two salient attributes. Studying the effectiveness of our method on more complex scenarios, with multiple/correlated attributes, is an interesting research avenue.

A key challenge in StyleGAN-based manipulation is the distortion–editability trade-off~\cite{tov2021designing}: low-dimensional spaces ($W$, $W^+$) are well regularized and easy to edit but lose fine detail, whereas high-dimensional feature spaces ($F$) restore detail but risk entangling irrelevant attributes. Recent work therefore manipulates in $F$-space while conditioning generation in $W$~\cite{liu2023delving} or $W^+$~\cite{bobkov2024devil}. Following this strategy, we first train CS-StyleGAN in $W^+$ with strong regularization to obtain cleaner common/salient separation, and then refine image details in $F$-space. Unlike prior work~\cite{bobkov2024devil}, which conditions features using global semantic directions, our $F$-space refinement uses \emph{image-specific} latents and supports bidirectional training ($X\!\leftrightarrow\!Y$), enabling salient-factor swapping between images while maintaining high fidelity. A promising direction is to extend our approach to \emph{3D medical imaging} (CT/MRI), using volumetric backbones (e.g., 3D U-Net or StyleGAN) to separate \emph{common} healthy tissue from \emph{salient} pathology across volumes.

About Diffusion models, one important limitation is the computational cost, in particular compared to StyleGAN. Indeed, unlike StyleGAN’s single forward pass, DDIM requires a multi-step forward–reverse process (typically 50–100 steps; see \YL{Suppl., Table~XV} for per-image reconstruction times). In our experiments, naively applying adversarial losses in high-dimensional diffusion latents (e.g., $h$-space of size $512 \times 64$) often caused instability or collapse and yielded negligible gains. Moreover, tuning the associated hyperparameters and auxiliary networks becomes substantially time-consuming in a standard DDIM pipeline. A promising direction is to leverage recent high-order ODE solvers (e.g., DPM-Solver++~\cite{lu2025dpmsolverpp}) to reduce inference to only a few steps, thereby lowering the cost of regularization. Practically, one can also replace separate adversarial regressors with a gradient reversal layer (GRL) applied to intermediate representations to improve training stability. Finally, mining cross-dataset subspaces from Stable Diffusion attention maps \cite{liu2024towards}, or learning \textit{common}/\textit{salient} embeddings to guide text-to-image diffusion~\cite{miyake2025negative}, are complementary avenues for future work.

\section*{Acknowledgments}
This work was performed using HPC resources from GENCI–IDRIS (A0160615058) and it was supported by the EUR BERTIP (ANR‑18‑EURE‑0002).

\bibliographystyle{IEEEtran}
\bibliography{myrefs}


\clearpage

\makeatletter
\twocolumn[
\begin{@twocolumnfalse}

\begin{center}
{\LARGE\bfseries Supplementary Material\par}
\vspace{0.6em}
{\LARGE Learning Common and Salient Generative Factors Between Two Image Datasets\par}
\end{center}
\vspace{0.6em}
\end{@twocolumnfalse}
]
\makeatother

\setcounter{figure}{12}
\setcounter{table}{5}
\setcounter{equation}{28}
\setcounter{section}{0}

\section*{\textbf{Notation}}
All notation (e.g., $\mathcal{L}_{\text{lat}}$, $\|s_x\|_2^2$, etc.) follows the same definitions as in the main paper unless otherwise stated. We reuse the same notation throughout the Supplementary for consistency.

\section{Implementation details}
\label{sec:imp_details}
This section describes training and inference for \textbf{CS-StyleGAN} and \textbf{CS-Diffusion} (see \YL{Section~III} of the main paper for method details). All images are resized to $256 \times 256$ before being fed into the networks. At each training iteration, we sample two mini-batches of equal size, one from dataset $X$ and one from dataset $Y$, and the final losses are computed using both batches. All experiments are conducted on a single NVIDIA A100 GPU. Our code is available at \url{https://github.com/yunlongH/CA-with-stylegan2-pSp/}.

\subsection{Training and Inference of CS-StyleGAN}
The training of the proposed CS-StyleGAN consists of two stages. First, we train the separating network to learn common and salient factors in $\mathscr{S}$ on $W^+$ space. Second, we train the F-space refinement network to improve the image quality. In both stages, we use a fixed StyleGAN2 generator $G$ and pSp~\cite{richardson2021encoding} as the pre-trained encoder $E$.

\textit{\textbf{Stage 1.}} We first train $\mathscr{S}$ by minimizing $\mathcal{L}_{\text{lat}}$, $\|s_x\|^2_2$, and $\mathcal{L}_{\text{img}}$. After 130k steps, we use the trained $\mathscr{S}$ to produce latent factors $(c, s)$, which are then used as samples to train the networks $\mathcal{D}$ and $\mathcal{R}$ for 500 epochs by maximizing the losses $\mathcal{L}_{\text{adv-D}}$ and $\mathcal{L}_{\text{adv-R}}$. After this warm-up period, we train $\mathscr{S}$ by adversarially minimizing $\mathcal{L}_{\text{adv-D}}$ and $\mathcal{L}_{\text{adv-R}}$ (computed using the frozen $\mathcal{D}$ and $\mathcal{R}$), together with the other losses in \YL{Eq.~(12)} of the main paper. In addition, every 5k steps we retrain $\mathcal{D}$ and $\mathcal{R}$ for 500 epochs to further refine their capabilities. We use a learning rate of 0.01 for training $\mathscr{S}$ without adversarial losses and 0.001 when training $\mathscr{S}$ with adversarial losses. Stage 1 is trained for 150k steps for $\mathscr{S}$ using the Adam optimizer. Empirically, we found the best hyperparameters to be $\lambda_4 = 0.01$, $\lambda_{\text{id}} = 0.4$, and $\lambda_{\text{lpips}} = 0.8$, while keeping all other $\lambda$ values at 1 to maintain balanced training.

\textit{\textbf{Stage 2.}} After training the separating network $\mathscr{S}$, we use the frozen pSp encoder $E$ and $\mathscr{S}$ to train the refinement framework for both reconstruction and swapping tasks, as illustrated in \YL{Fig.~2} of the main paper. Training minimizes $\mathcal{L}_{x \rightarrow y}$ and $\mathcal{L}_{y \rightarrow x}$ for F-space swapping, $\mathcal{L}_r$ for real image reconstruction, and $\mathcal{L}_{\text{adv-}y}$ to enhance realism. For the computation of $\mathcal{L}_{adv}$ losses, we employ a pre-trained StyleGAN2 discriminator $D$, which is fine-tuned during training. We use $\lambda_{adv}=0.02$, $\lambda_{id}=0.2$, and $\lambda_{lpips}=0.8$, and optimize with a Ranger optimizer (learning rate 0.0002) for 200k training steps. 

\textit{\textbf{Inference.}}
Given real images $x$ and $y$, we first use \textbf{pSp-cs} (pSp encoder and separating network $\mathscr{S}$ pretrained after Stage~1) to encode them into common and salient factors: $(c_x,s_x)$ and $(c_y,s_y)$. Reconstructions use same-image factors, $G(c_x + s_x)$ and $G(c_y + s_y)$. Attribute exchange swaps the salient factor while keeping the common factor fixed, yielding $G(c_x + s_y)$ and $G(c_y + s_x)$. Once Stage~2 is trained, we compute the shifts $\Delta f_{x \rightarrow y}$ and $\Delta f_{y \rightarrow x}$ using the common/salient factors via \YL{Eqs.~(13) and (14)} of the main paper. Let $f_x = F_\text{Adpt}(E_f(x))$ and $f_y = F_\text{Adpt}(E_f(y))$ denote the F-space features from real images, then reconstructions are $G(f_x)$ and $G(f_y)$. Swapped results are obtained via $G(\widehat{f}_{y \rightarrow x})$ and $G(\widehat{f}_{x \rightarrow y})$, where $\widehat{f}_{x \rightarrow y} = F_{\text{Adpt}}(f_x \oplus \Delta f_{x \rightarrow y})$ and $\widehat{f}_{y \rightarrow x} = F_{\text{Adpt}}(f_y \oplus \Delta f_{y \rightarrow x})$.

\subsection{Training and Inference of CS-Diffusion}
For the network $\mathscr{D}$ operating on \textbf{h-space}, real images $x$ and $y$ are first inverted to their noisy versions $x_T$ and $y_T$ via the DDIM forward process, and then denoised using the DDIM reverse process. At each reverse step $t$, we extract $h_x^t$ and $h_y^t$ (the 8th-layer output of the U-Net) and pass them through $\mathscr{D}$ to obtain common and salient factors $(c_x^t, s_x^t)$ and $(c_y^t, s_y^t)$. These yield the latent loss $\mathcal{L}_{\mathrm{lat}_t}$. In parallel, $(c_x^t, s_x^t)$ and $(c_y^t, s_y^t)$ are fed into the U-Net decoder to produce the noise prediction $\epsilon_t^{\theta}(x_t | c_x^t+s_x^t)$ and $\epsilon_t^{\theta}(y_t | c_y^t+s_y^t)$). These are then compared with $\epsilon_t^{\theta}(x_t | h_x^t)$ and $\epsilon_t^{\theta}(y_t | h_y^t)$) to compute the $\mathcal{L}_{\mathrm{P}_t\text{-}x}$ and $\mathcal{L}_{\mathrm{P}_t\text{-}y}$. Together with $\mathcal{L}_{\mathrm{lat}_t}$, this gives the losses in \YL{Eq.~(27)} of the main paper for updating $\mathscr{D}$. We set $\lambda_{id} = 0.2$ and $\lambda_{lpips} = 0.8$, with all other $\lambda$'s at $1$. Training uses Adam with a learning rate of $1\mathrm{e}{-}4$ for 14k steps. After training, test images $x$ and $y$ are inverted to $x_T$ and $y_T$. At each DDIM reverse step $t$, the middle-layer feature $h^t$ is decomposed by the pretrained $\mathscr{D}$ into $(c^t, s^t)$, enabling reconstruction and attribute swapping, e.g., for reconstruction of $x$, replace $h_x^t$ with  $(c_x^t, s_x^t)$ and compute $\epsilon_t^{\theta}(x_t \mid c_x^t+s_x^t)$; for swapping, use $\epsilon_t^{\theta}(x_t \mid c_x^t{+}s_y^t)$. The $y$-case is analogous ($s_y^t \leftrightarrow s_x^t$).

For the network $\mathscr{S}$ trained on \textbf{DiffAE}, we compute the latent losses in \YL{Eqs.~(1)--(3)} (main paper) by replacing $w_x, w_y$ with $z_{\mathrm{sem}\text{-}x}, z_{\mathrm{sem}\text{-}y}$ obtained from DiffAE’s encoder $E_{\mathrm{sem}}$. Besides, the decomposed $(c,s)$ from $z_{\mathrm{sem}}$ are then combined with $x_T$ and $y_T$ (produced by the encoder $E_{\mathrm{sto}}$, e.g., $E_{\mathrm{sto}}(x, z_{\mathrm{sem}}) \rightarrow x_T$) and fed to the DDIM decoder $\mathrm{Dec}$ to compute the image-space losses in \YL{Eq.~(28)} of the main paper. We set $\lambda_{id}{=}0.2$ and $\lambda_{lpips}{=}0.8$, keep all other $\lambda$ at $1$, and optimize $\mathscr{S}$ with Adam (learning rate $1\mathrm{e}{-}3$) for 20k steps. At inference, real images are mapped to $z_{\mathrm{sem}\text{-}x}$, $z_{\mathrm{sem}\text{-}y}$ and to $x_T$, $y_T$. Reconstruction uses $z_{\mathrm{sem}\text{-}x}{=}c_x{+}s_x$ (and analogously for $y$). Attribute exchange swaps the salient components: $z_{\mathrm{sem}\text{-}x}{=}c_x{+}s_y$ and $z_{\mathrm{sem}\text{-}y}{=}c_y{+}s_x$.

\section{Architecture Details}
\label{sec:arch_details}

\begin{figure*}[ht!]
\begin{center}
   \includegraphics[width=1.0\linewidth]{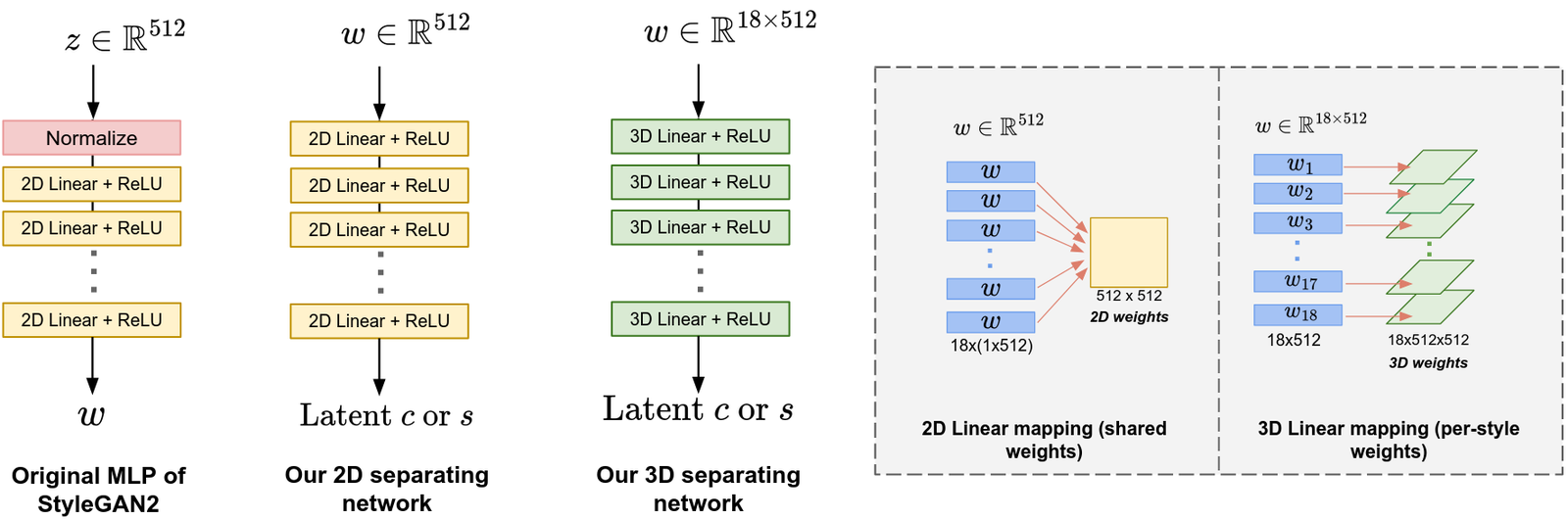}
   \caption{
   Comparison of the original StyleGAN2 mapping network and our proposed 2D and 3D separating networks.
   }
\label{fig:archi_detail_1}
\end{center}
\end{figure*}

\subsection{Separating Network}
Our separating network is built upon the MLP mapping architecture of StyleGAN2~\cite{karras2020analyzing}, which was originally designed to map latent codes from $\mathcal{Z}$ to the $W$ space. 
As shown in Fig.~\ref{fig:archi_detail_1}, we implement both 2D and 3D variants, where the 3D version extends the linear weights to enable per-style transformations. Specifically, in the original StyleGAN2 MLP, the latent code $z \in \mathbb{R}^{512}$ is mapped to a style vector $w \in \mathbb{R}^{512}$, 
which is then broadcast 18 times and injected into the corresponding modulation layers of the generator for image synthesis. Our 2D variant reuses the linear-layer architecture of the MLP used in StyleGAN2, where each transformation is expressed as:
\begin{align}
f(w^k) = w^k M^\top + b, \quad k = 1, 2, \ldots, d,
\end{align}
where $M \in \mathbb{R}^{512 \times 512}$ is the (2D) weight matrix of the fully connected layer, and $b \in \mathbb{R}^{512}$ is the bias vector. However, when using recent StyleGAN encoders such as pSp, the latents $w = [w_1, w_2, \ldots, w_{18}]$ consists of distinct style codes rather than repeated copies of a single $w$. In this case, an MLP-based network with a 2D weight matrix performs poorly, as the shared weights across all 18 style codes lack the expressive capacity to capture their style-specific variations. Thus, we extend $M$ to a 3D matrix $M^{*} \in \mathbb{R}^{18 \times 512 \times 512}$. 
The 3D weight multiplication is then formulated as:
\begin{align}
f(w^k) = w^k {M_k^*}^\top + b_k, \quad k = 1, 2, \ldots, d,
\end{align}
where each $M_k^*$ and $b_k$ correspond to an independent linear transformation for the $k$-th style code. 
This modification allows each style component of $w$ to be weighted individually, thereby better preserving the expressive capacity when mapping $w$ to $c$ and $s$.

We select the 2D or 3D weight module based on the dimensionality of the latent space. For CS-StyleGAN with the pSp encoder ($w \in \mathbb{R}^{18 \times 512}$), we adopt a 3D weight module with $M^{*} \in \mathbb{R}^{18 \times 512 \times 512}$ for $\mathscr{S}$. 
For DiffAE, we employ a 2D weight module with $M \in \mathbb{R}^{512 \times 512}$ for $\mathscr{S}$ to match the latent representation $z_{\mathrm{sem}} \in \mathbb{R}^{512}$. Additionally, a 3D weight module with $M^{*} \in \mathbb{R}^{64 \times 512 \times 512}$ is used for $\mathscr{D}$ 
to accommodate the higher-dimensional representation of the $h$-space.

For \textit{\textbf{background/target}} expriments, all separating networks consist of two independent branches, with each branch comprising $n$ layers of MLP—one branch outputs $c$ and the other outputs $s$. We set $n=12$ for $\mathscr{S}$ used in CS-StyleGAN, and $n=4$ for both the separating network $\mathscr{S}$ trained with DiffAE and the network $\mathscr{D}$ trained on $h$-space. 
For the \textit{\textbf{multiple-salients}} case, we used three independent branches for output $c$, $s1$, $s2$ (see architecture in \YL{Fig.~(3)} of the main paper), each branch comprises 12 layers of MLP with 3D-weight multiplication, using pSp as the $W^+$ space encoder.

\subsection{Architectures of Evaluation Classifiers}
To evaluate image editing performance, we train an image classifier on real images and apply it to detect target attributes in the edited images. Following TIME \cite{jeanneret2024text}, we use a DenseNet-121 classifier with the same architectural settings and training hyperparameters as reported in their work. We keep this classifier frozen during evaluation. The code of TIME is available at \url{https://github.com/guillaumejs2403/TIME/tree/main}. For human face images, we fine-tune an ImageNet-pretrained DenseNet-121 on our datasets for binary attribute classification (e.g., FFHQ glasses vs. non-glasses), whereas for non-face attributes (e.g., AFHQv2 cat vs. dog) we train the classifier from scratch on the corresponding datasets. To evaluate the quality of the latent separation, we train a linear classifier (logistic regression implemented in scikit-learn) on the learned common and salient factors; each latent factor is flattened and normalized before being fed to the classifier. 

\subsection{Regularization Networks \texorpdfstring{$\mathcal{D}$ and $\mathcal{R}$}{D and R}}
For the discriminator $\mathcal{D}$, we explored multiple architectures and empirically found that a simple neural network with one fully connected hidden layer, followed by ReLU, before mapping to the final output. We used a learning rate of $0.0001$ and a batch size of $4$ for training $\mathcal{D}$. The regressor network $\mathcal{R}$ follows the same architecture as $\mathscr{S}_c$ (or $\mathscr{S}_s$) but with a different depth: using a 10-layer MLP to produce a single output. We used a learning rate of $0.0002$ with a batch size of $4$ during training. 

\subsection{Other Architectures and Codes}
The original StyleGAN2 MLP implementation is available at \url{https://nn.labml.ai/gan/stylegan/index.html}. 
For the refinement networks, we used those proposed by SFE~\cite{bobkov2024devil}, including their F-space encoder ($E_f$), trainable adapter ($F_{\text{Adpt}}$), and pretrained StyleGAN2 generator $G$ and discriminator (for $\mathcal{L}_{adv}$). The SFE code is available at \url{https://github.com/ControlGenAI/StyleFeatureEditor}. 
For the CS-Diffusion networks, we followed the architectures of Asyrp and DiffAE, including their pretrained DDIM and DiffAE encoders, and they are available at \url{https://github.com/kwonminki/Asyrp_official} and \url{https://github.com/phizaz/diffae}.

\section{Ablation Studies}
\label{sec:ablation}
\subsection{Ablations on Network Architecture}
\begin{figure}[ht!]
\begin{center}
   \includegraphics[width=0.9\linewidth]{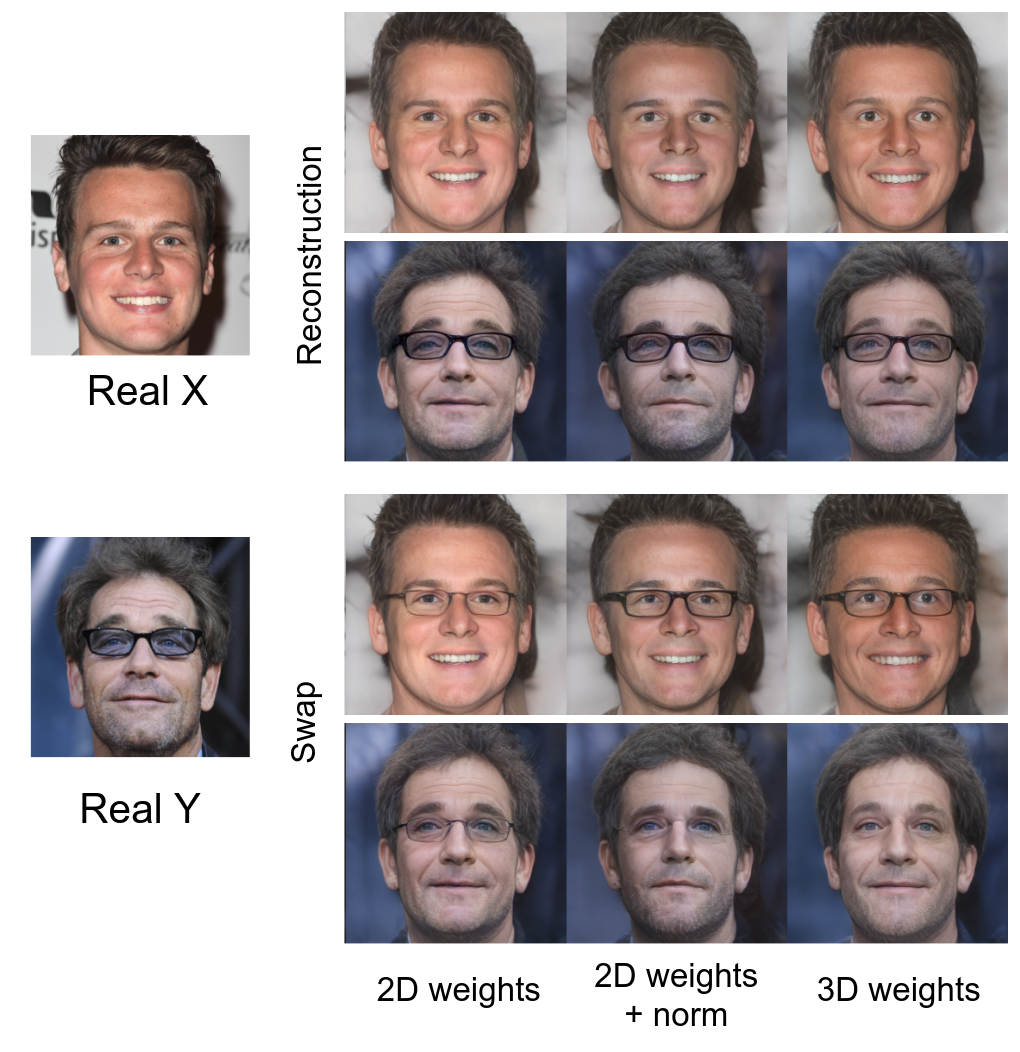}
   \caption{\textbf{Ablation study.} 2D vs. 3D Linear weight multiplication for reconstruction and attribute swapping.}
\label{fig:2d_vs_3d}
\end{center}
\end{figure}

\paragraph{2D network vs. 3D network}
Fig.~\ref{fig:2d_vs_3d} compares 2D (with/without input normalization) and 3D weight multiplication for reconstruction and attribute swapping in the $W^+$ space of StyleGAN (see Fig.~\ref{fig:archi_detail_1} for architecture details). Table~\ref{tab:Fid_2d_3d} reports their FID scores for reconstruction and for generations after swapping, along with the FID when applying 2D weight multiplication in the $W$ space. Table~\ref{tab:classification_results} shows their latent separation ability. These results demonstrate that when training the separating network $\mathscr{S}$ in the $W^+$ space, 3D weight multiplication yields superior performance for both reconstruction and attribute manipulation.


\begin{table}[ht!]
\caption{\textbf{Ablation study.} FID Scores of our spearating network on both $\mathcal{W}$ and $\mathcal{W}+$ space using 2D, normalized 2D or 3D weights. First two rows indicate results for reconstruction and last two rows for swapping.}
\label{tab:Fid_2d_3d}
\centering
\resizebox{0.9\linewidth}{!}{
\begin{tabular}{@{}c|c|ccc@{}}
\toprule
\multirow{2}{*}{} & $\mathcal{W}$& \multicolumn{3}{c}{$\mathcal{W}+$} \\ \cmidrule(lr){1-5}  
 & 2D & 2D & 2D (norm) & 3D\\ \cmidrule(r){1-5}
$G(c_x+s_x)$ & 85.59 & 45.57 & 42.19 & \textbf{41.56} \\
$G(c_y+s_y)$ & 74.78 & 36.96 & 34.21 & \textbf{33.50} \\ \cmidrule(r){1-5}
$G(c_x+s_y)$ & 75.42 & 53.33 & 55.88 & \textbf{49.15} \\
$G(c_y+s_x)$ & 97.98 & 63.01 & 63.61 & \textbf{54.58} \\ \bottomrule
\end{tabular}
}
\end{table}

\begin{table}[ht!]
\caption{\textbf{Ablation study.} Accuracy on test set of FFHQ using $X$: no glasses and $Y$: with glasses. We train latent classifier (logistic regression) on $c$ and $s$ for distinguishing No Glasses vs. Glasses or Male vs. Female. Glass information should be encoded in $S$ and gender information in $C$.}
\label{tab:classification_results}
\centering
\resizebox{0.9\linewidth}{!}{
\begin{tabular}{@{}ccccc@{}}
\toprule
\multirow{2}{*}{Method} & \multicolumn{2}{c}{\makecell{No glasses vs. Glasses \\ }} & \multicolumn{2}{c}{\makecell{Male vs. \\ Female}} \\ 
\cmidrule(lr){2-3} \cmidrule(lr){4-5}
                 & $C$ $\downarrow$   & $S$ $\uparrow$   & $C$ $\uparrow$    & $S$ $\downarrow$   \\ 
\midrule
2D            & 0.95  & 0.98  & 0.89  & 0.72  \\
2D (norm)     & 0.82  & 0.98  & 0.88  & 0.71  \\
3D            & 0.75  & 0.98  & 0.88  & 0.69  \\
3D + $\mathcal{L}_{\text{adv-D}}$     & 0.60  & 0.98  & 0.88  & 0.70  \\
3D + $\mathcal{L}_{\text{adv-D}}$ + $\mathcal{L}_{\text{adv-R}}$ & \textbf{0.52}  & \textbf{0.98}  & \textbf{0.89}  & \textbf{0.68}  \\ 
\midrule
Expected & 0.5 & 1 &1 & 0.5\\
\bottomrule
\end{tabular}
}
\end{table}

\begin{table}[!ht]
\caption{\textbf{Ablation study.} Effect of using shared or independent branches for $c$ and $s$ in $\mathscr{S}$}
\label{tab:shared_branch}
\centering
\resizebox{1.0\linewidth}{!}{
\begin{tabular}{@{}cccccc@{}}
\toprule
\multicolumn{2}{c}{Layer Config.} & \multicolumn{2}{c}{Reconstruction} & \multicolumn{2}{c}{Swap} \\ 
\cmidrule(lr){1-2} \cmidrule(lr){3-4} \cmidrule(lr){5-6}
Shared & Independent & $G(c_x + s_x)$ & $G(c_y + s_y)$ & $G(c_y + s_x)$ & $G(c_x + s_y)$ \\ 
\midrule
12 & 0  & 45.61 & 39.72 & 72.33 & 77.72 \\  
10 & 2  & 42.64 & 36.94 & 69.03 & 67.78 \\  
8  & 4  & 42.26 & 36.17 & 66.53 & 70.92 \\  
6  & 6  & \textbf{40.48} & 35.69 & 63.28 & 65.81 \\  
4  & 8  & 41.32 & 35.66 & 61.61 & 60.94 \\  
2  & 10 & 40.70 & 34.51 & 55.60 & 51.54 \\  
0  & 12 & 41.56 & \textbf{33.50} & \textbf{54.58} & \textbf{49.15} \\  
\bottomrule
\end{tabular}
}
\end{table}
\begin{figure}[ht!]
\begin{center}
   \includegraphics[width=0.9\linewidth]{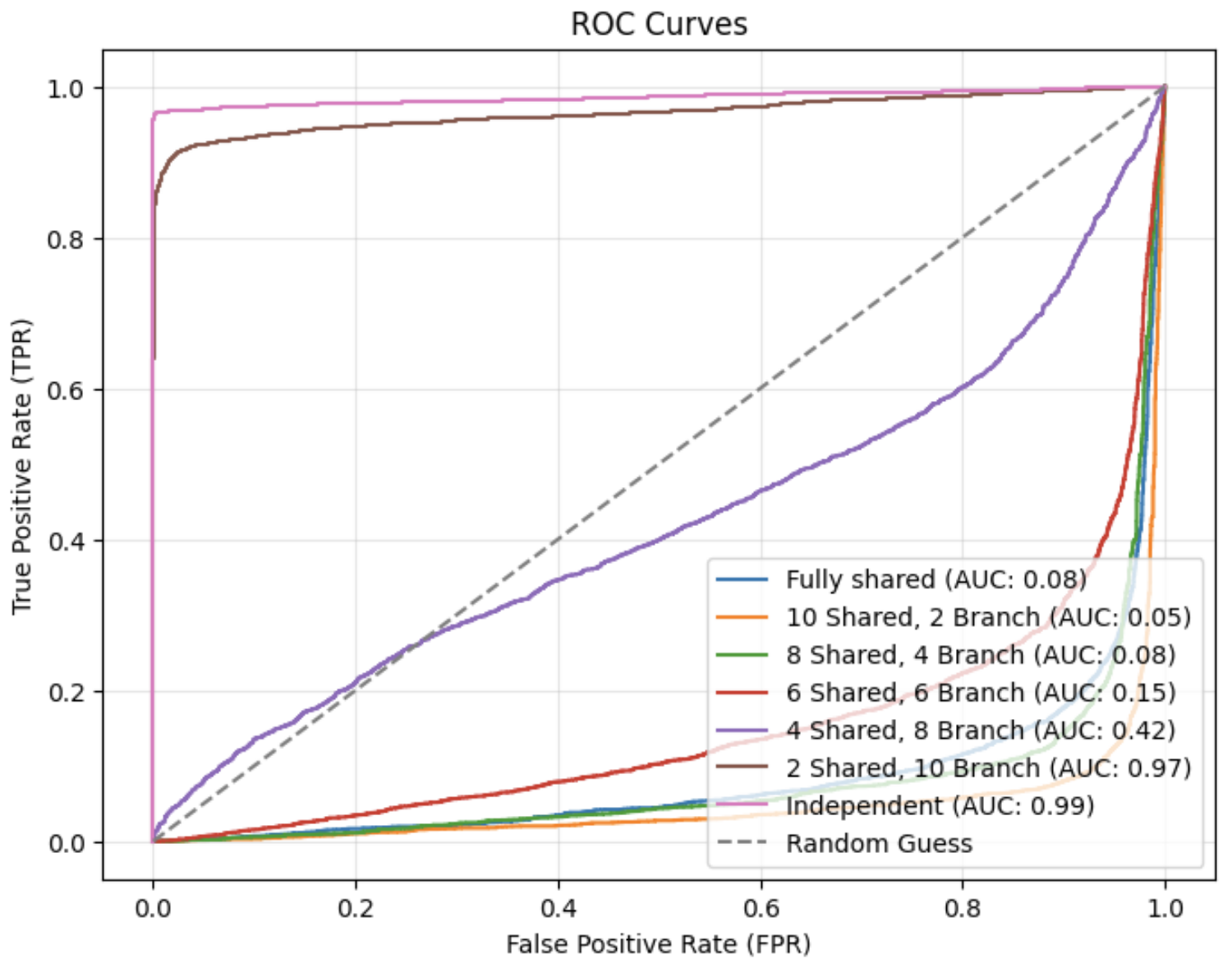}
   \caption{\textbf{Ablation study.} ROC curves for different configurations of network $\mathscr{S}$ on the classification (glasses vs. no glasses) of images obtained swapping the salient factors $s$ of reconstructed images from both $X$ and $Y$. We use a pretrained Grounding DINO as detection model.}
\label{fig:roc_shared}
\end{center}
\end{figure}

\begin{figure}[ht!]
\begin{center}
   \includegraphics[width=0.9\linewidth]{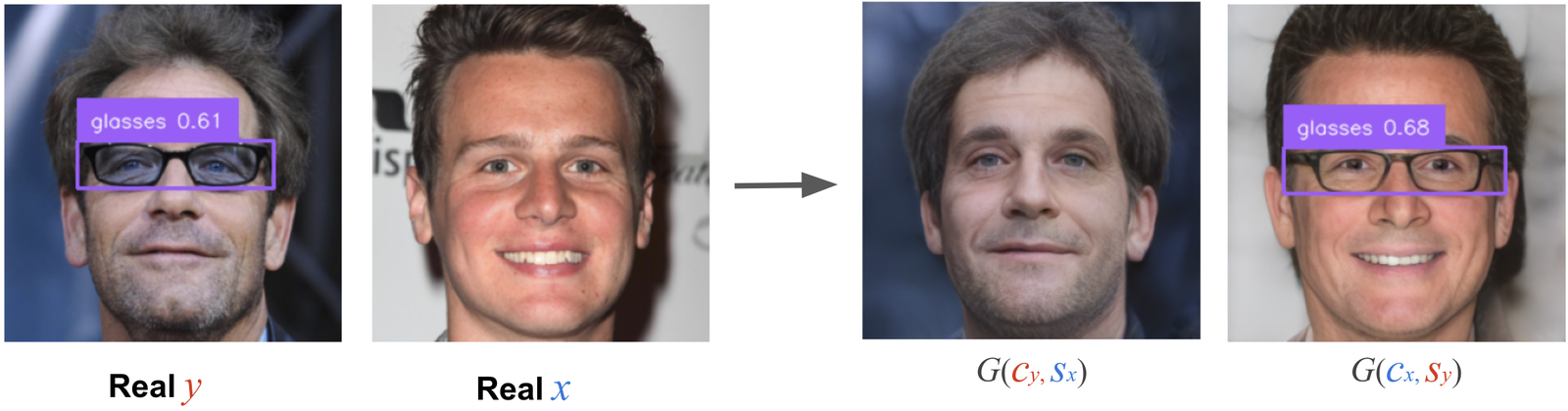}
   \caption{Grounding DINO detection confidences for salient-swap results on FFHQ.}
\label{fig:ffhq_gdino}
\end{center}
\end{figure}

\begin{figure}[ht!]
\begin{center}
   \includegraphics[width=0.9\linewidth]{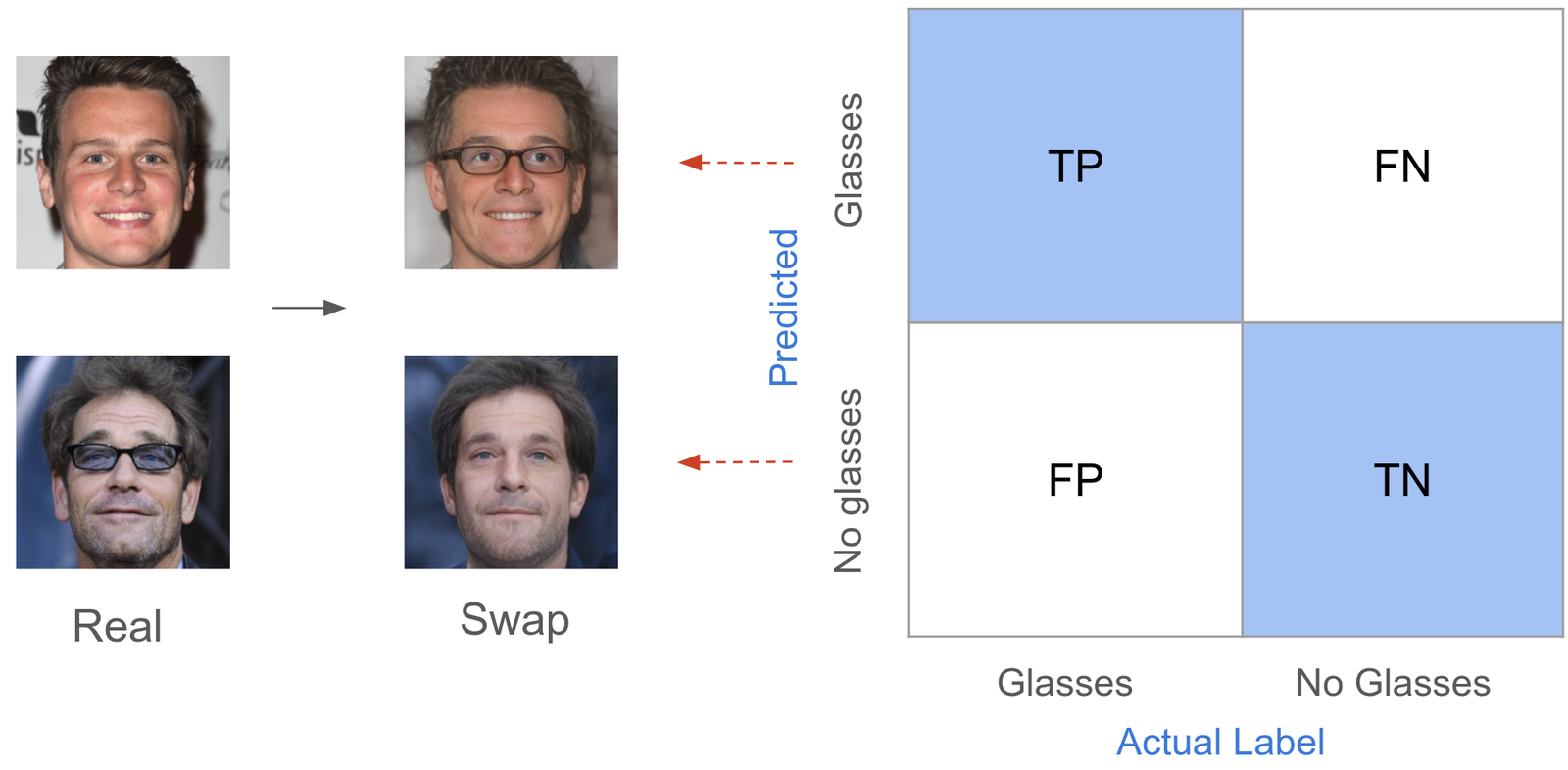}
   \caption{Evaluation metrics for classification on swapped image results}
\label{fig:cls_metrics}
\end{center}
\end{figure}

\paragraph{Effect of Shared vs.\ Independent Layers (Separating network)}
In Table~\ref{tab:shared_branch} and Fig.~\ref{fig:roc_shared}, we compare shared vs.\ independent layers between the $c$ and $s$ branches of the separating network $\mathscr{S}=\{\mathscr{S}_c,\mathscr{S}_s\}$. We vary the number of shared layers while keeping the total depth fixed (e.g., 12 layers). Table~\ref{tab:shared_branch} reports the FID scores for the reconstruction and swapping for different configurations. We further evaluate our method on synthesizing background and target images via salient-factor swapping. Here, we use a pretrained detection model, Grounding DINO \cite{liu2024grounding}, with the text input ``glasses" to obtain detection confidences. Fig.~\ref{fig:ffhq_gdino} shows an example of Grounding DINO \cite{liu2024grounding} detection applied to the salient swap results on FFHQ. In this context, see Fig.~\ref{fig:cls_metrics}, true positives (TP) correspond to images containing glasses that are correctly classified as 1; false negatives (FN) represent images with glasses that are misclassified as 0; false positives (FP) refer to images without glasses (0) that are incorrectly classified as 1; and true negatives (TN) denote images without glasses (0) that are correctly classified as 0. Based on this definition, we then label the swapped images, $G(c_x, s_y)$ and $G(c_y, s_x)$, as 0 and 1, respectively. The confidences serve as probabilities that can be classified using a simple threshold of 0.5. After classification, ROC can be computed, as shown in Fig.~\ref{fig:roc_shared}. We can notice that increasing the number of independent layers (branch) improves both the FID score and classification performance. This suggests that a fully independent network configuration is optimal for separation.


\begin{figure}[ht!]
\begin{center}
   \includegraphics[width=1.0\linewidth]{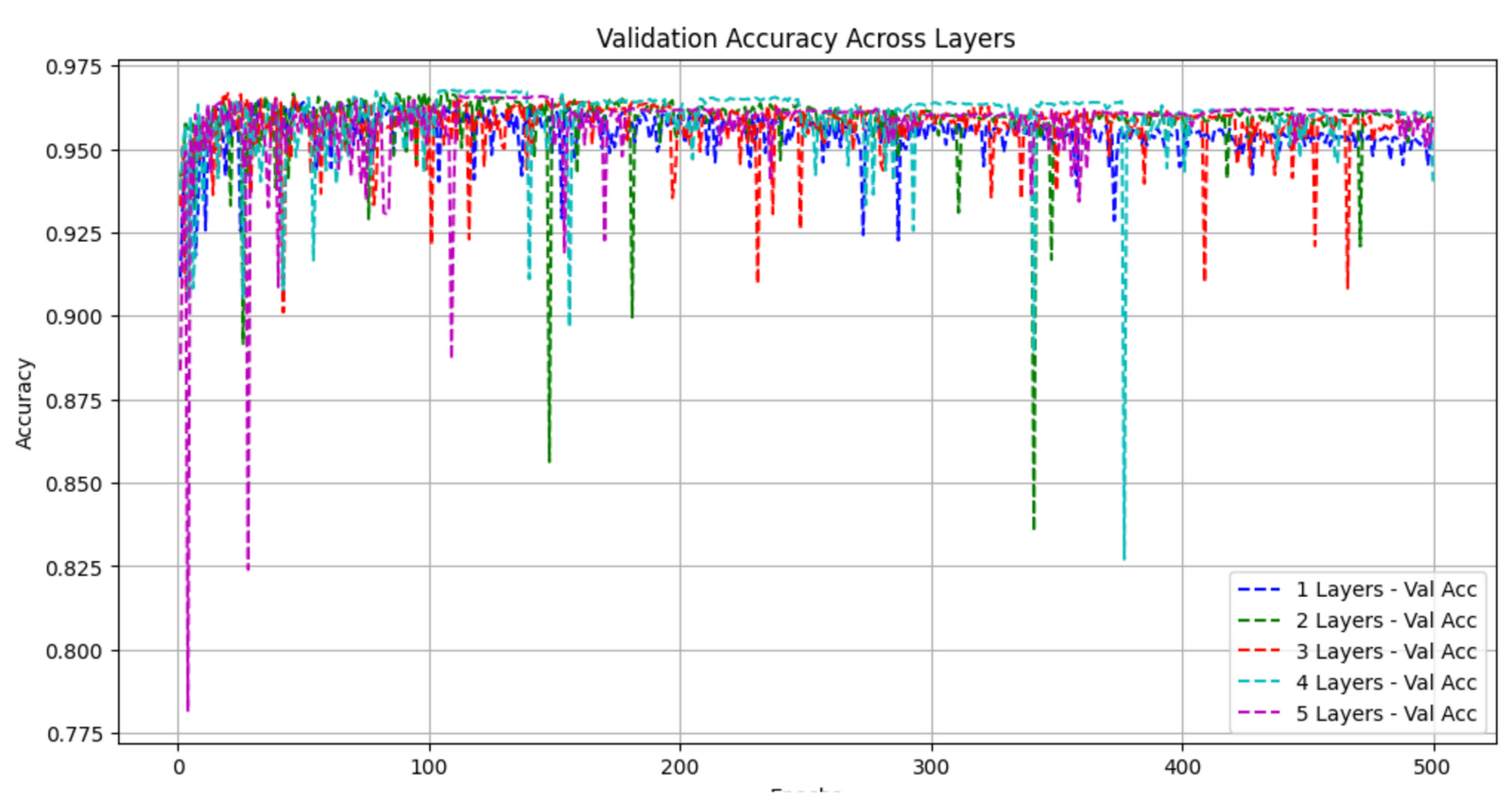}
   \caption{\textbf{Ablation study.} Effect of training $\mathcal{D}$ using different number of layers.}
\label{fig:D_acc}
\end{center}
\end{figure}

\paragraph{Discriminator Depth}
Fig.~\ref{fig:D_acc} reports the validation accuracy of discriminator $D$ when trained with different numbers of layers $\{1,2,3,4,5\}$ while keeping all other settings fixed. Across all depths, the model typically converges within 100 epochs, and a single layer already achieves competitive classification accuracy.

\subsection{Effect of the Choice of StyleGAN Inversion Encoder}
\begin{figure}[ht!]
    \centering
    \includegraphics[width=1.0\linewidth]{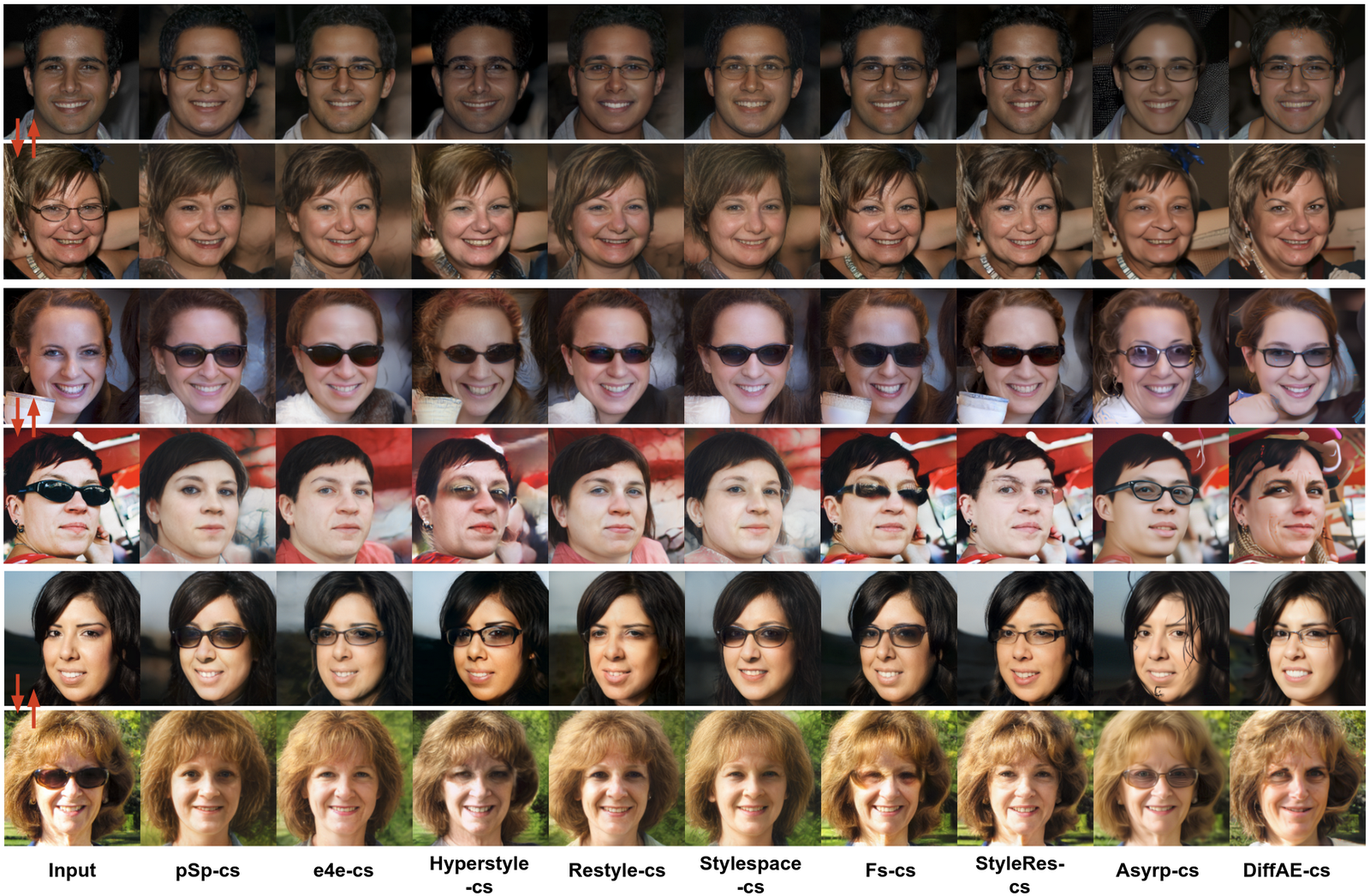}
    \caption{\textbf{Ablation study.} Results of our method applied on diverse StyleGAN-based encoders. Red arrows indicate attribute transfer between background and target images.}
    \label{fig:all_cs_applied}
\end{figure}

In Fig.~\ref{fig:all_cs_applied} and Table~\ref{tab:cs_sep_effect}, we present comparative results of our models applied to different StyleGAN-based encoders, covering latent spaces ranging from $W$, $W^+$, and $\mathcal{S}$ to $F$. In Fig.~\ref{fig:all_cs_applied}, we observe that results from the $W^+$ space (e.g., pSp-cs, e4e-cs) achieve clean attribute swapping (such as glasses) better then the others, albeit with slightly lower reconstruction quality. The values in Table~\ref{tab:cs_sep_effect} show that pSp-cs achieves the best overall separation results among the evaluated encoders. These findings suggest that pSp, combined with our refined framework, is the most suitable choice for both reconstruction and attribute swapping tasks.

\begin{table*}[ht!]
\caption{\textbf{Ablation study.} Results of common and salient separation for various pretrained encoders and latent spaces. The metric $\Delta$ quantifies separation quality, where $\Delta = |0.5 - C| + |1.0 - S|$ for glasses, and $\Delta = |1.0 - C| + |0.5 - S|$ for gender and smile. Lower $\Delta$ indicates better separation.}
\label{tab:cs_sep_effect}
\centering
\footnotesize
\resizebox{1.0\textwidth}{!}{
\begin{tabular}{l|l|ccc|ccc|ccc}
\toprule
\multicolumn{2}{c|}{\textbf{}} 
& \multicolumn{3}{c|}{\textbf{No Glasses vs. Glasses}} 
& \multicolumn{3}{c|}{\textbf{Male vs. Female}} 
& \multicolumn{3}{c}{\textbf{Smile vs. Non-Smiling}} \\
\textbf{Model} & \textbf{Latent Space} 
& $C \downarrow$ & $S \uparrow$ & $\Delta \downarrow$ 
& $C \uparrow$ & $S \downarrow$ & $\Delta \downarrow$ 
& $C \uparrow$ & $S \downarrow$ & $\Delta \downarrow$ \\
\midrule
Hyperstyle-cs        & W         & 0.686 & 0.985 & 0.201 & 0.810 & 0.516 & 0.206 & 0.846 & 0.500 & 0.154 \\
e4e-cs               & W$^+$     & 0.614 & 0.985 & 0.129 & 0.860 & 0.586 & 0.226 & 0.897 & 0.616 & 0.219 \\
pSp-cs               & W$^+$     & 0.516 & 0.980 & \textbf{0.036} & 0.804 & 0.514 & 0.210 & 0.903 & 0.523 & \textbf{0.120} \\
Restyle-e4e-cs       & W$^+$     & 0.622 & 0.986 & 0.136 & 0.836 & 0.570 & 0.234 & 0.878 & 0.539 & 0.161 \\
Restyle-pSp-cs       & W$^+$     & 0.671 & 0.979 & 0.192 & 0.836 & 0.662 & 0.326 & 0.884 & 0.566 & 0.182 \\
Stylespace-e4e-cs    & S         & 0.619 & 0.985 & 0.134 & 0.882 & 0.590 & 0.208 & 0.924 & 0.583 & 0.159 \\
Stylespace-pSp-cs    & S         & 0.543 & 0.981 & 0.038 & 0.917 & 0.597 & 0.086 & 0.917 & 0.598 & 0.081 \\
FS-cs                & W$^+$ \& F & 0.523 & 0.985 & 0.093 & 0.894 & 0.672 & 0.278 & 0.909 & 0.652 & 0.253 \\
StyleRes-cs          & W$^+$ \& F & 0.586 & 0.980 & 0.066 & 0.867 & 0.527 & \textbf{0.160} & 0.905 & 0.595 & 0.190 \\
\midrule
\multicolumn{2}{c|}{\textcolor{blue}{Expected}}
& \textcolor{blue}{0.5} & \textcolor{blue}{1.0} & \textcolor{blue}{0} 
& \textcolor{blue}{1.0} & \textcolor{blue}{0.5} & \textcolor{blue}{0} 
& \textcolor{blue}{1.0} & \textcolor{blue}{0.5} & \textcolor{blue}{0} \\

\bottomrule
\end{tabular}
}
\end{table*}

\subsection{\texorpdfstring{Ablation study about $s_x$}{Ablation study about s\_x}}
\begin{figure}[ht!]
\begin{center}
   \includegraphics[width=1.0\linewidth]{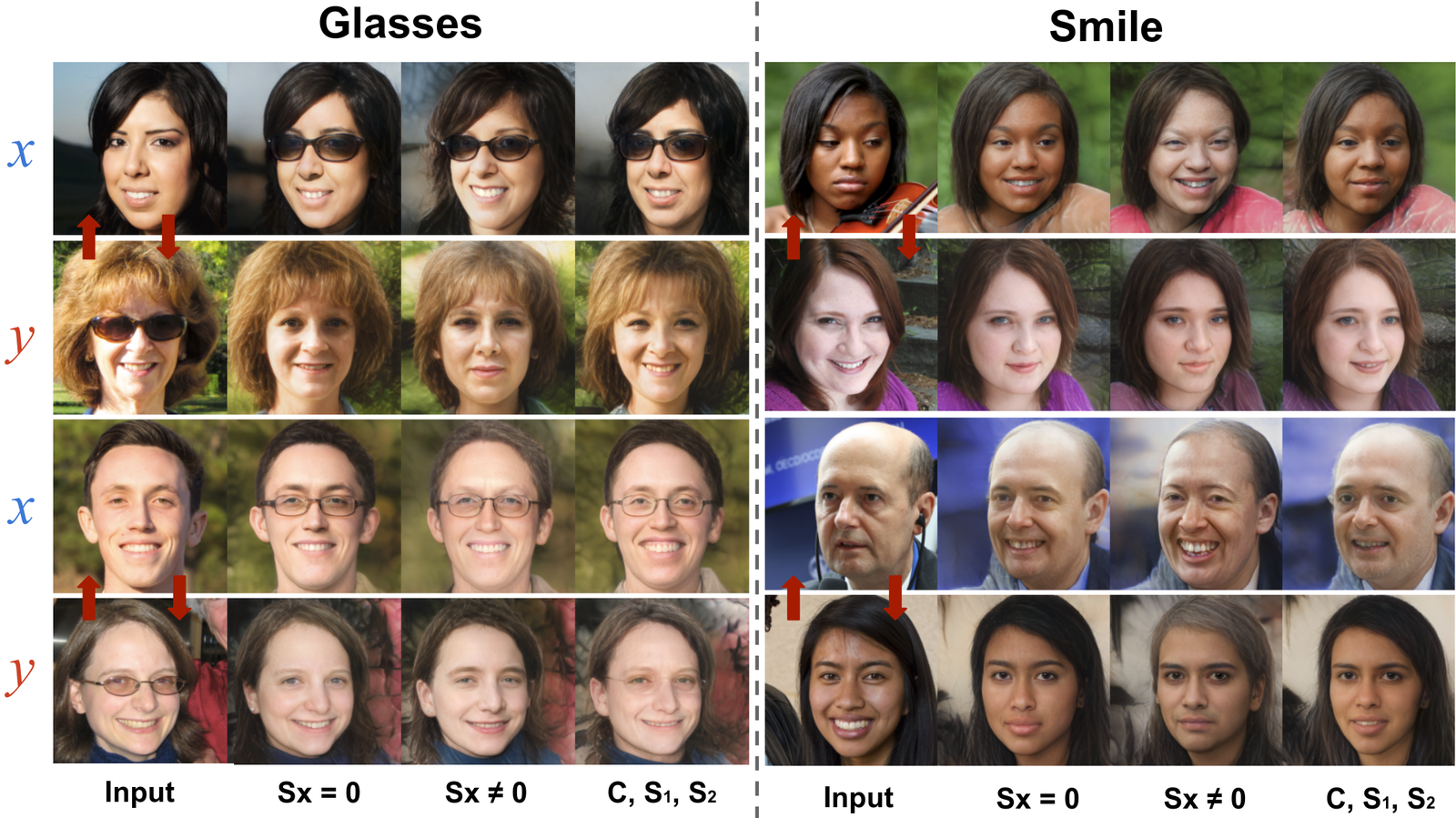}
   \caption{\textbf{Ablation study.} Impact on the swap results for different choices about $s_x$ using the \textit{background/target assumption}.}
\label{fig:sx_s1s2_ablation}
\end{center}
\end{figure}

In Fig.~\ref{fig:sx_s1s2_ablation}, we compare our model using three strategies for the salient output: (1) using $s_x$, (2) not using $s_x$ (i.e., setting $s_x$ to zero) in \YL{Eqs.(1), (2), and (11)} of the main paper during reconstruction training, and (3) using multiple-output saliency networks. The results show that for attributes involving added patterns, such as glasses and smiles, setting $s_x$ to zero yields the best performance. 

\subsection{F-space vs.\ \texorpdfstring{$W^+$}{W+}-space}
\begin{figure}[ht!]
    \centering    \includegraphics[width=1.0\linewidth]{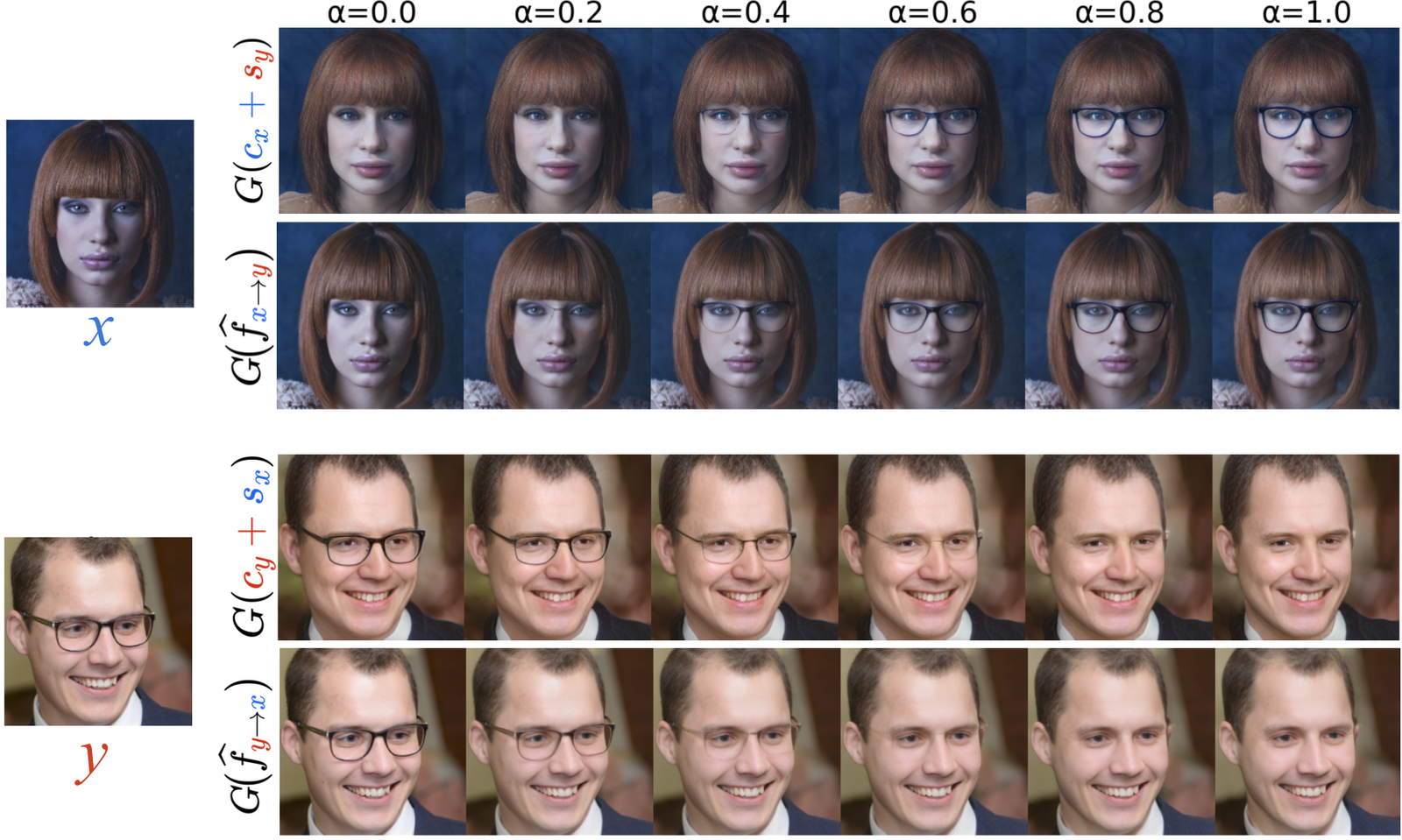}
    \caption{\textbf{Ablation study.} Comparison of latent interpolation in $W^+$ space and $F$ space.}
    \label{fig:interp_w_f}
\end{figure}

In Fig.\ref{fig:interp_w_f}, we compare the interpolation results of pSp-cs models in $W^+$ space with those of the pSp-cs-Ref models in F-space. The results demonstrate that F-space provides better reconstruction details than $W^+$ space. 

\subsection{Common/Salient Regularizations}
\label{sec:mutual_info}
Table~\ref{tab:classification_results} compares our method trained with different latent-separation regularizations. The results demonstrate the effectiveness of incorporating regularizations $\mathcal{L}_{\mathrm{adv}\text{-}R}$ and $\mathcal{L}_{\mathrm{adv}\text{-}D}$. Additionally, we also evaluate replacement regularizers for $\mathcal{L}_{\mathrm{adv}\text{-}R}$ that explicitly decrease $I(c, s)$ to encourage common–salient independence. Implementation details are provided below.

\paragraph{Discriminator-Based Estimates}
To assess statistical dependence between the learned \emph{common} and \emph{salient} factors, \(c\) and \(s\), we follow \cite{louiset2024sepvae} to measure their mutual information (MI): \(I(c;s)\). Specifically, the MI is defined by: $I(c;s) \;=\; \mathrm{KL} (q(c,s) \,\|\, q(c)\,q(s))$, i.e., the KL divergence between the joint \(q(c,s)\) and the product of marginals \(q(c)q(s)\). For the joint, we draw samples by taking a batch of image pairs \((x_i, y_i)\) and encoding them into latent pairs \([c_i, s_i]\), where \(i\) indexes items in the batch. To approximate the product of marginals, we use the same batch but shuffle the salient factors among images (e.g., \([c_1, s_2], [c_2, s_3],\) etc.).
After obtaining these samples, we train a discriminator \(D_{\mathrm{mi}}([c,s])\) to distinguish joint pairs from shuffled pairs using a BCE loss. 
Once the \(D_{\mathrm{mi}}\) is trained, we estimate the MI as:
\begin{align}
I(c;s)
&= \mathbb{E}_{q(c,s)}\!\left[
    \log \frac{q(c,s)}{q(c)q(s)}
\right] \notag\\
&\approx \frac{1}{N}\sum_{i=1}^{N}
\operatorname{ReLU}\!\left(
    \log \frac{D_{\mathrm{mi}}([c_i,s_i])}{1 - D_{\mathrm{mi}}([c_i,s_i])}
\right),
\label{eq:mi_est}
\end{align}
where the \(\operatorname{ReLU}\) ensures a non-negative KL estimate. We first trained \(D_{\mathrm{mi}}\) on latent factors produced by the our separating network \(\mathscr{S}\) trained without the regularizations $\mathcal{L}_{\mathrm{adv}\text{-}R}$ and $\mathcal{L}_{\mathrm{adv}\text{-}D}$. This yielded MI estimates \(I(c_x;s_x)=2.017\) and \(I(c_y;s_y)=2.036\) for background and target images, respectively. After training \(\mathscr{S}\) with the proposed regularizations, the MI dropped to \(I(c_x;s_x)=8\times10^{-4}\) and \(I(c_y;s_y)=1\times10^{-4}\). This substantial reduction indicates that our method effectively enforces independence between the common and salient factors in both domains.

\begin{table*}[!ht]
\centering
\caption{Quantitative comparison of mutual-information regularizations.}
\label{tab:mi_reg}
\resizebox{\textwidth}{!}{%
\begin{tabular}{l
                *{5}{c}
                *{5}{c}
                *{2}{c}
                *{2}{c}}
\toprule
& \multicolumn{5}{c}{\textbf{Reconstruction ($X$)}} & \multicolumn{5}{c}{\textbf{Reconstruction ($Y$)}}
& \multicolumn{2}{c}{\textbf{Swap X$\rightarrow$Y (+Glasses)}}
& \multicolumn{2}{c}{\textbf{Swap Y$\rightarrow$X (–Glasses)}} \\
\cmidrule(lr){2-6}\cmidrule(lr){7-11}\cmidrule(lr){12-13}\cmidrule(lr){14-15}
Regularizers &
LPIPS $\downarrow$ & MSE $\downarrow$ & MS-SSIM $\uparrow$ & ID-sim $\uparrow$ & FID $\downarrow$ &
LPIPS $\downarrow$ & MSE $\downarrow$ & MS-SSIM $\uparrow$ & ID-sim $\uparrow$ & FID $\downarrow$ &
ID-sim $\uparrow$ & FID-Y $\downarrow$ & ID-sim $\uparrow$ & FID-X $\downarrow$ \\
\midrule
W/o MI Reg. &
0.021 & 0.001 & 0.981 & 0.988 & 2.911 &
0.022 & 0.001 & 0.981 & 0.987 & 2.587 &
0.702 & 27.511 & 0.714 & 31.907 \\
kNN-MI &
0.022 & 0.001 & 0.982 & 0.987 & 2.607 &
0.022 & 0.001 & 0.981 & 0.990 & 2.442 &
0.715 & 27.897 & 0.729 & 30.734 \\
MINE &
0.020 & 0.001 & 0.982 & 0.991 & 2.534 &
0.021 & 0.001 & 0.982 & 0.989 & 2.421 &
0.801 & 25.884 & \textbf{0.820} & 27.320 \\
Disc-MI &
0.020 & 0.001 & 0.982 & 0.991 & 2.599 &
0.021 & 0.001 & 0.982 & 0.989 & 2.434 &
\textbf{0.806} & 25.511 & 0.811 & 28.907 \\
Regressor ($\mathcal{R}$) &
\textbf{0.019} & \textbf{0.001} & \textbf{0.983} & \textbf{0.992} & \textbf{2.406} &
\textbf{0.020} & \textbf{0.001} & \textbf{0.982} & \textbf{0.991} & \textbf{2.201} &
0.804 & \textbf{23.302} & 0.809 & \textbf{25.222} \\
\bottomrule
\end{tabular}%
}
\end{table*}

\paragraph{MI-Based Regularization Variants}
In addition to our regressor-based regularizer \(\mathcal{L}_{\text{adv-R}}\), we evaluate three MI–driven alternatives that can be used in place of \(\mathcal{L}_{\text{adv-R}}\) to promote independence between \(c\) and \(s\):
(1) Disc-MI, i.e., the discriminator-based estimates introduced above (Eq.~\eqref{eq:mi_est});
(2) kNN-MI~\cite{kraskov2004estimating}; and
(3) MINE~\cite{belghazi2018mutual}.
For Disc-MI, we first train the discriminator \(D_{\mathrm{mi}}\) to distinguish joint pairs from shuffled pairs, and then train the separating network \(\mathscr{S}\) by \emph{minimizing} the estimate in Eq.~\eqref{eq:mi_est}. For kNN-MI, we use the nonparametric kNN estimator proposed in \cite{kraskov2004estimating}. In our case, we build joint samples by concatenating each pair \(z_i = [c_i \oplus s_i]\), and we take the marginals as the separate sets \(\{c_i\}\) and \(\{s_i\}\). We compute the \(k\)-NN radius \(\varepsilon_i\) as the L2 distance from \(z_i\) to its \(k\)-th nearest neighbor among \(\{z_j\}_{j\neq i}\). Using this same radius \(\varepsilon_i\), we look in each marginal space separately and count how many other c or s samples lie within L2 distance \(\varepsilon_i\): denote these per-sample counts by \(n_c\) around \(c_i\) and \(n_s\) around \(s_i\). The MI is then estimated as :  $\widehat I_{kNN} \!=\! \psi(k)\! - \!\tfrac{1}{N}\sum_i\big[\psi(n_{c}+1)\!+\!\psi(n_{s}+1)\big] \!+ \!\psi(N)$, where \(\psi(\cdot)\) is the digamma function and \(N\) is the batch size (see details in \cite{kraskov2004estimating}). We use the estimate $\widehat I$ as an additional loss and minimize it when training $\mathscr{S}$. MINE estimates mutual information via the Donsker-Varadhan (DV) lower bound. Given a minibatch, we use the empirical objective: $\widehat V(\theta)\!=\!\frac{1}{B}\sum_{i=1}^B T_\theta(c_i,s_i)\!-\!\log\!(\frac{1}{B}\sum_{i=1}^B e^{\,T_\theta(c_i\tilde s_i)})$, where \(T_\theta\) is a neural critic, \((c_i,s_i)\) are joint samples from \(q(c,s)\), and \((c_i,\tilde s_i)\) are product-of-marginals samples from \(q(c)\,q(s)\) obtained by shuffling \(s\) within the batch; \(B\) is the batch size. We first maximize \(\widehat V(\theta)\) with respect to \(\theta\) to obtain a tight DV lower bound on \(I(c;s)\). Then, with \(T_\theta\) fixed, we train the separating network \(\mathscr{S}\) to minimize \(\widehat V(\theta)\), thereby reducing the mutual information between \(c\) and \(s\). 

Table~\ref{tab:mi_reg} presents the corresponding comparisons for reconstruction and swapped-image quality. Table~\ref{tab:cs_small} reports quantitative latent-separation performance of pSp-cs-Ref under the four regularization schemes. Overall, the regressor-based regularization achieves the best results across all metrics.

\begin{table}[ht!]
\centering
\caption{Effect of latent separation with MI regularizations. $X$ and $Y$ consist of facial images without glasses and with glasses.} 
\label{tab:cs_small}
\resizebox{0.9\linewidth}{!}{%
\begin{tabular}{l c c c}
\toprule
\textbf{Model} & \(\mathbf{C}\,\downarrow\) & \(\mathbf{S}\,\uparrow\) &
\(\boldsymbol{\Delta = |0.5{-}C| + |1.0{-}S|}\) \\
\midrule
W/o MI Reg.     & 0.60 & 0.98 & 0.12 \\
kNN-MI             & 0.66 & 0.98 & 0.18 \\
MINE            & 0.55 & 0.98 & 0.07 \\
Disc-MI     & 0.56 & 0.98 & 0.08 \\
Regressor ($\mathcal{R}$) & \textbf{0.52} & \textbf{0.98} & \textbf{0.04} \\
\midrule
\textcolor{blue}{Expected} & \textcolor{blue}{0.5} & \textcolor{blue}{1.0} & \textcolor{blue}{0} \\
\bottomrule
\end{tabular}%
}
\end{table}

\begin{figure*}[ht!]
    \centering  \includegraphics[width=1.0\linewidth]{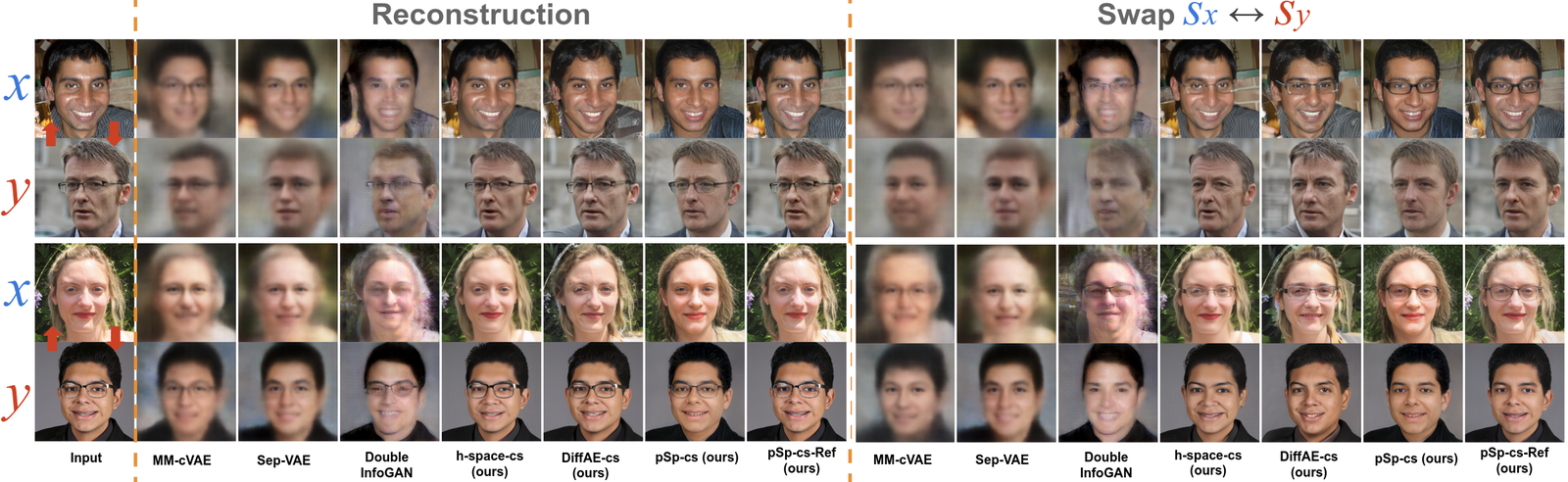}
    \caption{Visual comparisons of reconstructions and attribute-swapping results: our -cs models vs. CA baselines.}
    \label{fig:ca_sota_comparison_supp}
\end{figure*}

\section{Additional Results}
\label{sec:add_results}
Fig.~\ref{fig:ca_sota_comparison_supp} provides additional qualitative comparisons of our -cs models with contrastive analysis (CA) baselines, complementing the results shown in \YL{Fig.~4} of the main paper. Table~\ref{tab:recon_quality} reports quantitative reconstruction results, complementing the reconstruction-quality results shown in \YL{Fig.~5} of the main paper. 

\begin{figure*}[ht!]
\begin{center}
   \includegraphics[width=1.0\linewidth]{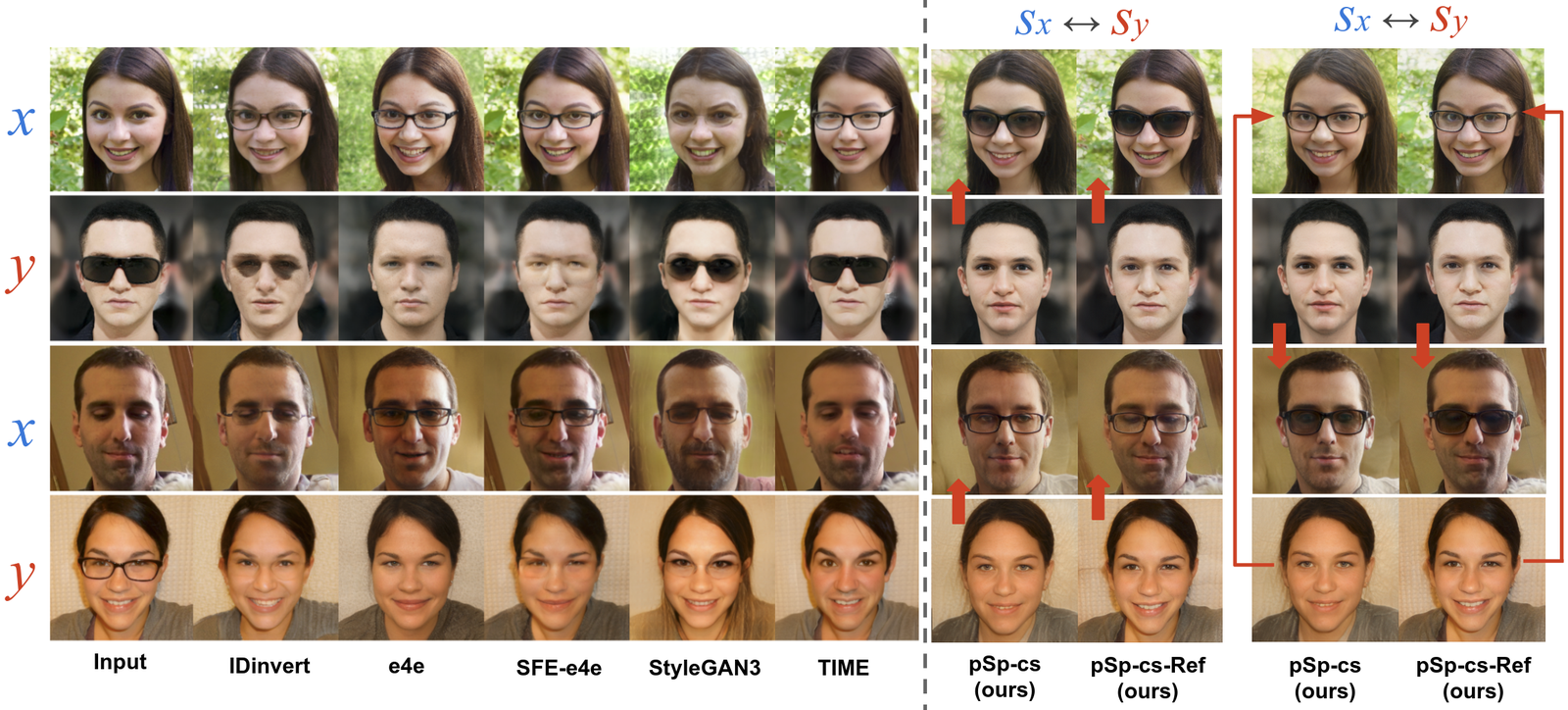}
    \caption{Comparison with SOTA methods for glasses editing. Columns~2–5: supervised baselines. Columns~6–10: weakly supervised approaches. Red arrows indicate transfer of \emph{glasses} from $y$ to $x$ via salient-factor swapping. Our method enables instance-specific swaps (e.g., transferring glasses shape and lens tint), whereas supervised baselines produce only global edits (e.g., simply adding or removing generic reading glasses).}
\label{fig:sota_editing_sunglasses}
\end{center}
\end{figure*}

\begin{figure*}[ht!]
\begin{center}
   \includegraphics[width=0.5\linewidth]{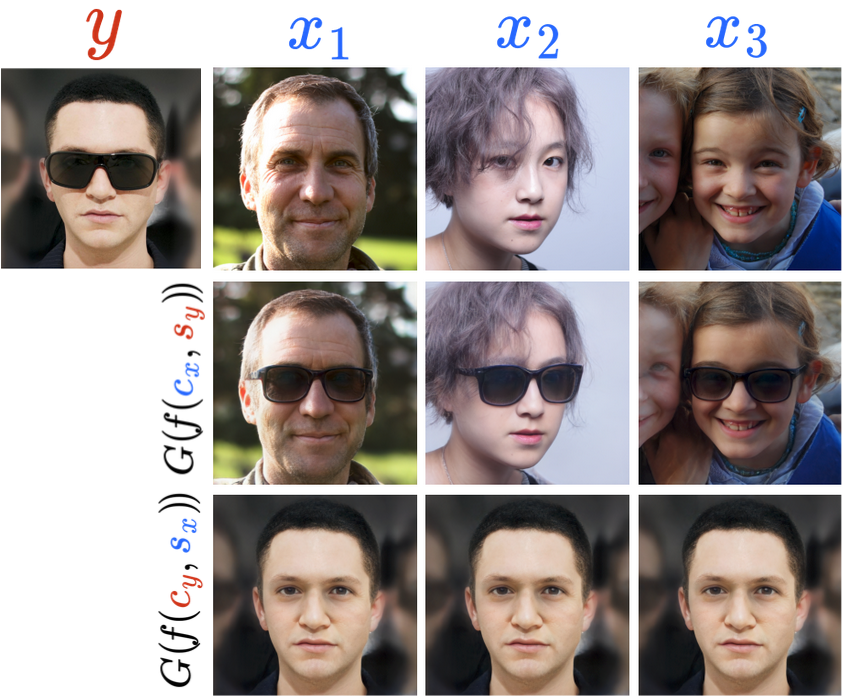}
   \caption{Faithful salient swap. Sunglasses are correctly encoded in $s_y$ and realistically added to the three background images  $x_1$, $x_2$, $x_3$.
   }
\label{fig:y_x1x2x3_sup}
\end{center}
\end{figure*}

\begin{figure*}[ht!]
    \centering
    \includegraphics[width=0.9\linewidth]{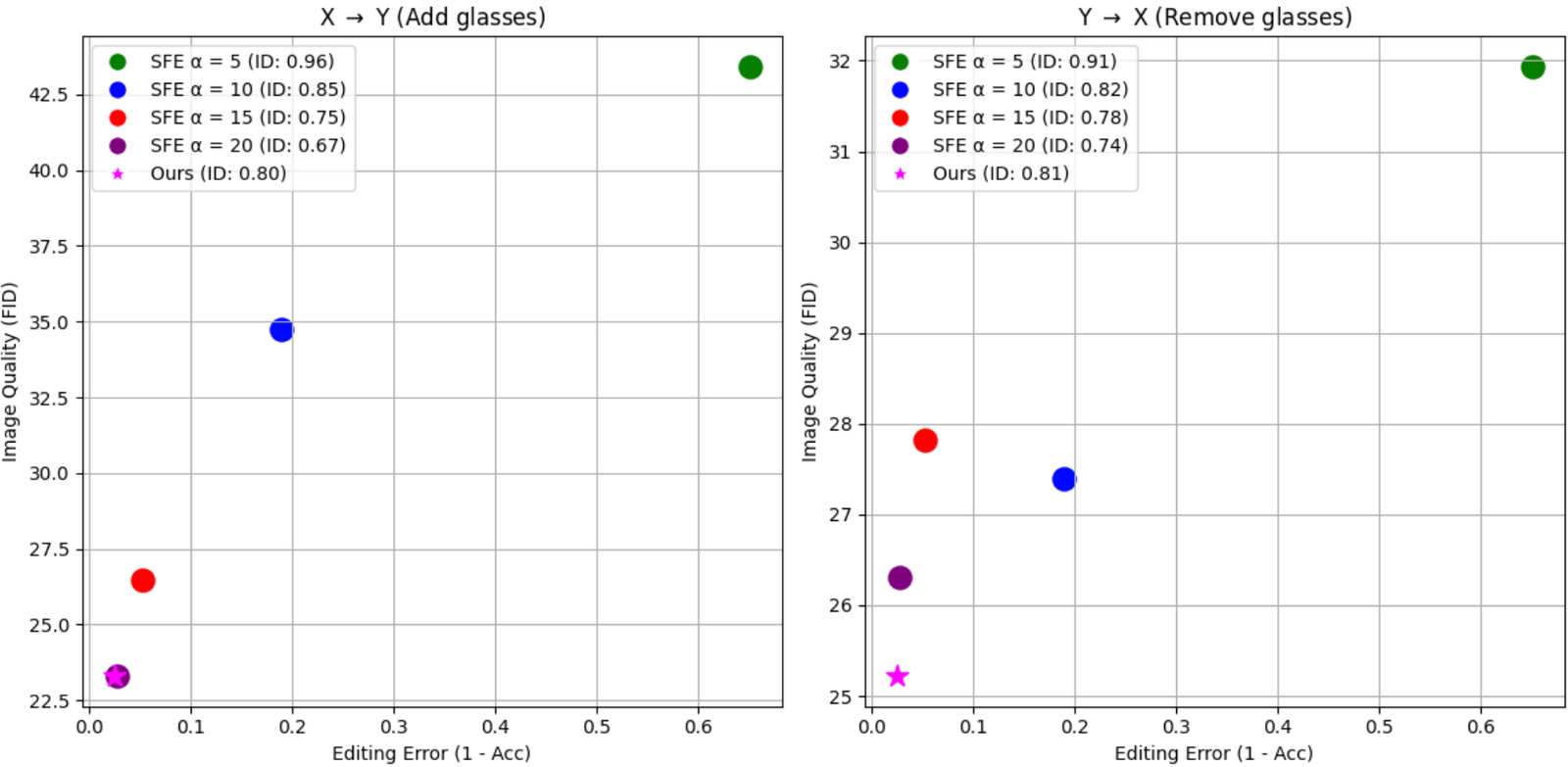}
    \caption{\textbf{Comparison of our method with SFE baseline.} SFE edits the image latent $\mathbf{w}$ by moving it along a global semantic direction $\mathbf{d}$ with strength $\alpha$ (i.e., $\hat{\mathbf{w}}=\mathbf{w}+\alpha\,\mathbf{d}$), whereas our method performs the edit in a single pass without $\alpha$ tuning. x-axis: editing error ($1-\mathrm{Acc}$, where $\mathrm{Acc}$ is the attribute-classification accuracy); y-axis: FID; ID: identity similarity.}
    \label{fig:balance_sfe_editing}
\end{figure*}

\begin{figure*}[ht!]
    \centering  \includegraphics[width=0.8\linewidth]{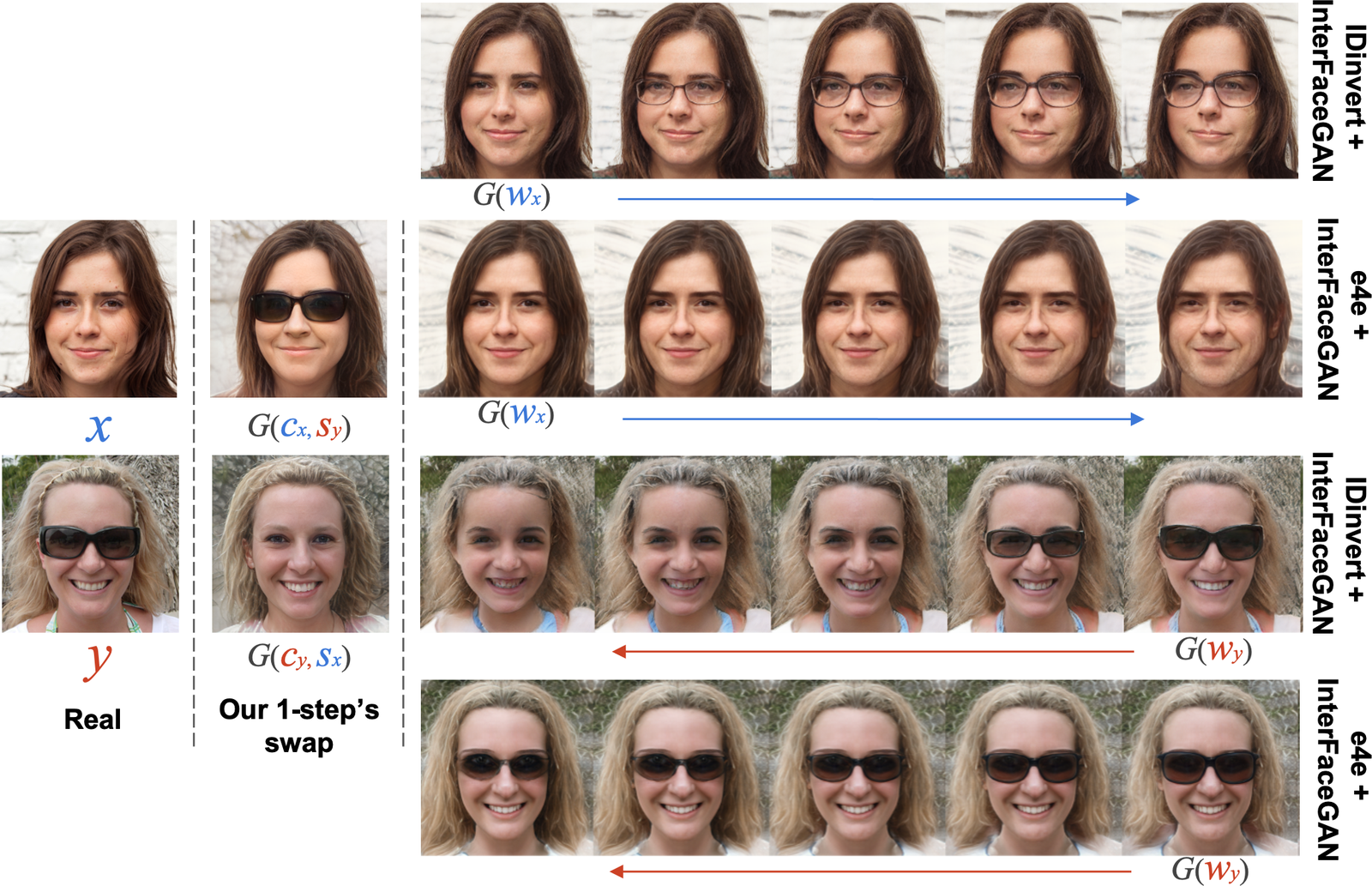}
    \caption{Comparison of our method with e4e and IDInvert with InterfaceGAN for glasses editing.}
    \label{fig:edit_idinvert}
\end{figure*}

Fig.~\ref{fig:sota_editing_sunglasses} provides additional results of our method compared with SOTA approaches for image editing, as in \YL{Fig.~6} of the main paper. Fig.~\ref{fig:y_x1x2x3_sup} shows an additional results for faithful salient swap as in \YL{Fig.~7} of the main paper.
Figs.~\ref{fig:balance_sfe_editing} and \ref{fig:edit_idinvert} provide additional results of our method compared with SFE, complementing the results described in \YL{Section~IV-D5} of the main paper. For Fig.~\ref{fig:edit_idinvert}, we first invert real images $x$ and $y$ into the \(W^+\) space using two advanced StyleGAN-based inversion methods: e4e~\cite{tov2021designing} (pretrained on FFHQ with a StyleGAN3 generator~\cite{alaluf2022third}) and IDInvert~\cite{zhu2020indomain} (pretrained on FFHQ with a StyleGAN1 generator~\cite{karras2019style}). After inversion, edits are applied with InterFaceGAN~\cite{shen2020interpreting}, which linearly shifts latent codes along an attribute direction: $w_{\text{edit}} \!=\! w \!+\! \alpha\mathbf{d}$, where \(\mathbf{d}\) is the semantic direction (estimated from a classifier’s decision boundary) and \(\alpha\) is an editing strength factor. Adding glasses uses a positive shift (\(\alpha>0\)), while removing glasses uses a negative shift (\(\alpha<0\)); we sweep \(\alpha \in \{1,2,3,4\}\). For both background and target images, we start from the reconstructions \(G(w_x)\) (or \(G(w_y)\)) and apply the edits. Results show that our method produces more realistic images and cleaner edits than either encoder combined with InterFaceGAN, while requiring only a single inference step.

\begin{table*}[ht!]
\caption{Quantitative comparison of reconstruction quality. Best results in bold, second-best underlined.}
\label{tab:recon_quality}
\centering
\footnotesize
\resizebox{1.0\textwidth}{!}{
\begin{tabular}{llcccccccccc}
\toprule
\multirow[b]{2}{*}{} & \multirow[b]{2}{*}{\textbf{Model}}
& \multicolumn{5}{c}{\textbf{X dataset (No glasses)}} 
& \multicolumn{5}{c}{\textbf{Y dataset (Glasses)}} \\
\cmidrule(lr){3-7}\cmidrule(lr){8-12}
 & & LPIPS ↓ & MSE ↓ & MS-SSIM ↑ & ID-sim ↑ & FID ↓ 
   & LPIPS ↓ & MSE ↓ & MS-SSIM ↑ & ID-sim ↑ & FID ↓ \\
\midrule

\multirow{11}{*}{\textbf{\shortstack[c]{StyleGAN \\ encoders}}}
& e4e & 0.191 & 0.013 & 0.717 & 0.803 & 41.828 & 0.194 & 0.014 & 0.706 & 0.783 & 35.158 \\
& pSp & 0.147 & 0.010 & 0.758 & 0.871 & 30.245 & 0.150 & 0.010 & 0.749 & 0.857 & 27.848 \\
& Hyperstyle & 0.086 & 0.007 & 0.832 & 0.944 & 22.278 & 0.090 & 0.007 & 0.828 & 0.937 & 20.321 \\
& ID-invert & 0.147 & 0.013 & 0.764 & 0.608 & 16.715 & 0.146 & 0.014 & 0.761 & 0.635 & 14.419 \\
& Restyle & 0.187 & 0.013 & 0.724 & 0.785 & 36.381 & 0.189 & 0.013 & 0.716 & 0.766 & 30.429 \\
& Stylespace & 0.163 & 0.011 & 0.745 & 0.867 & 28.946 & 0.165 & 0.012 & 0.692 & 0.852 & 26.171 \\
& FS & 0.069 & 0.005 & 0.894 & 0.924 & 12.695 & 0.076 & 0.006 & 0.885 & 0.911 & 11.775 \\
& StyleRes & 0.080 & 0.005 & 0.907 & 0.905 & 9.207 & 0.085 & 0.005 & 0.899 & 0.898 & 8.502 \\
& StyleGAN3-e4e & 0.223 & 0.019 & 0.671 & 0.710 & 41.008 & 0.228 & 0.020 & 0.654 & 0.722 & 34.192 \\
& StyleGAN3-pSp & 0.174 & 0.011 & 0.741 & 0.868 & 42.653 & 0.177 & 0.012 & 0.733 & 0.846 & 38.333 \\
& SFE & 0.023 & 0.001 &  \underline{0.986} &  \underline{0.994} & \underline{3.586} & 0.024 & 0.001 &  \underline{0.986} &  0.991 & \underline{3.148} \\

\midrule

\multirow{3}{*}{\textbf{\shortstack[c]{Diffusion \\ based}}}
& TIME & 0.025 & 0.001 & 0.981 & 0.988 & 8.551 & 0.026 & 0.001 & 0.982 & 0.986 & 8.640 \\
& DiffAE & \textbf{0.014} & \textbf{0.0001} & \textbf{0.994} & \textbf{0.996} & 9.257 & \textbf{0.014} & \textbf{0.0001} & \textbf{0.994} & \textbf{0.996} & 7.167 \\
& Asyrp & 0.134 & 0.002  & 0.941 & 0.941 & 21.347 & 0.146 & 0.002 & 0.938 & 0.937 & 20.697 \\

\midrule

\multirow{4}{*}{\textbf{\shortstack[c]{CS models \\ (ours)}}}
& DiffAE-cs & 0.188 & 0.011 & 0.831 & 0.732 & 29.816 & 0.118 & 0.006 & 0.907 & 0.877 & 16.018 \\
& H-space-cs & 0.179 & 0.004 & 0.908 & 0.838 & 30.776 & 0.190 & 0.004 & 0.907 & 0.840 & 29.271 \\
& pSp-cs & 0.196 & 0.015 & 0.692 & 0.756 & 41.709 & 0.183 & 0.014 & 0.704 & 0.790 & 33.095 \\
& pSp-cs-Ref & \underline{0.019} & 0.001 & 0.983 & 0.992 & \textbf{2.406} & \underline{0.020} & 0.001 & 0.982 &  \underline{0.991} & \textbf{2.201} \\
\bottomrule
\end{tabular}
}
\end{table*}

\begin{figure*}[ht!]
    \centering  \includegraphics[width=1.0\linewidth]{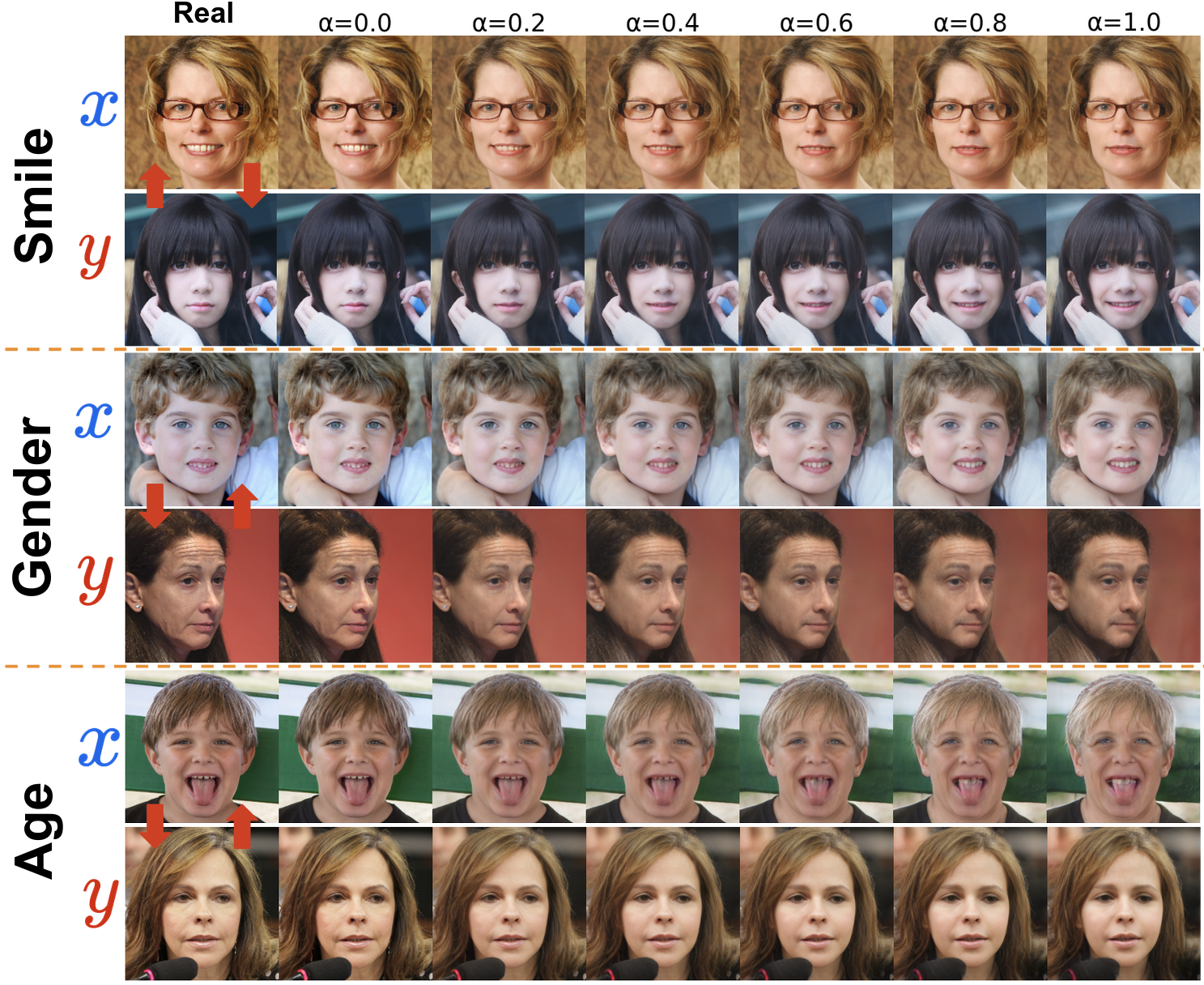}
    \caption{Interpolations between background and target images.}
    \label{fig:figure4_v2}
\end{figure*}

\begin{figure*}[ht!]
    \centering  \includegraphics[width=1.0\linewidth]{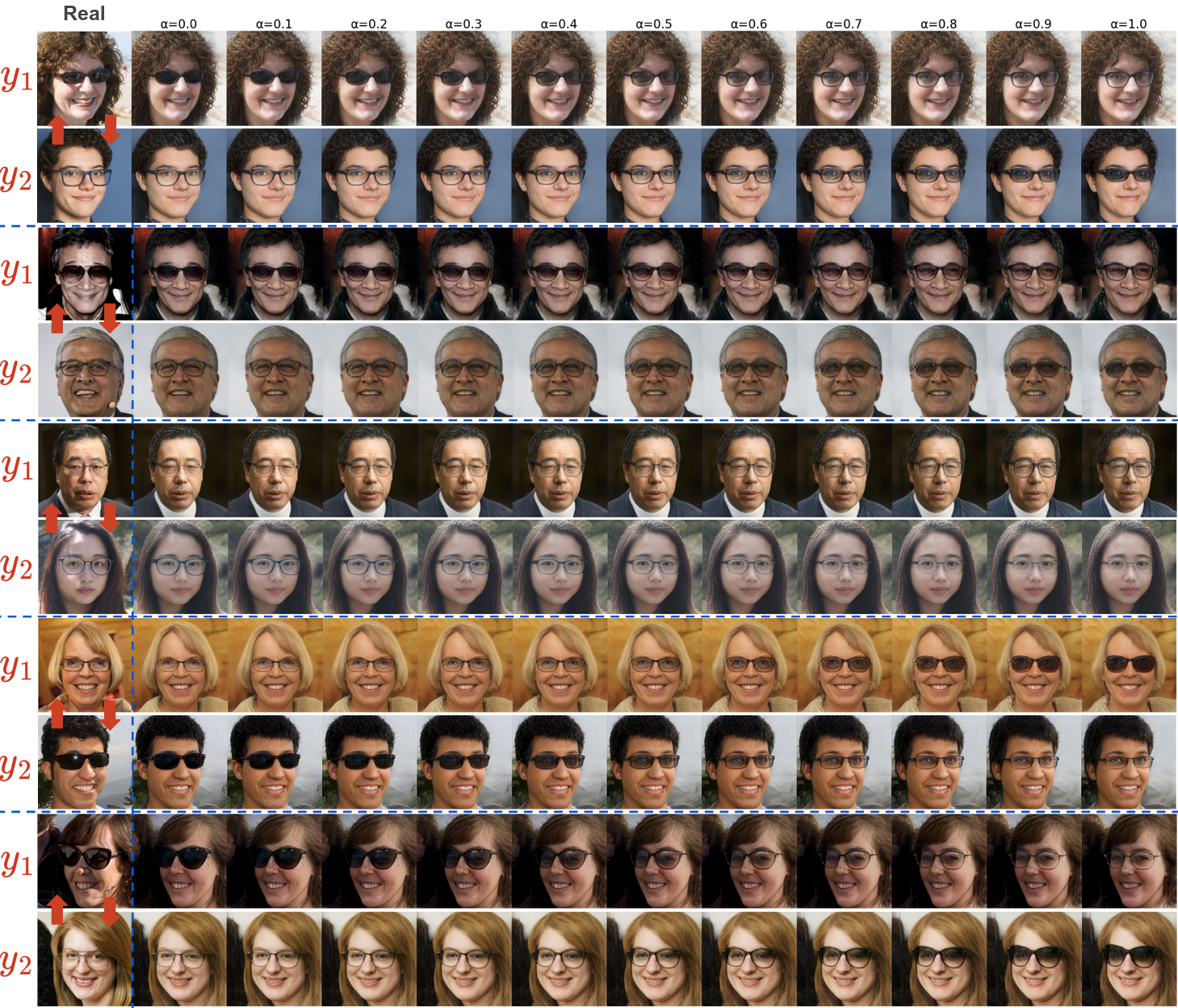}
    \caption{Additional results for interpolation along salient factors of target images.}
    \label{fig:interp_two_salient_supp}
\end{figure*}

\begin{figure*}[ht!]
\begin{center}
   \includegraphics[width=0.9\linewidth]{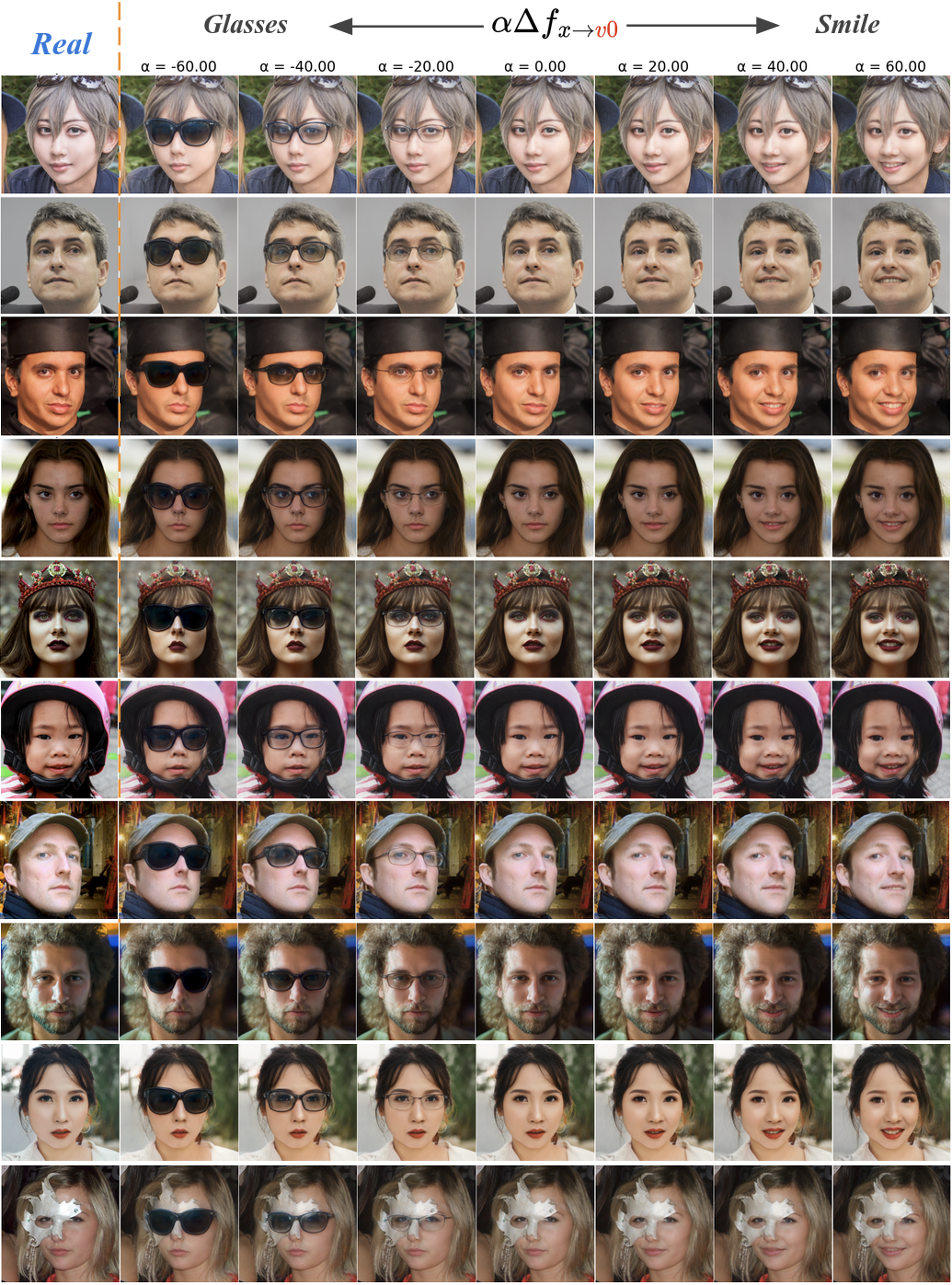}
\end{center}
   \caption{More results of semantic traversal along the \textbf{first} PCA direction $v_0$ of learned salient features.}
\label{fig:pca_1st_supp}
\end{figure*}

\begin{figure*}[ht!]
\begin{center}
   \includegraphics[width=0.9\linewidth]{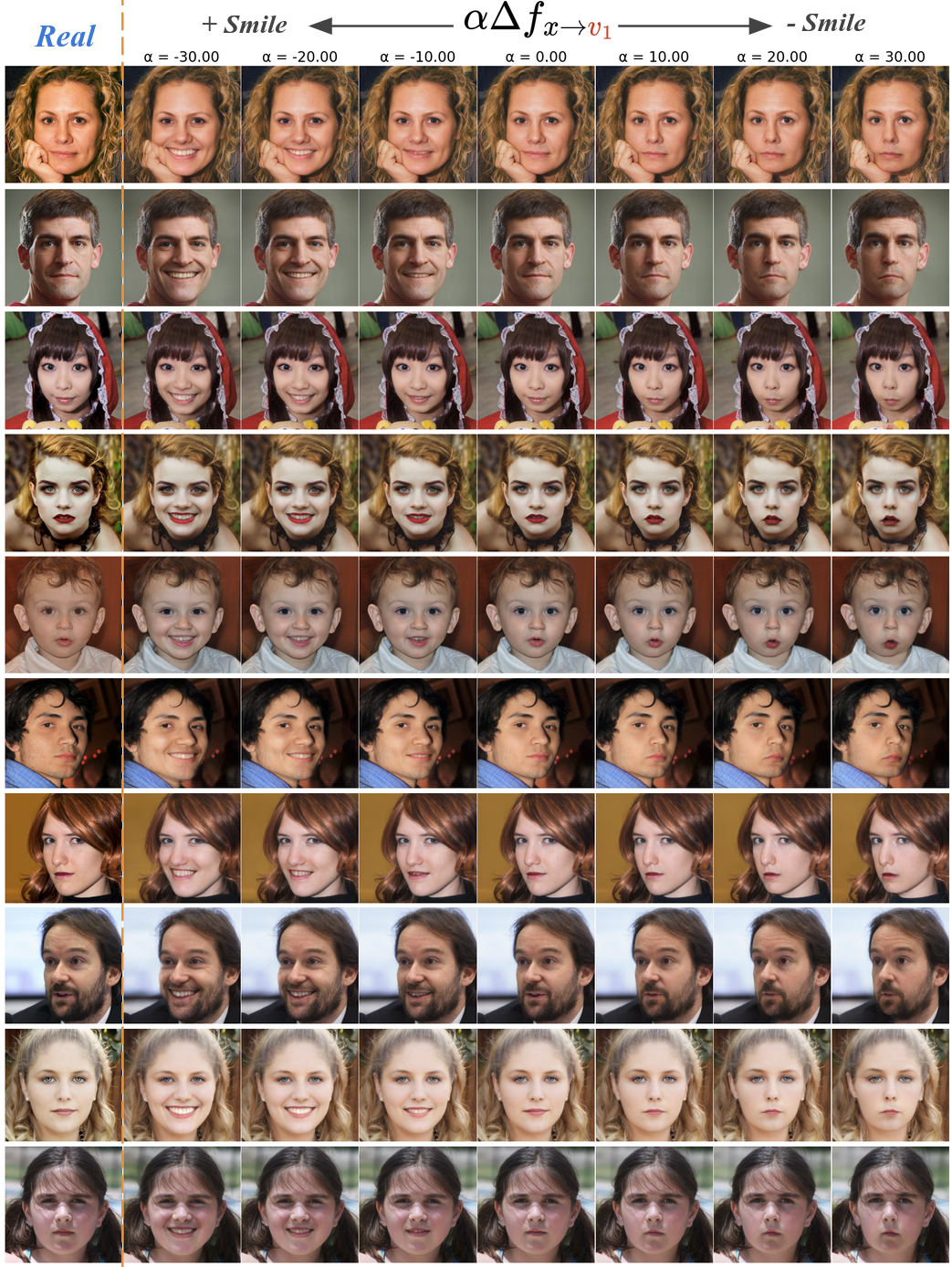}
\end{center}
   \caption{More results of semantic traversal along the \textbf{second} PCA direction $v_1$ of learned salient features.}
\label{fig:pca_2nd_supp}
\end{figure*}

\begin{figure*}[ht!]
\begin{center}
   \includegraphics[width=0.9\linewidth]{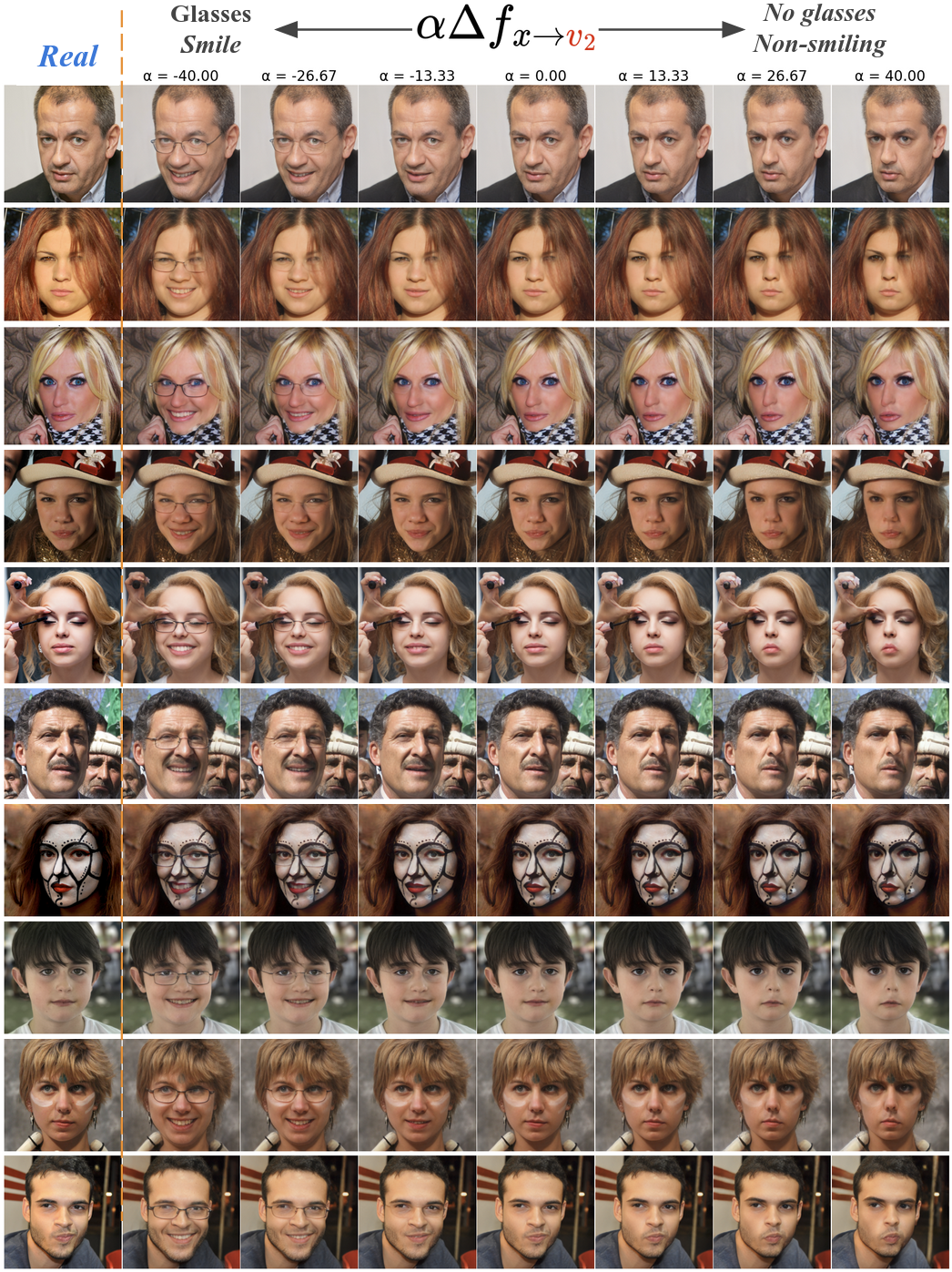}
\end{center}
   \caption{More results of semantic traversal along the \textbf{third} PCA direction $v_2$ of learned salient features.}
\label{fig:pca_3rd_supp}
\end{figure*}

Figs.~\ref{fig:figure4_v2}--\ref{fig:pca_3rd_supp} complement the results described in \YL{Section~IV-D6} of the main paper for interpolation along salient factors. Fig.~\ref{fig:figure4_v2} shows interpolations between background–target image pairs $x$ and $y$, while Fig.~\ref{fig:interp_two_salient_supp} shows interpolations between two target images $y_1$ and $y_2$. Figs.~\ref{fig:pca_1st_supp}, \ref{fig:pca_2nd_supp}, and \ref{fig:pca_3rd_supp} present additional results obtained by replacing the salient code $s_x$ of images from $X$ with traversals along the first ($v_0$), second ($v_1$), and third ($v_2$) PCA directions, where $v_k$ is estimated from the $Y$-salients that contain glasses and smile attributes absent in $X$. The results show that the first principal component (explaining $8.05\%$ of the variance) corresponds to a transition from glasses to smile. The second principal component (explaining $2.91\%$) reflects a change from smiling to non-smiling. 
The third principal component (explaining $2.66\%$) moves from faces exhibiting both glasses and a smile toward faces without glasses and not smiling.

\begin{figure*}[ht!]
\begin{center}
\includegraphics[width=0.9\linewidth]{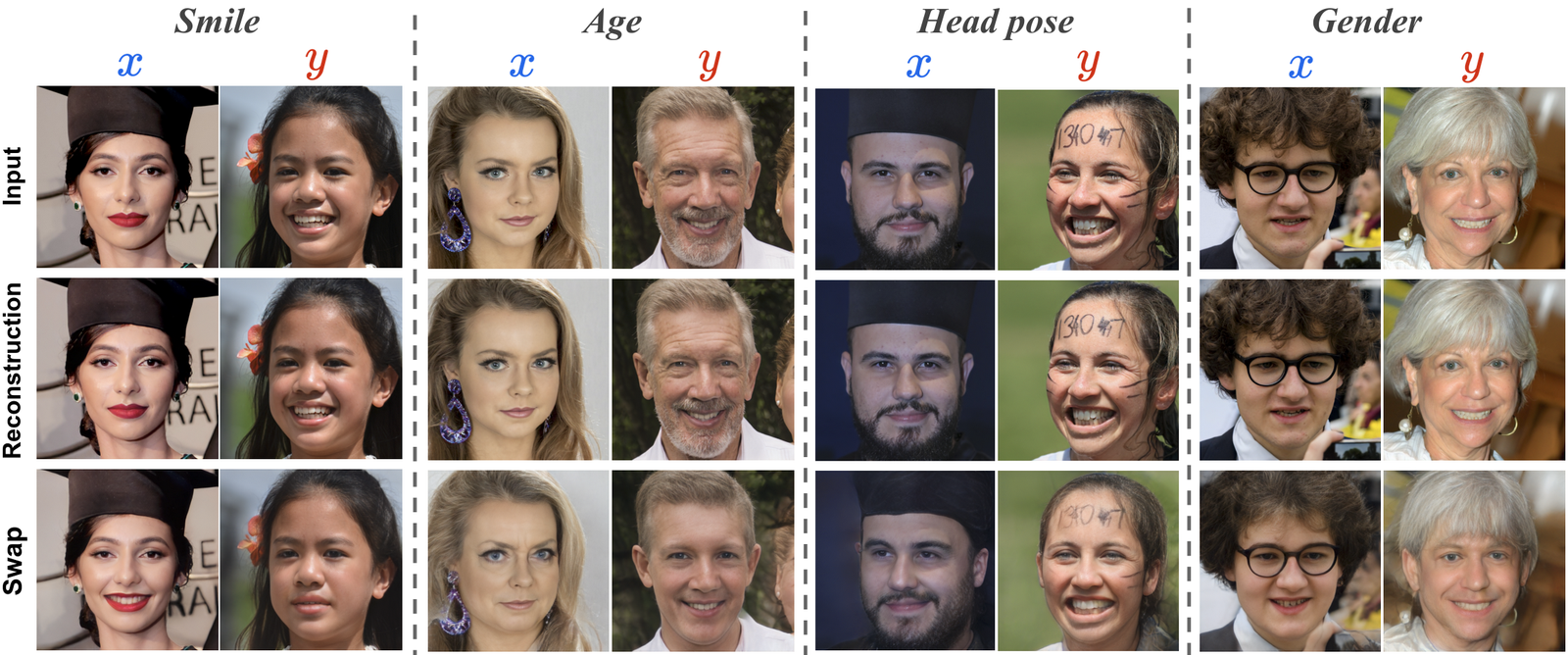}
\end{center}
   \caption{Results of the proposed pSp-cs-Ref (CS-StyleGAN) on four common/salient attributes (smile, age, head pose, gender) under the \textit{background/target} assumption.}
\label{fig:pSp_cs_ref_other_attri}
\end{figure*}

\begin{figure*}[t!]
    \centering
    \includegraphics[width=0.9\linewidth]{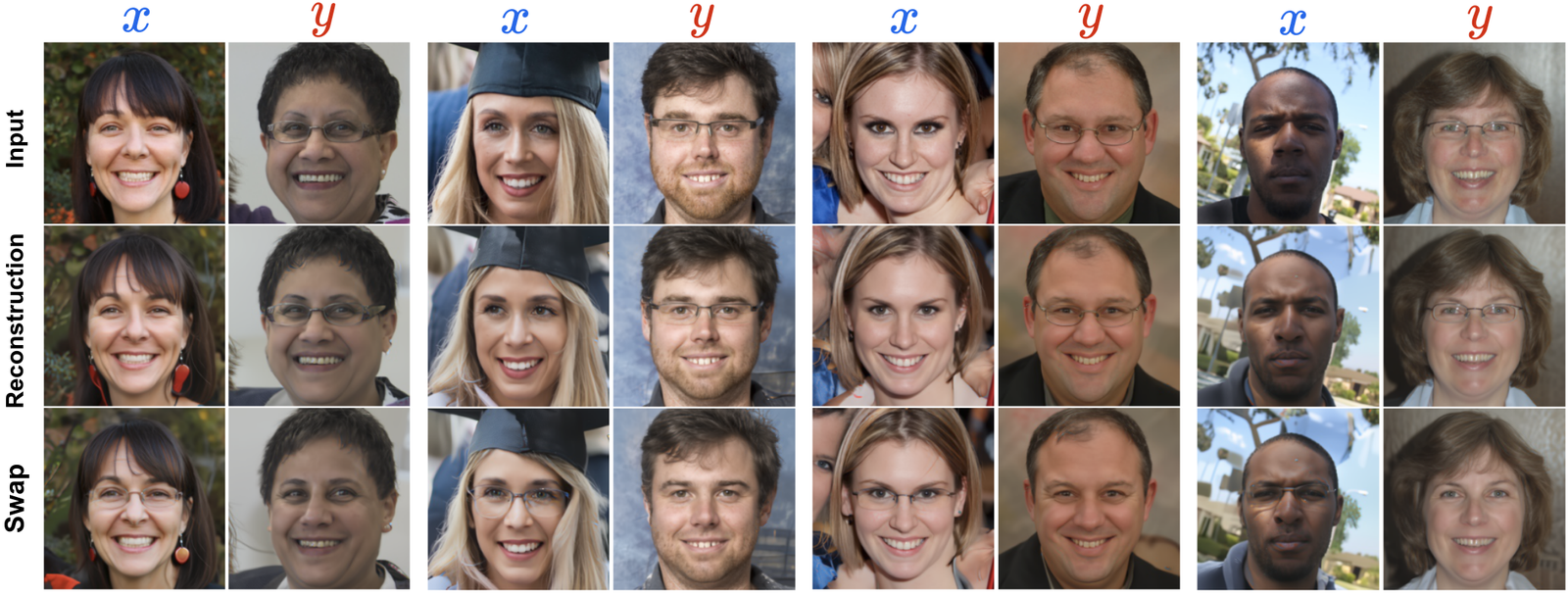}
    \caption{Results of our framework applied to \textbf{DiffAE (DiffAE-cs)}.}
    \label{fig:DiffAE}
\end{figure*}

\begin{figure*}[t!]
\begin{center}
   \includegraphics[width=0.85\linewidth]{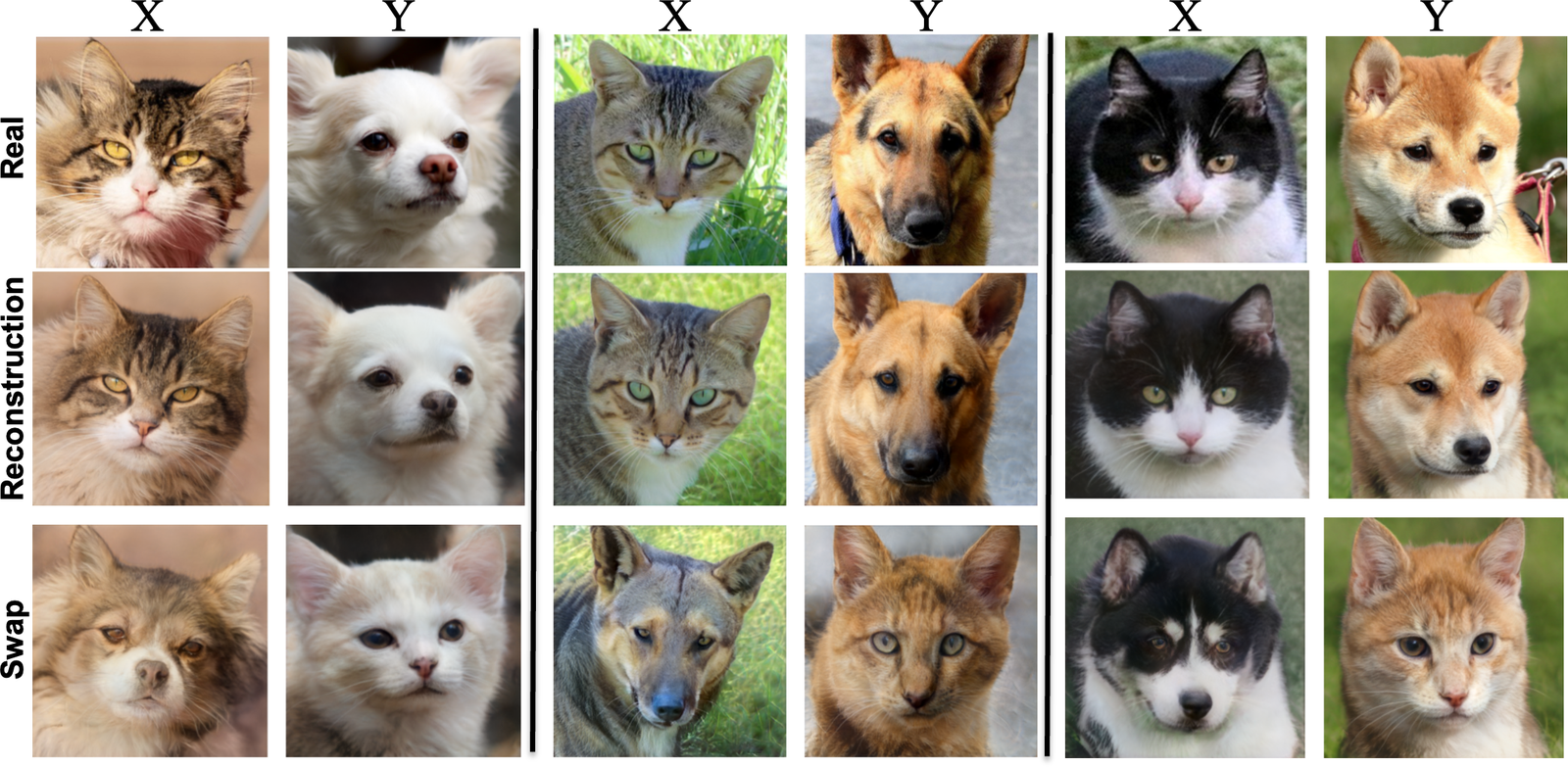}
   \caption{More results on AFHQv2 dataset.}
\label{fig:cat_dogs}
\end{center}
\end{figure*}

\begin{figure*}[ht!]
    \centering
    \includegraphics[width=1.0\linewidth]{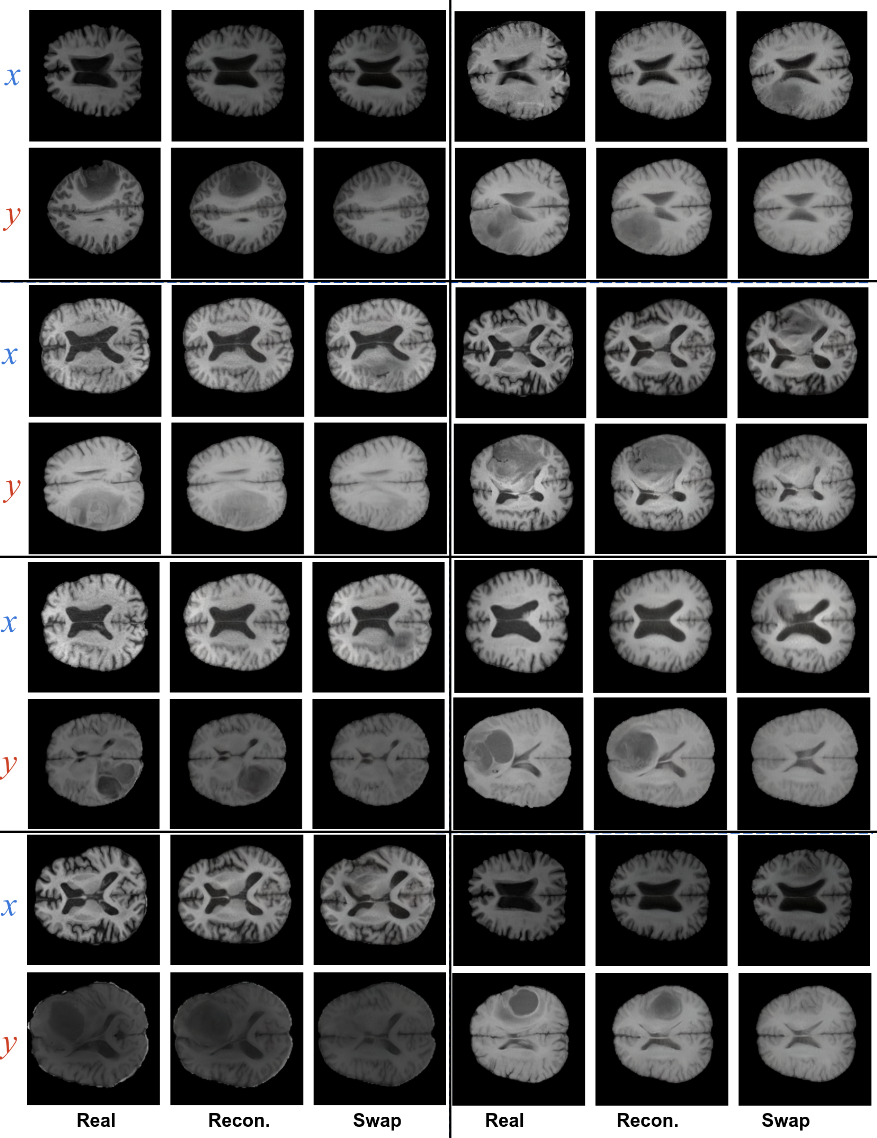}
    \caption{Reconstruction and swap on images of BraTS (brain MRI) datasets.}
    \label{fig:BraTS_sup}
\end{figure*}

\begin{figure*}[t!]
    \centering
    \includegraphics[width=1.0\linewidth]{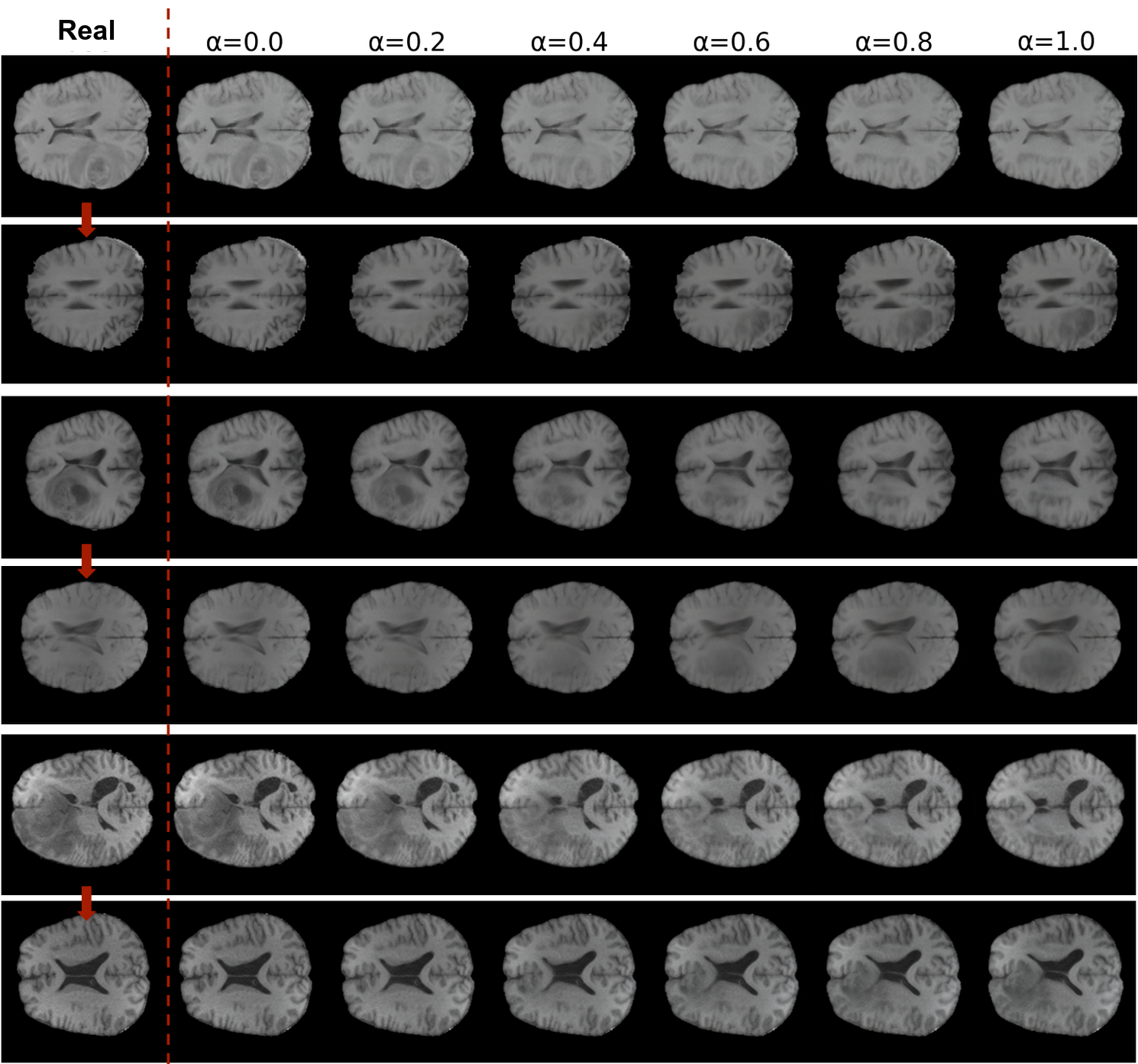}
    \caption{Interpolations between MR images of healthy brains ($X$) and brains with tumors ($Y$). Red arrows indicate transfer of \emph{tumor} from tumored dataset $Y$ to healthy dataset $X$ via salient factor swapping.}
    \label{fig:brats_interp}
\end{figure*}

Figs.~\ref{fig:pSp_cs_ref_other_attri}--\ref{fig:brats_interp} complement \YL{Section~IV-D7} of the main paper by providing results using diverse attributes and on several datasets. 
Fig.~\ref{fig:pSp_cs_ref_other_attri} presents results of pSp-cs-Ref for reconstruction and image swapping on FFHQ smile, age, head pose, and gender attributes. 
Fig.~\ref{fig:DiffAE} shows results of DiffAE-cs for reconstruction and swapping on FFHQ glasses. 
Fig.~\ref{fig:cat_dogs} provides additional results on AFHQv2 (cat $X$ vs.\ dog $Y$). 
Fig.~\ref{fig:BraTS_sup} presents results of reconstructions and salient swaps on the BraTS dataset. 
Fig.~\ref{fig:brats_interp} shows interpolations between MR images of healthy brains and brains with tumors. The results suggest that our model isolates tumor-specific features in the salient space while preserving overall brain morphology. 

\begin{figure*}[t!]
    \centering
    \includegraphics[width=1.0\linewidth]{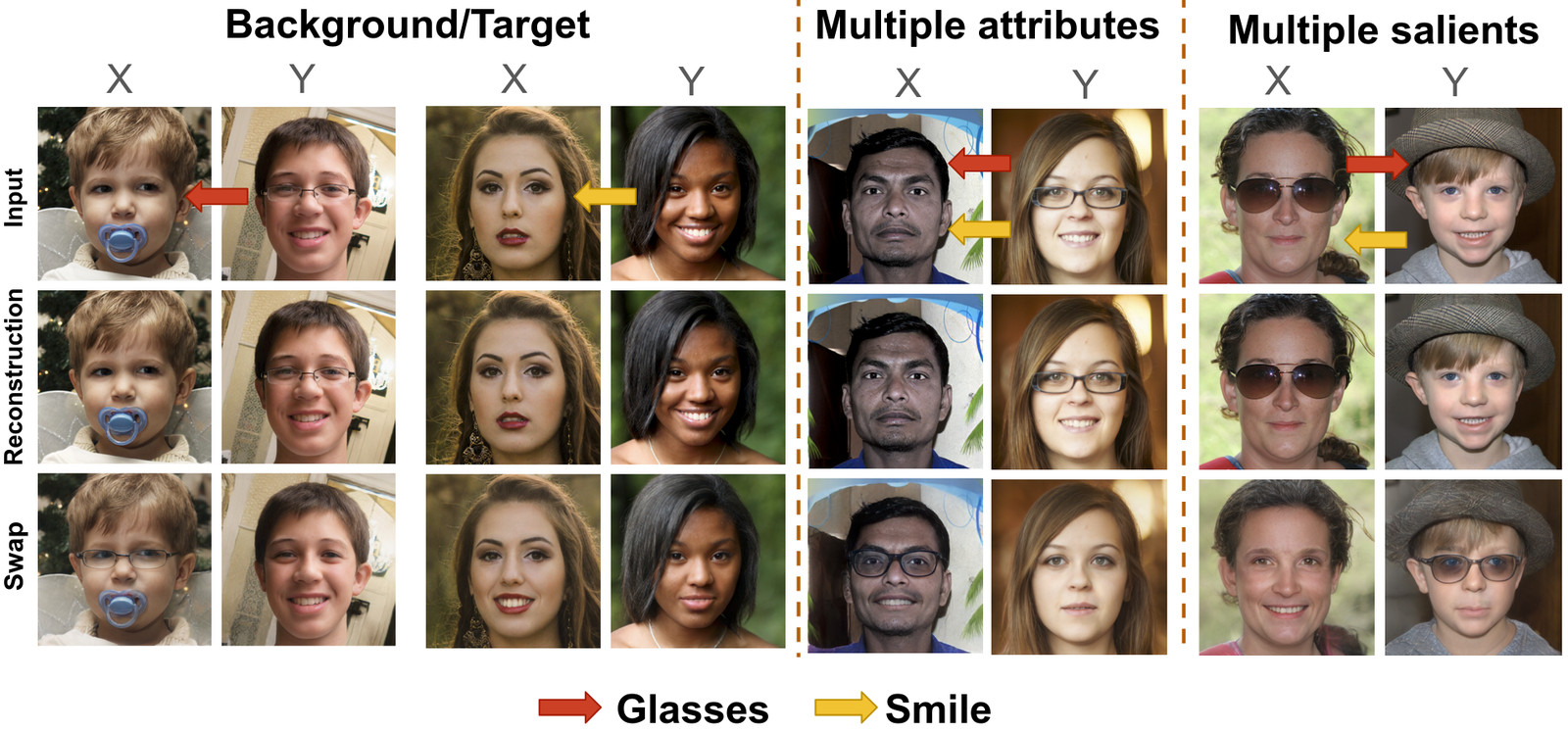}
    \caption{Examples illustrating the results under the \emph{background/target} assumption, where $Y$ contains one/two salient attributes, and the \emph{multiple-salient} assumption, where each of $X$ and $Y$ contains a salient attribute (glasses in $X$ and smiles in $Y$). Red arrows indicate the transfer of \emph{glasses} between $X$ and $Y$ datasets, while yellow arrows represent the transfer of \emph{smile}.}
    \label{fig:bg_t_vs_multiple}
\end{figure*}

Fig.~\ref{fig:bg_t_vs_multiple} shows additional results under the \emph{background/target} setting, \emph{multiple-attribute} and the \emph{multiple-salient} case, complementing the challenging CA scenarios described in \YL{Section~IV-D8} of the main paper. Table~\ref{tab:multi_attr_salient} reports the quantitative image-quality results for the two cases.


\begin{table*}[ht!]
\centering
\caption{Results of latent separation for facial attributes. \textbf{First column}: training attributes (see the main paper, \YL{Table~1} for $X$/$Y$ split details); \textbf{Other columns}: evaluation attributes; cells show 5-fold mean accuracy (\(\pm\) std) on each test set. \textbf{Case 1}: Multiple attributes; \textbf{Case 2}: Multiple salients. \textbf{\(s_1\)} and \textbf{\(s_2\)} denote two salient factors learned in the multiple–salient setting (case 2).}
\label{tab:sep_full_attributes}
\resizebox{\textwidth}{!}{%
\begin{tabular}{l cc cc cc cc cc}
\toprule
 & \multicolumn{2}{c}{No Glasses vs.\ Glasses}
 & \multicolumn{2}{c}{Male vs.\ Female}
 & \multicolumn{2}{c}{Head pose (Frontal vs.\ Turned)}
 & \multicolumn{2}{c}{Smile vs.\ Non-smiling}
 & \multicolumn{2}{c}{Young vs.\ Old} \\
\cmidrule(lr){2-3}\cmidrule(lr){4-5}\cmidrule(lr){6-7}\cmidrule(lr){8-9}\cmidrule(lr){10-11}
\multicolumn{1}{l}{Training Attributes} & C & S & C & S & C & S & C & S & C & S \\
\midrule
Glasses
  & $0.52 \pm 0.032$ & $0.98 \pm 0.005$
  & $0.80 \pm 0.008$ & $0.59 \pm 0.029$
  & $0.95 \pm 0.007$ & $0.52 \pm 0.048$
  & $0.90 \pm 0.009$ & $0.52 \pm 0.010$
  & $0.84 \pm 0.016$ & $0.63 \pm 0.105$ \\
\rowcolor{blue!8}
\textcolor{blue}{Expected}
  & $\textcolor{blue}{0.5}$ & $\textcolor{blue}{1.0}$ & $\textcolor{blue}{1.0}$ & $\textcolor{blue}{0.5}$ & $\textcolor{blue}{1.0}$ & $\textcolor{blue}{0.5}$ & $\textcolor{blue}{1.0}$ & $\textcolor{blue}{0.5}$ & $\textcolor{blue}{1.0}$ & $\textcolor{blue}{0.5}$ \\
\addlinespace
Gender
  & $0.95 \pm 0.006$ & $0.57 \pm 0.060$
  & $0.52 \pm 0.048$ & $0.91 \pm 0.007$
  & $0.94 \pm 0.006$ & $0.53 \pm 0.056$
  & $0.87 \pm 0.009$ & $0.58 \pm 0.008$
  & $0.90 \pm 0.008$ & $0.52 \pm 0.033$ \\
\rowcolor{blue!8}
\textcolor{blue}{Expected}
  & $\textcolor{blue}{1.0}$ & $\textcolor{blue}{0.5}$ & $\textcolor{blue}{0.5}$ & $\textcolor{blue}{1.0}$ & $\textcolor{blue}{1.0}$ & $\textcolor{blue}{0.5}$ & $\textcolor{blue}{1.0}$ & $\textcolor{blue}{0.5}$ & $\textcolor{blue}{1.0}$ & $\textcolor{blue}{0.5}$ \\
\addlinespace
Head pose
  & $0.95 \pm 0.003$ & $0.56 \pm 0.071$
  & $0.80 \pm 0.015$ & $0.52 \pm 0.042$
  & $0.56 \pm 0.112$ & $0.98 \pm 0.005$
  & $0.82 \pm 0.013$ & $0.60 \pm 0.056$
  & $0.90 \pm 0.016$ & $0.58 \pm 0.066$ \\
\rowcolor{blue!8}
\textcolor{blue}{Expected}
  & $\textcolor{blue}{1.0}$ & $\textcolor{blue}{0.5}$ & $\textcolor{blue}{1.0}$ & $\textcolor{blue}{0.5}$ & $\textcolor{blue}{0.5}$ & $\textcolor{blue}{1.0}$ & $\textcolor{blue}{1.0}$ & $\textcolor{blue}{0.5}$ & $\textcolor{blue}{1.0}$ & $\textcolor{blue}{0.5}$ \\
\addlinespace
Smile
  & $0.94 \pm 0.005$ & $0.52 \pm 0.044$
  & $0.79 \pm 0.008$ & $0.53 \pm 0.057$
  & $0.94 \pm 0.005$ & $0.58 \pm 0.098$
  & $0.57 \pm 0.058$ & $0.94 \pm 0.005$
  & $0.91 \pm 0.007$ & $0.55 \pm 0.066$ \\
\rowcolor{blue!8}
\textcolor{blue}{Expected}
  & $\textcolor{blue}{1.0}$ & $\textcolor{blue}{0.5}$ & $\textcolor{blue}{1.0}$ & $\textcolor{blue}{0.5}$ & $\textcolor{blue}{1.0}$ & $\textcolor{blue}{0.5}$ & $\textcolor{blue}{0.5}$ & $\textcolor{blue}{1.0}$ & $\textcolor{blue}{1.0}$ & $\textcolor{blue}{0.5}$ \\
\addlinespace
Age
  & $0.91 \pm 0.004$ & $0.65 \pm 0.102$
  & $0.70 \pm 0.022$ & $0.58 \pm 0.070$
  & $0.94 \pm 0.005$ & $0.54 \pm 0.074$
  & $0.84 \pm 0.007$ & $0.59 \pm 0.047$
  & $0.54 \pm 0.083$ & $0.97 \pm 0.003$ \\
\rowcolor{blue!8}
\textcolor{blue}{Expected}
  & $\textcolor{blue}{1.0}$ & $\textcolor{blue}{0.5}$ & $\textcolor{blue}{1.0}$ & $\textcolor{blue}{0.5}$ & $\textcolor{blue}{1.0}$ & $\textcolor{blue}{0.5}$ & $\textcolor{blue}{1.0}$ & $\textcolor{blue}{0.5}$ & $\textcolor{blue}{0.5}$ & $\textcolor{blue}{1.0}$ \\
\addlinespace
Glasses \& Smile (case 1)
  & $0.54 \pm 0.073$ & $0.97 \pm 0.004$
  & $0.78 \pm 0.006$ & $0.57 \pm 0.059$
  & $0.94 \pm 0.006$ & $0.59 \pm 0.073$
  & $0.56 \pm 0.051$ & $0.91 \pm 0.004$
  & $0.87 \pm 0.011$ & $0.54 \pm 0.072$ \\
\rowcolor{blue!8}
\textcolor{blue}{Expected}
  & $\textcolor{blue}{0.5}$ & $\textcolor{blue}{1.0}$ & $\textcolor{blue}{1.0}$ & $\textcolor{blue}{0.5}$ & $\textcolor{blue}{1.0}$ & $\textcolor{blue}{0.5}$ & $\textcolor{blue}{0.5}$ & $\textcolor{blue}{1.0}$ & $\textcolor{blue}{1.0}$ & $\textcolor{blue}{0.5}$ \\
\addlinespace
Glasses \& Smile (case 2, \(s_1\))
  & $0.53 \pm 0.062$ & $0.90 \pm 0.012$
  & $0.81 \pm 0.007$ & $0.52 \pm 0.046$
  & $0.93 \pm 0.010$ & $0.55 \pm 0.095$
  & $0.57 \pm 0.058$ & $0.56 \pm 0.058$
  & $0.81 \pm 0.018$ & $0.53 \pm 0.067$ \\
\rowcolor{blue!8}
\textcolor{blue}{Expected}
  & $\textcolor{blue}{0.5}$ & $\textcolor{blue}{1.0}$ & $\textcolor{blue}{1.0}$ & $\textcolor{blue}{0.5}$ & $\textcolor{blue}{1.0}$ & $\textcolor{blue}{0.5}$ & $\textcolor{blue}{0.5}$ & $\textcolor{blue}{0.5}$ & $\textcolor{blue}{1.0}$ & $\textcolor{blue}{0.5}$ \\
Glasses \& Smile (case 2, \(s_2\))
  &  & $0.54 \pm 0.022$
  &  & $0.52 \pm 0.034$
  &  & $0.56 \pm 0.122$
  & & $0.91 \pm 0.008$
  & & $0.54 \pm 0.094$ \\
\rowcolor{blue!8}  
\textcolor{blue}{Expected}
  & $\textcolor{blue}{}$ & $\textcolor{blue}{0.5}$ & $\textcolor{blue}{}$ & $\textcolor{blue}{0.5}$ & $\textcolor{blue}{}$ & $\textcolor{blue}{0.5}$ & $\textcolor{blue}{}$ & $\textcolor{blue}{1.0}$ & $\textcolor{blue}{}$ & $\textcolor{blue}{0.5}$ \\
\bottomrule
\end{tabular}
}
\end{table*}

\begin{table*}[ht!]
\centering
\renewcommand{\arraystretch}{1.3}
\setlength{\tabcolsep}{6pt}
\caption{Performance of common and salient separation on AFHQ, BraTS, and CelebA-HQ datasets. $\Delta = |0.5-C| + |1.0-S|$.}
\label{tab:cs_sep_4datasets}
\resizebox{\textwidth}{!}{%
\begin{tabular}{l *{3}{c} *{3}{c} *{3}{c} *{3}{c}}
\toprule
& \multicolumn{3}{c}{\textbf{BraTS}} &
  \multicolumn{3}{c}{\textbf{AFHQ}} &
  \multicolumn{3}{c}{\textbf{CelebA-HQ Gender}} &
  \multicolumn{3}{c}{\textbf{CelebA-HQ Smile}} \\
\cmidrule(lr){2-4}\cmidrule(lr){5-7}\cmidrule(lr){8-10}\cmidrule(lr){11-13}
\textbf{Model} & $C$ & $S$ & $\Delta$ & $C$ & $S$ & $\Delta$ & $C$ & $S$ & $\Delta$ & $C$ & $S$ & $\Delta$ \\
\midrule
\textbf{Double InfoGAN} &
0.65 $\pm$ 0.027 & 0.73 $\pm$ 0.026 & 0.42 &
0.78 $\pm$ 0.022 & 0.82 $\pm$ 0.029 & 0.46 &
0.74 $\pm$ 0.004 & 0.84 $\pm$ 0.013 & 0.40 &
0.79 $\pm$ 0.007 & 0.80 $\pm$ 0.009 & 0.49 \\
\textbf{Ours} &
\textbf{0.54 $\pm$ 0.086} & \textbf{0.98 $\pm$ 0.008} & \textbf{0.06} &
\textbf{0.56 $\pm$ 0.114} & \textbf{0.99 $\pm$ 0.004} & \textbf{0.07} &
\textbf{0.59 $\pm$ 0.111} & \textbf{0.97 $\pm$ 0.003} & \textbf{0.12} &
\textbf{0.57 $\pm$ 0.057} & \textbf{0.84 $\pm$ 0.005} & \textbf{0.23} \\
\textbf{\textcolor{blue}{Expected}} &
\textcolor{blue}{0.5} & \textcolor{blue}{1.0} & \textcolor{blue}{0} &
\textcolor{blue}{0.5} & \textcolor{blue}{1.0} & \textcolor{blue}{0} &
\textcolor{blue}{0.5} & \textcolor{blue}{1.0} & \textcolor{blue}{0} &
\textcolor{blue}{0.5} & \textcolor{blue}{1.0} & \textcolor{blue}{0} \\
\bottomrule
\end{tabular}%
} 
\end{table*}

\begin{table*}[ht!]
    \caption{Per-image reconstruction time (seconds) for different models.}
    \label{tab:inference_time}
    \centering
    \resizebox{0.9\linewidth}{!}{
    \begin{tabular}{lcccccccc}
        \toprule
        Method & pSp & pSp-cs & pSp-cs-Ref & TIME & Asyrp & h-space-cs & DiffAE & DiffAE-cs  \\
        \midrule
        Time (s/image) & 0.03 & 0.04 & 0.18 & 18.46  & 6.82 & 7.46 & 22.75 & 24.81 \\
        \bottomrule
    \end{tabular}
    }
\end{table*}

\begin{table*}[ht!]
\centering
\caption{Pearson correlations between the row attribute defining the training datasets ($X$ vs. $Y$) and each column attribute within those training datasets. For each attribute pair, both attributes are labeled as 0/1 ($X{=}0$, $Y{=}1$); entries report Pearson’s $r$ on the binary vectors. 
}
\label{tab:phi_corr}
\resizebox{\linewidth}{!}{%
\begin{tabular}{l *{5}{c}}
\toprule
\multicolumn{1}{c}{\textbf{Training datasets (X vs.\ Y)}} &
\multicolumn{1}{c}{No Glasses vs.~Glasses} &
\multicolumn{1}{c}{Male vs.~Female} &
\multicolumn{1}{c}{Head pose (Frontal vs.~Turned)} &
\multicolumn{1}{c}{Smile vs.~Non\mbox{-}smiling} &
\multicolumn{1}{c}{Young vs.~Old} \\
\midrule
No Glasses vs.~Glasses         &  1.000 & -0.195 & -0.014 & -0.012 &  \textbf{0.301} \\
Male vs.~Female                & -0.043 &  1.000 &  0.011 &  0.170 &  0.001 \\
Head pose (Frontal vs.~Turned) & -0.024 &  0.056 &  1.000 & \textbf{-0.224} &  0.035 \\
Smile vs.~Non\mbox{-}smiling   &  0.007 &  0.002 & -0.158 &  1.000 &  0.034 \\
Young vs.~Old                  &  \textbf{0.366} &  0.001 &  0.115 &  0.076 &  1.000 \\
\bottomrule
\end{tabular}}
\end{table*}

\begin{figure*}[ht!]
\begin{center}
   \includegraphics[width=1.0\linewidth]{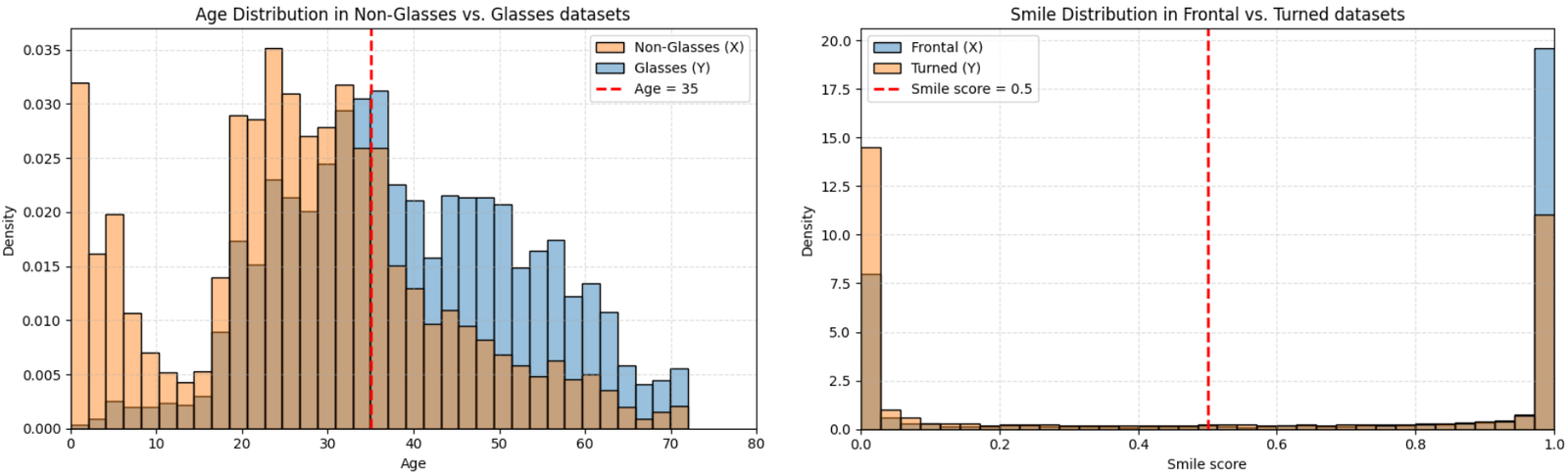}
   \caption{Attribute imbalance in the training data. Right: age distributions in Non-Glasses ($X$) vs. Glasses ($Y$) datasets. Bottom: smile-score distributions in Frontal ($X$) vs. Turned poses ($Y$) datasets.}
\label{fig:data_bias}
\end{center}
\end{figure*}

\begin{table*}[ht!]
\caption{Quantitative results for the multiple-attributes (case~1) and multiple-salients (case~2) settings.}
\label{tab:multi_attr_salient}
\centering
\footnotesize
\resizebox{\textwidth}{!}{%
\begin{tabular}{c c
                *{5}{c}
                *{5}{c}
                *{4}{c}}
\toprule
& &
\multicolumn{5}{c}{\textbf{Reconstruction (X)}} &
\multicolumn{5}{c}{\textbf{Reconstruction (Y)}} &
\multicolumn{2}{c}{\textbf{Swap X$\rightarrow$Y}} &
\multicolumn{2}{c}{\textbf{Swap Y$\rightarrow$X}} \\
\cmidrule(lr){3-7}\cmidrule(lr){8-12}\cmidrule(lr){13-14}\cmidrule(lr){15-16}
\textbf{Setting} & \textbf{Model} &
LPIPS$\downarrow$ & MSE$\downarrow$ & MS\mbox{-}SSIM$\uparrow$ & ID\mbox{-}Sim$\uparrow$ & FID$\downarrow$ &
LPIPS$\downarrow$ & MSE$\downarrow$ & MS\mbox{-}SSIM$\uparrow$ & ID\mbox{-}Sim$\uparrow$ & FID$\downarrow$ &
ID\mbox{-}sim$\uparrow$ & FID\mbox{-}Y$\downarrow$ &
ID\mbox{-}sim$\uparrow$ & FID\mbox{-}X$\downarrow$ \\
\midrule
\multirow{2}{*}{\begin{tabular}{c} Multiple Attributes \\ (case 1) \end{tabular}} 
& pSp\mbox{-}cs       
& 0.211 & 0.017 & 0.675 & 0.779 & 55.137
& 0.185 & 0.015 & 0.699 & 0.837 & 43.670
& 0.609 & 58.613
& 0.675 & 68.422 \\

& pSp\mbox{-}cs\mbox{-}Ref 
& 0.020 & 0.001 & 0.982 & 0.992 & 5.859
& 0.021 & 0.001 & 0.982 & 0.992 & 5.662
& 0.759 & 39.857
& 0.773 & 47.328 \\
\midrule
\multirow{2}{*}{\begin{tabular}{c} Multiple Salients \\ (case 2) \end{tabular}} 
& pSp\mbox{-}cs       
& 0.212 & 0.020 & 0.659 & 0.783 & 76.613
& 0.204 & 0.019 & 0.674 & 0.791 & 81.204
& 0.590 & 65.422
& 0.650 & 70.755 \\

& pSp\mbox{-}cs\mbox{-}Ref 
& 0.021 & 0.001 & 0.981 & 0.991 & 10.854
& 0.020 & 0.001 & 0.981 & 0.991 & 11.580
& 0.729 & 45.695
& 0.719 & 48.828 \\
\bottomrule
\end{tabular}%
}
\end{table*}

\end{document}